\documentclass[manuscript]{acmart}

\AtBeginDocument{%
  }

\acmConference[]{}{}{}

\usepackage{multirow}
\usepackage{booktabs}
\usepackage{caption,subcaption}

\usepackage{algorithmic}
\usepackage{color}
\usepackage{adjustbox}
\usepackage{tikz}
\usepackage{makecell}
\usepackage{tabularx}
\usepackage{bm}
\usepackage{amsmath}
\usepackage{ising_v1}
\def\submission{1}

\usepackage{soul}

\begin{document}

\title{A Survey on Vertical Federated Learning: From a Layered Perspective}

\author{Liu Yang}
\email{lyangau@cse.ust.hk}
\affiliation{%
  \institution{Hong Kong University of Science and Technology}
  \city{Hong Kong}
  \country{China}
}

\author{Di Chai}
\email{dchai@cse.ust.hk}
\affiliation{%
  \institution{Hong Kong University of Science and Technology}
  \city{Hong Kong}
  \country{China}
}

\author{Junxue Zhang}
\email{jzhangcs@connect.ust.hk}
\affiliation{%
  \institution{Hong Kong University of Science and Technology}
  \city{Hong Kong}
  \country{China}
}
\affiliation{%
  \institution{Clustar Co., Ltd}
  \city{Shenzhen}
  \country{China}
}

\author{Yilun Jin}
\email{yilun.jin@connect.ust.hk}
\affiliation{%
  \institution{Hong Kong University of Science and Technology}
  \city{Hong Kong}
  \country{China}
}

\author{Leye Wang}
\email{leyewang@pku.edu.cn}
\affiliation{%
  \institution{Peking University}
  \city{Beijing}
  \country{China}
}

\author{Hao Liu}
\email{liuh@ust.hk}
\affiliation{%
  \institution{Hong Kong University of Science and Technology (Guangzhou)}
  \city{Guangzhou}
  \country{China}
}

\author{Han Tian}
\email{htianab@cse.ust.hk}
\affiliation{%
  \institution{Hong Kong University of Science and Technology}
  \city{Hong Kong}
  \country{China}
}

\author{Qian Xu}
\email{qianxu@ust.hk}
\affiliation{%
  \institution{Hong Kong University of Science and Technology (Guangzhou)}
  \city{Shenzhen}
  \country{China}
}

\author{Kai Chen}
\email{kaichen@cse.ust.hk}
\affiliation{%
  \institution{Hong Kong University of Science and Technology}
  \city{Hong Kong}
  \country{China}
}


\renewcommand{\shortauthors}{Yang et al.}


\begin{abstract}
Vertical federated learning (VFL) is a promising category of federated learning for the scenario where data is vertically partitioned and distributed among parties. VFL enriches the description of samples using features from different parties to improve model capacity. Compared with horizontal federated learning, in most cases, VFL is applied in the commercial cooperation scenario of companies. Therefore, VFL contains tremendous business values. In the past few years, VFL has attracted more and more attention in both academia and industry. In this paper, we systematically investigate the current work of VFL from a layered perspective. From the hardware layer to the vertical federated system layer, researchers contribute to various aspects of VFL. Moreover, the application of VFL has covered a wide range of areas,~\eg, finance, healthcare,~\etc. At each layer, we categorize the existing work and explore the challenges for the convenience of further research and development of VFL. Especially, we design a novel MOSP tree taxonomy to analyze the core component of VFL,~\ie, secure vertical federated machine learning algorithm. Our taxonomy considers four dimensions,~\ie, machine learning model (M), protection object (O), security model (S), and privacy-preserving protocol (P), and provides a comprehensive investigation.
\end{abstract}

\ccsdesc[300]{Security and privacy~Distributed systems security}
\ccsdesc[300]{Computing methodologies~Cooperation and coordination}
\ccsdesc[500]{Security and privacy~Privacy-preserving protocols}

\keywords{vertical federated learning, survey}


\maketitle

\section{Introduction}
\label{sec:introduction}
Federated learning is an emerging field of secure distributed machine learning, which addresses data silo and privacy problems together~\cite{mcmahan2021advances}. Nowadays, machine learning has become ubiquitous, with applications ranging from video recommendations~\cite{covington2016deep} to facial recognition~\cite{parkhi2015deep}. Consequently, the performance of machine learning algorithms is crucial, which is largely affected by the quality and quantity of training data. Unfortunately, in the real world, data are always distributed over different data silos. Privacy concerns and security regulations such as the GDPR~\footnote{GDPR is a regulation in EU law on data protection and privacy in the European Union and the European Economic Area. https://gdpr.eu/.} make it impossible to collect data directly for centralized model training. Under this circumstance, Google introduced cross-device federated learning (FL)~\cite{mcmahan2017communication}, which enables device-level model training for keyboard input prediction without exchanging raw training data. Yang~\textit{et al.} generalized the FL concept to cross-silo scenarios where participants are organizations or companies~\cite{yang2019federated}, meanwhile categorizing FL into horizontal federated learning (HFL), vertical federated learning (VFL), and federated transfer learning (FTL) according to different data partition situations. 




 


\begin{figure}[t]
\centering
\includegraphics[width=0.6\linewidth]{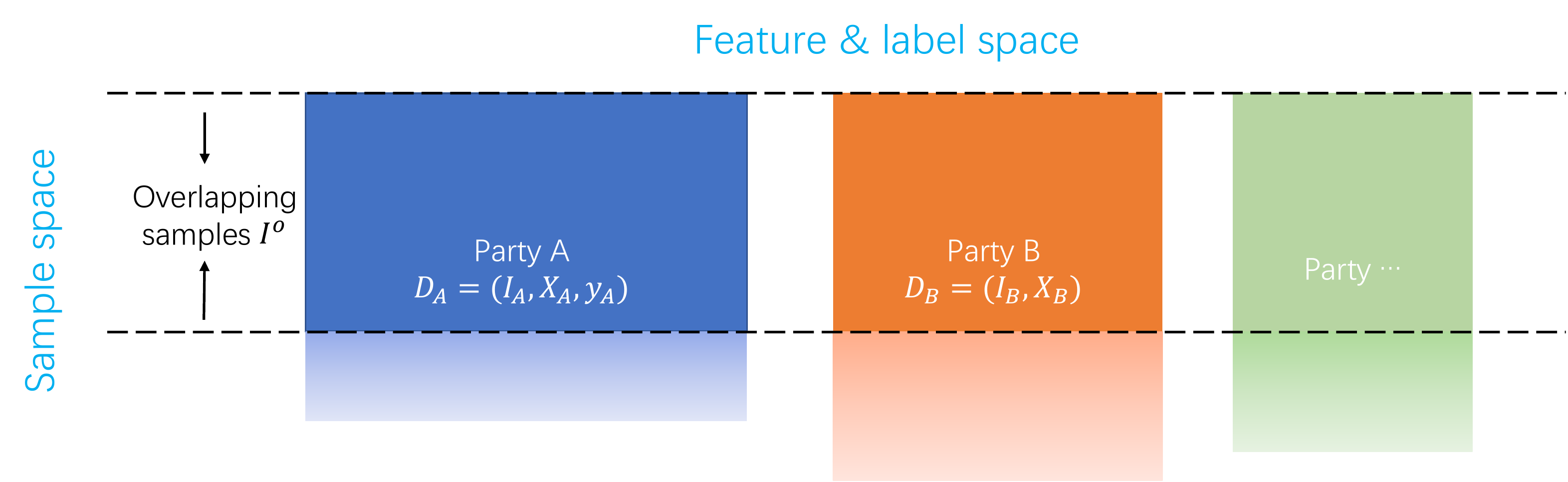}
\caption{The scenario of vertical federated learning. Participants contain overlapping samples but different feature spaces. They want to collaboratively train an effective machine learning model which utilizes all their features with no mutual leakage of training data.}
\vspace{-0.3cm}
\label{fig:scenario_VFL}
\end{figure}

Vertical federated learning (VFL) aims to collaboratively train machine learning models among parties with vertically-partitioned data without privacy leakage, as shown in Fig.~\ref{fig:scenario_VFL}. Descriptions of samples could be enriched with different aspects of features from various parties. We take the VFL cooperation between a traditional bank $A$ and an Internet company $B$ as an example, which is a common application for VFL. Bank $A$ wants to train a machine learning model to predict customers' repayment ability. However, it is difficult for bank $A$ to achieve the goal independently due to its one-sidedness and scarcity of local features. On the one hand, bank $A$ only has financial behavior data,~\eg, loan and transaction records. The knowledge provided by the financial features is insufficient to find all qualified customers because of one-sidedness. On the other hand, the financial behavior data of bank A only exists for a small part of customers. In reality, not all customers maintain frequent interactions with the bank. For instance, many customers are not qualified to apply for a mortgage loan. Hence, the financial features cannot cover the whole customer group because of scarcity. Meanwhile, company $B$ owns abundant Internet behavior data,~\eg, social networks, which could supplement the financial features. Therefore, bank $A$ and company $B$ could cooperatively train a better model via VFL, which can be regarded as the privacy-preserving model parallelism~\cite{shen2020distributed,liu2021distributed} or feature-distributed learning~\cite{zhang2018feature}.

In this paper, we provide a comprehensive investigation of the current VFL research and aim to facilitate the growth of VFL since VFL has been less explored by both industry and academia but is more common in cross-silo scenarios with huge commercial value~\cite{chen2021homomorphic,fink2021artificial}. As an  interdisciplinary domain of machine learning, cryptography, and Economics~\cite{lu2022truthful,zhang2022data}, VFL 
involves many aspects, ranging from low-level physical deployments to high-level applications. Particular work could only cover one or several aspects of VFL. Thus current research or the industrial community lacks a complete view of VFL designs. We divide VFL into five layers,~\ie, hardware (layer one~\S\ref{sec:hardware_design}), privacy-preserving primitive (layer two~\S\ref{sec:primitive_design}), secure vertical federated algorithm (layer three~\S\ref{sec:protocol_design}), vertical federated system (layer four~\S\ref{sec:framework_design}), and application (layer five~\S\ref{sec:application_design}), which provides a clear view to our readers of how a real-world VFL framework is constructed. We categorize and analyze the current works, as well as summarize the promising directions in each layer. Particularly for the secure vertical federated machine learning algorithms, the core component of VFL, in layer three, we design a novel taxonomy, considering four principal dimensions.

\section{Overview}
This section describes the scenario and definition of vertical federated learning, explains its general architecture, and summarizes the designs at different layers.
\label{sec:VFL}

\vspace{-0.2cm}

\subsection{Scenario and Definition}

There could be multiple parties in the vertical federated learning scenario, shown in Fig.~\ref{fig:scenario_VFL}. Without loss of generality, we take the two-party situation as an example. Data $\mathcal{D}_A = \{ \mathcal{I}_A, {\bf X}_A \in \mathbb{R}^{|\mathcal{I}_A| \times d_B}, {\bf y}_A \in \mathbb{R}^{|\mathcal{I}_A| \times 1} \}$ and $\mathcal{D}_B = \{ \mathcal{I}_B, {\bf X}_B \in \mathbb{R}^{|\mathcal{I}_B| \times d_B}\}$ are distributed in party A and party B respectively. Their feature spaces $\mathbb{R}^{d_A}$ and $\mathbb{R}^{d_B}$ are different. However, their sample spaces have overlappings $
	\mathcal{I}^o \subseteq \mathcal{I}_A, \mathcal{I}^o \subseteq \mathcal{I}_B$. Party A is often called active party because of containing labels, while party B is called passive party.

\begin{definition}
	Assume $N$ parties whose data $\{ \mathcal{D}_i \}_i^N$ contain common entities $\mathcal{I}^o$ but different feature space $\mathbb{R}^{d_i}$. Vertical federated learning (VFL) aims to collaboratively train a machine learning model $\{ \bm{\theta}_i^{\rm{VFL}} \}_i^N$ among $N$ parties to improve the model performance of one specific prediction task $\bm{y}_k$ without mutual leakage of original data $\{ \mathcal{D}_i \}_i^N$:
	
	\begin{equation}
	    \{ \bm{\hat{\theta}}_i^{\rm{VFL}} \}_i^N = \mathop{\arg\min}_{\{ \bm{\theta}_i^{\rm{VFL}} \}_i^N} L(\bm{y}_k, f(\{ \bm{X}_i \}_i^N | \{ \bm{\theta}_i^{\rm{VFL}} \}_i^N)).
	    \label{eqn:vfl-formulation}
	\end{equation}
\end{definition}

The performance of the vertical federated model $\{ \bm{\theta}_i^{\rm{VFL}} \}_i^N$ should be better than the performance of the local model trained only on each data $\bm{X}_k$, and be very close to the performance of the global model trained on the centrally collected data. Besides, private training data is securely maintained locally at each party. Only intermediate results calculated for updating models over the original data are exchanged among parties. Furthermore, VFL utilizes privacy-preserving methods,~\eg, homomorphic encryption~\cite{gentry2009fully} and differential privacy~\cite{dwork2014algorithmic}, to protect the intermediate results.

\vspace{-0.2cm}

\subsection{General Procedure}
\begin{figure*}[htbp]
\centering
\includegraphics[width=0.55\linewidth]{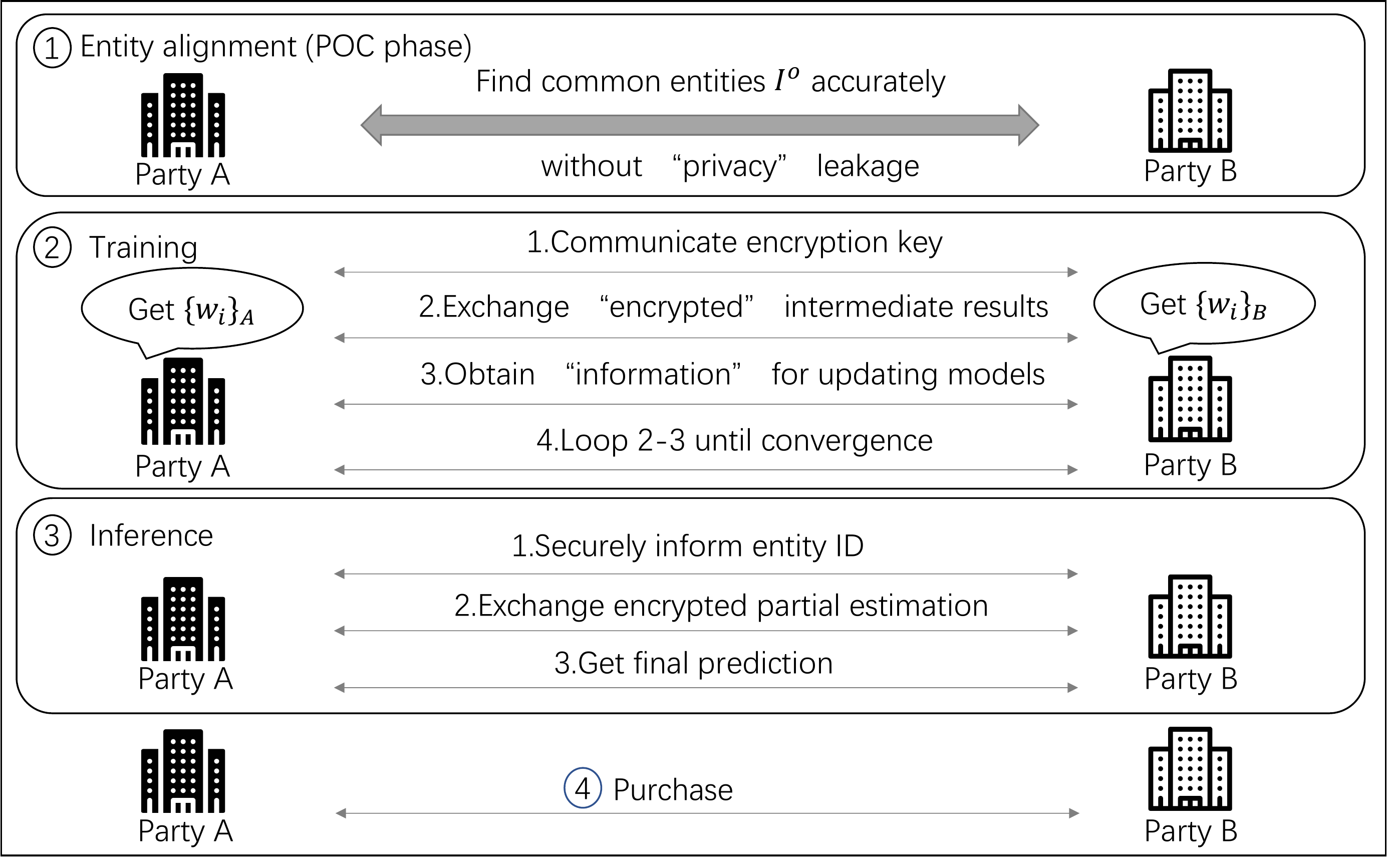}
\caption{The general procedure of vertical federated learning. There are four phases,~\ie, entity alignment, training, inference, and purchase phases by sequence. The purchase operations could exist in any of the previous three phases.}
\vspace{-0.4cm}
\label{fig:architecture_VFL}
\end{figure*}

In a general vertical federated learning task, there are four steps,~\ie, entity alignment, training, inference, and purchase phases. In this subsection, we introduce the details of them one by one. The details are presented in Fig.~\ref{fig:architecture_VFL}. For a better illustration, we also take a two-party task as an example.

\subsubsection{Entity Alignment Phase.}
In the very beginning, party A and party B carry out id matching procedure to find common entities $I^o$. One of the easiest methods can be implemented based on only the one-way hash function~\footnote{https://en.wikipedia.org/wiki/Cryptographic\_hash\_function}. First, both parties map their sample IDs into hash values using the same hash function. Next, party A sends its hash values to party B. Then, party B compares them with its own ones and obtains the ID intersection. However, this method is not safe enough because party B could get all the sample IDs of party A via a brute-force search.
To further protect ID privacy, a more standard method utilizes the blind signature scheme of RSA~\cite{liang2004privacy}. Firstly, both parties apply a hash function to their IDs. Secondly, party A creates RSA keys and sends the public key to party B. Thirdly, party B blinds its IDs with randomly generated numbers and sends them to party A. Fourthly, party A signs the received messages and its own sample IDs to get two signature sets separately, and sends them back. Finally, party B unblinds the first signature set, compares it with the second set, and obtains the ID intersection. This method could protect the privacy of IDs not in the intersection set.

\subsubsection{Training Phase.}
Vertical federated training is conducted on the overlapping samples $I^o$. In general, the procedure can be divided into four steps. And third-party arbiters may be incorporated in some scenarios. We take the vertical federated linear regression~\cite{yang2019federated} as an example. First, the arbiter C generates encryption keys and sends public keys to party A and party B. Second, party A and party B calculate, encrypt, and exchange intermediate results separately. The intermediate results are used to compute gradients and loss. Third, party A and B calculate gradients in the ciphertext state, add masks, and send them to arbiter C for decryption. Party B also calculates encrypted loss and sends it to arbiter C. Finally, both parties receive the decrypted masked values from arbiter C, remove the additional masks and update local models. The training procedure goes on if the federated model does not converge in terms of loss.

Before the training process, we could also adopt vertical federated feature engineering. For example, vertical federated feature binning and selection are both implemented in FATE~\cite{liu2021fate}. Vertical federated feature binning converts numerical features into categorical features. Then, the Weight of Evidence (WoE) and Information Value (IV) could be calculated for each feature~\cite{weed2005weight}. The party without label information must securely communicate with the party with labels to compute WoE and IV. Next, vertical federated feature selection could be performed to filter irrelevant features for model performance improvement according to WoE and IV.

\subsubsection{Inference Phase.}
After obtaining a well-trained federated model, we could conduct inference for the overlapping samples. At first, arbiter C sends the sample ID to party A and B. Then, each party computes its point estimate for this sample and sends it to arbiter C. Finally, arbiter C could add these two values to obtain the final result.

\subsubsection{Purchase Phase.}
The participants of VFL could be divided into the model users and data providers. In our example, party A is the model user and party who owns labels and needs auxiliary features to improve model performance, while B is the data provider that has abundant features. Since VFL is typically driven by commercial applications, the model users need to pay a certain amount of money to the data owner as a payoff. If we go into detail, many issues need to be solved during the purchase phase,~\eg, how much the model users need to pay to the data providers, how to allocate the money fairly if there is more than one data provider,~\etc. And we denote all these problems as the incentive mechanism design further introduced in section~\S\ref{sec:incentive_mechanism}. It is worth noting that the purchase phase is not necessarily carried out after the training and inference phase. A recent study~\cite{yu2020sustainable} has shown that decreasing the payment delay (~\eg, paying by each training iteration) is beneficial and can reduce the risk of exceeding the budget.

\subsection{Layered Approach}
\label{sec:layered}

\begin{figure*}[htbp]
\centering
\vspace{-0.2cm}
\includegraphics[width=0.7\linewidth]{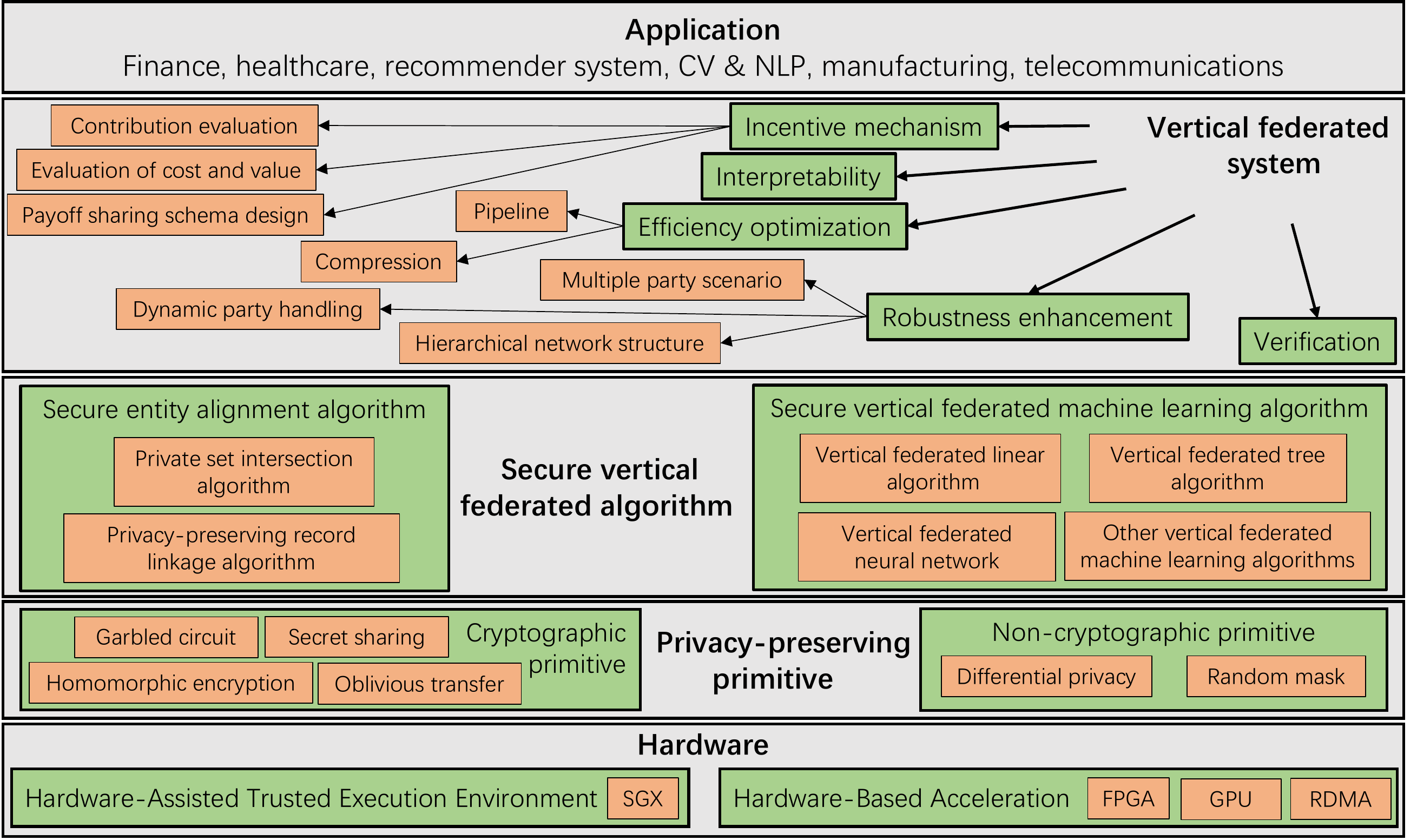}
\caption{The designs of vertical federated learning in a layered approach. From bottom to top, the techniques could be divided into five layers,~\ie, hardware layer, privacy-preserving primitive layer, secure vertical federated algorithm layer, vertical federated system layer, and application layer.}
\vspace{-0.2cm}
\label{fig:whole_design}
\end{figure*}

Our proposed layer approach is shown in Fig.~\ref{fig:whole_design}. In the following sections, we introduce the details of each VFL layer sequentially. We separately provide promising further directions (D) to guide the development of VFL:




\begin{itemize}
	\item Hardware layer: {\bf (D1) Hardware acceleration for cryptography-based VFL}
	\item Privacy-preserving primitive layer: {\bf (D2) Advanced and tailored primitive}
	\item Secure vertical federated algorithm layer: {\bf (D3) Secure and efficient entity alignment algorithm; (D4) Compiler of vertical federated machine learning algorithm; (D5) Hybrid privacy-preserving protocol}
	\item Vertical federated system layer: {\bf (D6) Fair incentive mechanism; (D7) Reliable verification} 
\end{itemize}

\section{Layer One: Hardware Design of VFL}
\label{sec:hardware_design}
The hardware layer is the most fundamental part of VFL. The whole system builds upon the hardware. Therefore, hardware design is highly related to the efficiency and security of VFL. The design of the hardware layer can be divided into hardware-based acceleration and hardware-assisted trusted execution environment.

\subsection{Hardware-Based Acceleration}
The design of hardware-based acceleration has been becoming more and more critical in VFL because the adopted privacy-preserving methods could result in significant computation and communication costs. The existing researches utilize FPGA~\cite{yang2020fpga}, GPU~\cite{cheng2021haflo}, and RDMA~\cite{liu2020accelerating} to achieve acceleration. In this subsection, we discuss the corresponding related works respectively.

\subsubsection{FPGA-Based Acceleration.}
FPGA (field-programmable gate array) is a semiconductor-integrated circuit configured for any task after manufacturing~\cite{kuon2008fpga}. Therefore, it could be customized to accelerate major workloads and enable designers to adapt to changing requirements. FPGA has been applied in traditional machine learning~\cite{zhang2015optimizing}, cryptography~\cite{san2016efficient},~\textit{etc}. Recently, FPGA is also adopted to accelerate VFL. For example, Yang~\textit{et al.} proposed an FPGA-based homomorphic encryption framework to expedite the training phase~\cite{yang2020fpga}. They carefully offload the modular multiplication operation,~\ie, the core operation of Paillier~\cite{paillier1999public}, to FPGA. In addition, they designed a compact architecture of the Paillier cryptosystem to incorporate the FPGA framework into VFL. Furthermore, FLASH~\cite{zhang2023flash} provided a more delicate design for Paillier-based VFL, which supports more prosperous cryptographic operations. Besides, FLASH proposed novel inter- \& intra-engine pipelining and dataflow scheduling mechanisms, thus achieving superior performance. 


\subsubsection{GPU-Based Acceleration.}
GPU (graphics processing unit) is an electronic circuit designed initially to power traditional graphics applications~\cite{owens2008gpu}. However, due to its highly parallel structure, GPU is primarily utilized to accelerate the training process of machine learning models, especially deep neural networks~\cite{schlegel2015deep}. In the area of VFL, Cheng~\etal proposed a GPU-based solution of acceleration for the vertical federated logistic regression algorithm named HAFLO~\cite{cheng2021haflo}. HAFLO summarizes and optimizes the core homomorphic operators to reduce the overhead that Paillier has introduced to VFL. In addition, IO and storage are also optimized by HAFLO for the vertical federated logistic regression algorithm. In practice, besides federated logistic regression, there are a large number of federated algorithms that need a delicately-designed GPU acceleration. Besides, for VFL based on fully homomorphic encryption (FHE)~\cite{gentry2009fully},~\eg,~\cite{jin2022towards}, acceleration could refer to~\cite{zhang2022sok}. In addition to FPGA and GPU,~\cite{zhang2022sok} also discusses FHE acceleration based on Application-specific Integrated Circuit (ASIC)~\cite{einspruch2012application}.


\subsubsection{RDMA-Based Acceleration.}
RDMA (remote direct memory access) is a kind of direct memory access technique that allows for remotely accessing the memory of a machine from another one without CPU intervention~\cite{kalia2016design}. RDMA is commonly utilized to accelerate applications intra-datacenter~\cite{guo2016rdma}. Recently, Liu proposed accelerating intra-party communication in VFL with RDMA~\cite{liu2020accelerating}. As model complexity and data volume increase, each participant of VFL may choose to adopt a distributed framework with multiple servers locally in a datacenter. In this situation, the intra-party communication cost could be the bottleneck of the whole vertical federated training process. Therefore, this work transmits data with RDMA for intra-party communication, with no need for modifications to the original VFL applications, to improve the network efficiency. Besides RDMA, advanced congestion control methods~\cite{cardwell2017bbr} could also be utilized in VFL to optimize inter-party communication.

{\bf (D1) Hardware acceleration for cryptography-based VFL.} The cryptography primitives are common choices for preserving privacy in VFL, such as Paillier or secret sharing. However, they could result in extensive computation and communication overheads, which severely hinder the development of VFL. Therefore, the design of hardware acceleration,~\eg, FPGA or RMDA, is promising with urgent need.


\subsection{Hardware-Assisted Trusted Execution Environment.}
Existing works have leveraged hardware-assisted trusted execution environments (TEE) to protect privacy in federated learning~\cite{mo2021ppfl,zhang2020enabling}. TEE creates a secure area on the processor to isolate data and codes from the operating system, providing integrity and confidentiality guarantee.~\cite{mo2021ppfl} adopts the Intel Software Guard Extensions (SGX)~\cite{costan2016intel} to perform secure aggregation on the server, a typical TEE implementation. Besides security, TEE is also very efficient because it only involves cleartext computation, orders of magnitude faster than ciphertext computation.~\cite{zhang2020enabling} leverages SGX to check if the training process is executed as intended. SGX could preserve privacy while incurring a small computational overhead. However, the application of TEE to VFL is still under-explored. How to delicately solve the problems of limited memory and frequent page swapping in TEE remains an immense challenge.
 
        

\section{Layer Two: Privacy-Preserving Primitive Design of VFL}
\label{sec:primitive_design}
Various cryptographic primitives have been utilized to protect data privacy in different steps of VFL. They could be classified as cryptographic primitives and non-cryptographic primitives. This section briefly introduces their basic mathematical principles and application areas to provide a comprehensive understanding.

\subsection{Cryptographic Primitive}

Cryptographic primitives used in VFL contain homomorphic encryption~\cite{yang2019federated,khodaparast2018privacy,feng2019securegbm,cheng2021secureboost,liang2004privacy,cristofaro2010practical}, oblivious transfer~\cite{ren2022improving}, garbled circuit~\cite{gascon2017privacy}, and secret sharing~\cite{wu2020privacy,chen2021homomorphic,huang2022cheetah}. We introduce them roughly in the order of dependency.


\subsubsection{Homomorphic Encryption (HE)} allows computing directly on the encrypted data without decrypting it~\cite{acar2018survey}. The homomorphic encryption conducts like: $[[ x_0 \star x_1 ]] = [[ x_0 ]] \ast [[ x_1 ]]$, where $x_0$ and $x_1$ could be all possible messages, $[[ \cdot ]]$ stands for encryption, and $\star$ and $\ast$ could be any operation,~\eg, addition or multiplication. Furthermore, HE could be divided into partially homomorphic encryption (PHE) and fully homomorphic encryption (FHE). PHE is only homomorphic over limited operations, while FHE is homomorphic over all operations. In addition, PHE and FHE perform differently in the view of efficiency. PHE is often much faster than FHE. Therefore, PHE is a common choice in the task of VFL. Two conmmonly utilized PHE methods are Parillier~\cite{paillier1999public} and Rivest–Shamir–Adleman (RSA)~\cite{rivest2019method}, whose security is guaranteed by the hardness of factoring two large prime numbers' product~\cite{montgomery1994survey}. Paillier is commonly utilized in the training process of VFL to protect the intermediate results from leakage~\cite{yang2019federated,khodaparast2018privacy,feng2019securegbm,cheng2021secureboost}.
%
Paillier is called additive PHE because we could obtain the plaintexts' addition result by multiplying the corresponding ciphertexts.
Besides, Paillier also supports the operation between a ciphertext and a plaintext.
These properties make Paillier popular in VFL because machine learning contains a lot of matrix addition and multiplication operations. Besides, RSA is a multiplicative PHE algorithm commonly used in the entity alignment process of VFL to protect the privacy of entities outside the intersection~\cite{liang2004privacy,cristofaro2010practical}. RSA algorithm is often used to sign a message to verify the sender of the message. Furthermore, the RSA-based entity alignment process also benefits from the blind signature ability of RSA. The techniques of FHE have been developing rapidly in recent years, such as GSW~\cite{gentry2013homomorphic} and CKKS~\cite{cheon2017homomorphic}. In terms of the efficiency of FHE, there is still a gap away from practical use. Naively utilizing FHE will result in too many computation and communication overheads. However,~\cite{sav2021poseidon,wei2021privacy} also delicately apply FHE in FL and achieve better efficiency than PHE.
 

%
%

\subsubsection{Oblivious Transfer (OT)} is a cryptographic protocol where a party sends one of his several messages to the other party but remains oblivious about which message has been transferred~\cite{yadav2021survey}. OT has been developed for decades and utilized in many applications,~\eg, private information retrieval~\cite{yekhanin2010private}. In the 1-out-of-2 OT, which can be implemented based on RSA, Alice owns two messages $x_0$ and $x_1$. Bob wants one but does not want Alice to know which one he receives. Besides, Bob should not see any information of $x_{1 - i}$. 1-out-of-2 OT could be extended to 1-out-of-n OT, which is more useful in real-world tasks. For example, 1-out-of-n OT could be utilized to protect privacy in the inference phase of VFL~\cite{ren2022improving}.



\subsubsection{Garbled Circuit (GC)}  is a cryptographic protocol that allows two mistrusting parties to conduct secure computation via a Boolean circuit~\cite{goldreich2003cryptography}. GC only reveals the output of the calculation and maintains the input and intermediate results safe. GC is first introduced in Yao's Millionaires' Problem~\cite{yao1986generate}, where two millionaires want to know who is richer between them but do not expose their actual wealth. OT can be utilized to implement the corresponding GC protocol, where the goal function (\eg, comparison function) is described as a Boolean circuit and known by both parties. GC is commonly used in VFL because of its high efficiency and security~\cite{gascon2017privacy}.

 

\subsubsection{Secret Sharing (SS)} is splitting a secret into multiple shares among parties~\cite{beimel2011secret}. Each party cannot recover the secret using only his own share. However, the secret can be reconstructed with all (sufficient) shares. Therefore, SS schemes are important for storing sensitive information,~\eg, missile launch codes and encryption keys. In addition, SS benefits the VFL tasks because of its homomorphic property. We explain it with the arithmetic sharing of ABY~\cite{demmler2015aby} as an example, which is a secure two-party computation framework. Based on additively sharing private values, x is split as an $l$-bit $\langle x \rangle \equiv \langle x \rangle_0 + \langle x \rangle_1$, where $\langle x \rangle_0, \langle x \rangle_1 \in \mathbb{Z}_{2^l}$. For sharing secret $x$, party $i$ chooses a random number $r \in_R \mathbb{Z}_{2^l}$, sets $\langle x \rangle_i = x - r$ and sends $r$ to party $1 - i$ as $\langle x \rangle_{1 - i}$. For reconstructing secret $x$, party $1 - i$ sends its share $\langle x \rangle_{1 - i}$ to party $i$ to compute $x = \langle x \rangle_0 + \langle x \rangle_1$. In such a sharing scheme, the addition operation $\langle z \rangle = \langle x \rangle + \langle y \rangle$ is easy via local computation: $
	\langle z \rangle_i = \langle x \rangle_i + \langle y \rangle_i$, while the multiplication operation is much harder because of the interactive terms between parties. However, it could be solved through a multiplication triple $\langle c \rangle = \langle a \rangle \langle b \rangle$. First, each party $i$ calculates $\langle e \rangle_i = \langle x \rangle_i - \langle a \rangle_i$ and $\langle f \rangle_i = \langle y \rangle_i - \langle b \rangle_i$. Then, both parties construct $e$ and $f$. Next, each party $i$ could compute $\langle z \rangle_i$ as: $\langle z \rangle_i = i e f + f \langle a \rangle_i + e \langle b \rangle_i + \langle c \rangle_i$. The triples could be pre-computed in the offline stage via HE or OT. Besides, SS could also be implemented via garbled circuit~\cite{demmler2015aby,mohassel2018aby3}. SS is a common building block for SS-based VFL protocols~\cite{wu2020privacy,chen2021homomorphic,huang2022cheetah}. However, SS-based VFL methods suffer from large communication overhead.


\subsection{Non-Cryptographic Primitive}
There are many non-cryptographic primitives utilized in VFL. We mainly introduce differential privacy and random mask.

\subsubsection{Differential Privacy (DP)} is a statistical primitive for protecting individual information of a dataset~\cite{dwork2014algorithmic}. The main idea of DP is adding noises to only expose the statistical data of group information but limit the leakage of individual details. We explain the DP mechanism using a database example. A database $\mathcal{D}$ is a collection of records from a universe $\mathcal{X}$. A randomized algorithm $M$ with domain $\mathbb{N}^{|\mathcal{X}|}$ is $(\epsilon, \delta)$ - DP if: $\mbox{Pr}[M(\mathcal{D}_0) \in \mathcal{S}] \leq e^{\epsilon} \mbox{Pr}[M(\mathcal{D}_1) \in \mathcal{S}] + \delta$, for $\forall \mathcal{S} \subseteq Range(M)$ and two databases $\mathcal{D}_0$ and $\mathcal{D}_1$ only differ in one data record $|| \mathcal{D}_0 - \mathcal{D}_0 ||_1 \leq 1,\ \forall \mathcal{D}_0, \mathcal{D}_1 \in \mathbb{N}^{|\mathcal{X}|}$. Considering a database query scenario, $M$ could be regarded as adding noises to the query results to hide the existence of a single record. $\epsilon$ is the privacy budget. Generally, a small $\epsilon$ means a better privacy protection and worse data utility due to the large number of injected noises. The adoption of $\delta$ further relaxes the striction and improves the practicability of DP. 
Although the technique of DP has already been applied in VFL because of its efficiency~\cite{hu2019fdml,wang2020hybrid,xu2019achieving,peng2021differentially}, there exist hurdles. The principal reason is that VFL is often utilized in the cross-silo scenarios. On the one hand, DP leads to accuracy loss of the federated model, which should harm the business cooperation. On the other hand, data processed by DP is still plaintext, which may violate the related laws and regulations during communication among parties.

\subsubsection{Random Mask (RM)} is an intuitive primitive commonly used in VFL~\cite{yang2019federated,zhang2020additively}. Taking a Paillier-based two-party scenario as an example~\cite{yang2019federated}, we explain the mechanism of RM. Party A generates keys of Paillier, while party B contains $[[ x ]]$, which is the number $x$ encrypted by party A's public key. If party B wants to obtain plaintext $x$, he can send $[[ x ]]$ to party A for decryption. However, party B does not wish to reveal $x$ to party A. Therefore, party B randomly generates an additional mask $r$, computes $[[ x ]] + r $, and sends it to party A. Party A decrypts the received message and get $x + r$. Because of the random mask $r$, party A cannot know $x$. Finally, party A sends back the masked data $x + r$ to party B to remove the mask. Similarly, multiplicative RM is utilized in the vertical federated neural network to protect the privacy of high-order feature embeddings~\cite{zhang2020additively}. Besides,~\cite{chai2021federated} designed a novel removable RM technique for privacy-preserving singular vector decomposition. Moreover, RM can also be used to design secure aggregation protocols~\cite{bonawitz2017practical}. Without encryption overhead, RM is an efficient and flexible method for preserving privacy in VFL.

{\bf (D2) Advanced and tailored primitive.} On the one hand, researchers of VFL should keep up with the latest developments in privacy-preserving primitives. For example, although less adopted in VFL, FHE is proven more efficient than Paillier with delicate designs in some recent works~\cite{sav2021poseidon}. On the other hand, we could tailor the primitive for better performance in VFL, similar to~\cite{jiang2021flashe}, which re-designed Paillier for the secure aggregation and key distribution operations of HFL.
\vspace{-0.2cm}


\section{Layer Three: Secure Vertical Federated Algorithm Design of VFL}
\label{sec:protocol_design}
In this section, we explain how to utilize the aforementioned privacy-preserving primitives to design secure vertical federated algorithms for various requirements of different VFL tasks. According to different phases of VFL, we divide the secure vertical federated algorithms to secure entity alignment algorithms and secure vertical federated machine learning algorithms.

\vspace{-0.2cm}

\subsection{Secure Entity Alignment Algorithm}
Secure entity alignment protocols are only contained in the VFL scenarios but not in the HFL scenarios. They securely match the same entities of the samples in different parties. Then the aligned entities can form a virtual dataset for the down-stream vertical training process. According to the existence of a unique identifier,~\eg, sample ID, secure entity alignment protocols could be categorized into private set intersection protocols and privacy-preserving record linkage protocols.

\subsubsection{Private Set Intersection (PSI) Algorithm}

is the more common choice for current VFL tasks, whose input is the unique identifier of different parties. Considering privacy, many intuitive protocols,~\eg, the hash-based entity alignment protocol, are not safe enough. In the hash-based protocol, one party computes the hash values of their sample IDs and sends them to the other parties. Sample membership can be revealed under the rainbow table attack~\footnote{https://en.wikipedia.org/wiki/Rainbow\_table}. The two-party RSA-based protocol~\cite{liang2004privacy} is the commonly used PSI protocol, utilizing the blind signature operation. It can further hide the sample IDs outside the intersection but only leaks the privacy of IDs in the intersection.
%
Based on this standard solution, several works designed OT-based PSI methods for more efficiency~\cite{pinkas2014faster, pinkas2018scalable}.~\cite{lu2020multi} investigated the PSI protocols which support the dropout of participants. Zhou~\etal extended the number of parties in PSI from 2 to $n$~\cite{zhou2021privacy}. In addition, some pilot works also try to conduct PSI while further protecting the privacy of IDs in the intersection~\cite{luo2021secure}. Besides,~\cite{pinkas2019efficient} provides a circuit-based PSI to achieve linear communication complexity.

\subsubsection{Privacy-Preserving Record Linkage (PPRL) Algorithm} is suitable for the VFL tasks where quasi-features are utilized to conduct entity alignment due to the absence of the unique identifier~\cite{vatsalan2017privacy}. For example, GPS location and address cannot be directly used as the unique identifier in the entity alignment process. Therefore, we must calculate the similarities and choose the most similar entities for matching~\cite{nock2021impact}. PPRL is a continuously studied topic in the field of database. Besides, embedding-based PPRL protocols are also explored in the domain of knowledge graph in recent years. In addition, a similarity learning algorithm exists for PPRL~\cite{lee2018privacy}. However, PPRL is relatively less commonly used in VFL compared with PSI. And there exist various attacks~\cite{vidanage2020graph} on current PPRL methods. Besides the PSI and PPRL protocols,~\cite{sun2021vertical} proposed to utilize private set union (PSU) protocol for protecting the intersection membership. However, the generated synthetic features may harm the effectiveness of down-stream VFL models. Worth to mention, the entity alignment process is also prone to make mistakes in reality~\cite{nock2021impact}, which could harm the effectiveness of the down-stream VFL training. Nevertheless, it is still an underexplored problem.

{\bf (D3) Secure and efficient entity alignment algorithm.} The existing entity alignment algorithms suffer from either security or efficiency. As the very first step of VFL, our final vision is to protect the privacy of all IDs as well as maintain good efficiency.

\vspace{-0.2cm}

\subsection{Secure Vertical Federated Machine Learning Algorithm}

THe secure vertical federated machine learning algorithm is the core component of VFL. However, the adopted secure protocols contain too many aspects and are always hard for us to analyze. In this subsection, we first introduce a novel taxonomy method involving four dimensions,~\ie, machine learning model (M), protection object (O), security model (S) and privacy-preserving protocol (P). Then, we discuss and compare the existing works according to the proposed categorization method.

\subsubsection{MOSP Tree Taxonomy Method.}
\begin{figure*}[h]
\centering
\includegraphics[width=0.8\linewidth]{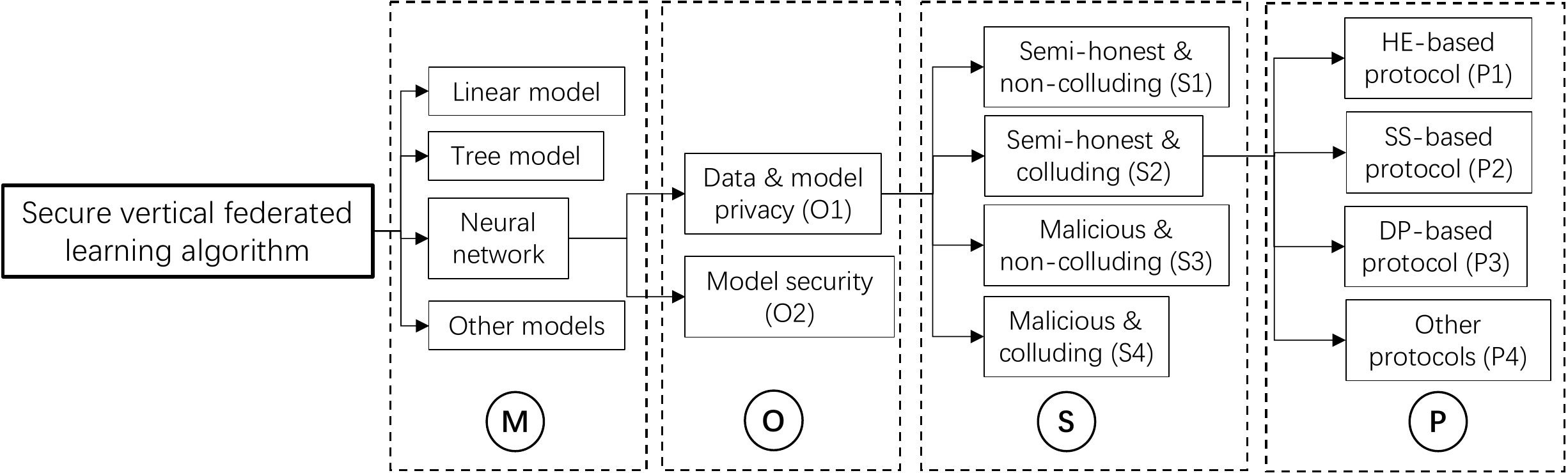}
\caption{The illustration of MOSP tree.}
\vspace{-0.4cm}
\label{fig:MOSP}
\end{figure*}

We design a novel taxonomy method,~\ie, MOSP tree, to analyze the existing secure VFL algorithms, shown in Fig.~\ref{fig:MOSP}. We consider four principal dimensions of all algorithms. Following the structure of MOSP tree, we can provide a fair and clear review of the related works.

\paragraph{\textbf{Machine Learning Model (M).}}
The first dimension to consider is the adopted machine learning model. As we have seen in recent years, various kinds of machine learning models have been utilized in the VFL scenario. We mainly summarize them as~\ul{\textit{linear model}},~\ul{\textit{tree model}},~\ul{\textit{neural network}}, and~\ul{\textit{other models}}.


\paragraph{\textbf{Protection Object (O).}}
Given a specific machine learning model, we could further categorize the secure VFL algorithms according to their protection objects,~\ie,~\ul{\textit{data~\& model privacy (O1)}} and~\ul{\textit{model security (O2)}}. The protection of the former one refers to preventing the leakage of private data and models, while the safety of the latter one refers to preventing model tampering. Taking an example to explain model security,~\cite{liu2020backdoor} designed backdoor attacks to harm the utility (accuracy) of federated models by modifying the transmitted gradients.



\paragraph{\textbf{Security Model (S).}}
Besides, we need to know adversaries' ability to choose of different privacy-preserving protocols. The definition of the security model commonly contains two aspects: 1) the willingness of adversaries to deviate from protocols,~\ie, semi-honest or malicious; 2) the collusion situation of adversaries,~\ie, colluding or non-colluding. Participants corrupted by semi-honest adversaries follow the protocols but try to breach other parties' privacy from the received messages. In contrast, participants corrupted by malicious adversaries directly deviate from protocols to cheat and harm both data~\& model privacy and model security. We do not consider the honest adversary because it does not exist in a real-world scenario. Colluding adversaries cooperate, while non-colluding adversaries do not. Therefore, we consider four kinds of security models,~\ie,~\ul{\textit{semi-honest~\& non-colluding (S1)}},~\ul{\textit{semi-honest~\& colluding (S2)}},~\ul{\textit{malicious~\& non-colluding (S3)}}, and~\ul{\textit{malicious~\& colluding (S4)}}.



\paragraph{\textbf{Privacy-Preserving Protocol (P).}}
After the three dimensions are fixed, we could adopt appropriate protection methods to design the corresponding secure protocols. Protection methods are categorized into four types,~\ie,~\ul{\textit{HE-based protocol (P1)}}, \ul{\textit{SS-based protocol (P2)}}, \ul{\textit{DP-based protocol (P3)}}, and \ul{\textit{other protocols (P4)}}, according to the primarily adopted privacy-preserving primitives. For example, the main idea of SS-based protocols is splitting private data into secret shares and conducting computation over shares among parties. Other protocols could contain more efficient multi-party computation (MPC) techniques~\cite{cramer2015secure}. We summarize them and analyze their advantages and disadvantages in Tab.~\ref{tab:secure-ml-protocols}. 

 
\begin{table*}[h]
\centering
\footnotesize
\begin{tabular}{ccccc}
	\toprule
	 & HE-based protocols & SS-based protocols & DP-based protocols & Other protocols \\
	\midrule
	\makecell[c]{Utilized\\primitives} & HE combined with RM, SS & SS combined with GC, HE, OT & DP & RM, SS or others \\
	\midrule
	Pros and cons &~\makecell[c]{Protect privacy during\\training, mainly introduce extra\\computational costs and often\\expose plaintext intermediate results\\for better efficiency} &~\makecell[c]{Protect privacy during\\training, mainly introduce extra\\communicational costs and hide \\all intermediate results for\\end-to-end security} &~\makecell[c]{Protect the\\privacy of model\\prediction but cause\\accuracy loss} & ~\makecell[c]{Very efficient but may\\contain severe privacy\\risks} \\
	\bottomrule
\end{tabular}
\caption{Summary of privacy-preserving protocols studied in vertical federated learning.}
\vspace{-0.8cm}
\label{tab:secure-ml-protocols}
\end{table*}

%

\subsubsection{Categorization of Existing Secure Vertical Federated Machine Learning Algorithms.}
\label{subsubsec:categorization}
In the following part of this section, we separately categorize the existing secure VFL algorithms,~\ie, vertical federated linear algorithm in Tab.~\ref{tab:summary_vertical_linear}, vertical federated tree algorithm in Tab.~\ref{tab:summary_vertical_tree}, vertical federated neural network in Tab.~\ref{tab:summary_vertical_nn} and other vertical federated algorithms in Tab.~\ref{tab:summary_vertical_other} according to the proposed MOSP tree taxonomy method.

\parab{Vertical Federated Linear Algorithm}

\begin{table*}[h]
\centering
\footnotesize
\begin{tabularx}{\linewidth}{ccc|cccc|ccccl}
\toprule
\multirow{2}{*}{\bf Algorithm}
& \multicolumn{2}{c}{\makecell[c]{{\bf Protection} \\ {\bf objective (O)}}}
& \multicolumn{4}{c}{\bf Security model (S)}
& \multicolumn{4}{c}{\makecell[c]{{\bf Privacy-preserving} \\ {\bf protocol (P)}}} 
& \multirow{2}{*}{\bf Main innovation and comments} \\

& O1 & O2 & S1 & S2 & S3 & S4 & P1 & P2 & P3 & P4 & \\
\midrule

ABY, 2015~\cite{demmler2015aby} & \fullcirc[0.6ex] & \emptycirc[0.6ex] & \fullcirc[0.6ex] & \emptycirc[0.6ex] & \emptycirc[0.6ex] & \emptycirc[0.6ex] & \emptycirc[0.6ex] & \fullcirc[0.6ex] & \emptycirc[0.6ex] & \emptycirc[0.6ex] & Pioneering SS-based framework \\
\midrule
VERTIGO, 2016~\cite{li2016vertical} & \fullcirc[0.6ex] & \emptycirc[0.6ex] & \fullcirc[0.6ex] & \emptycirc[0.6ex] & \emptycirc[0.6ex] & \emptycirc[0.6ex] & \emptycirc[0.6ex] & \emptycirc[0.6ex] & \emptycirc[0.6ex] & \fullcirc[0.6ex] & Utilizing linear separation property of kernel matrix \\
\midrule
SecureML, 2017~\cite{mohassel2017secureml} & \fullcirc[0.6ex] & \emptycirc[0.6ex] & \fullcirc[0.6ex] & \emptycirc[0.6ex] & \emptycirc[0.6ex] & \emptycirc[0.6ex] & \emptycirc[0.6ex] & \fullcirc[0.6ex] & \emptycirc[0.6ex] & \emptycirc[0.6ex] & Support of secure arithmetic operations \\

Gasc{\'o}n~\etal, 2017~\cite{gascon2017privacy} & \fullcirc[0.6ex] & \emptycirc[0.6ex] & \fullcirc[0.6ex] & \emptycirc[0.6ex] & \emptycirc[0.6ex] & \emptycirc[0.6ex] & \emptycirc[0.6ex] & \fullcirc[0.6ex] & \emptycirc[0.6ex] & \emptycirc[0.6ex] & Hybrid protocol with SS and GC \\
Nathan~\etal, 2017~\cite{nathan2017optimization} & \fullcirc[0.6ex] & \emptycirc[0.6ex] & \fullcirc[0.6ex] & \emptycirc[0.6ex] & \emptycirc[0.6ex] & \emptycirc[0.6ex] & \emptycirc[0.6ex] & \emptycirc[0.6ex] & \emptycirc[0.6ex] & \fullcirc[0.6ex] & First-order dual optimization method \\
\midrule
ABY$^3$, 2018~\cite{mohassel2018aby3} & \fullcirc[0.6ex] & \emptycirc[0.6ex] & \fullcirc[0.6ex] & \emptycirc[0.6ex] & \fullcirc[0.6ex] & \emptycirc[0.6ex] & \emptycirc[0.6ex] & \fullcirc[0.6ex] & \emptycirc[0.6ex] & \emptycirc[0.6ex] & Extension of ABY to three parties \\
\midrule
Yang~\etal, 2019~\cite{yang2019federated} & \fullcirc[0.6ex] & \emptycirc[0.6ex] & \fullcirc[0.6ex] & \emptycirc[0.6ex] & \emptycirc[0.6ex] & \emptycirc[0.6ex] & \fullcirc[0.6ex] & \emptycirc[0.6ex] & \emptycirc[0.6ex] & \halfcirc[0.6ex] & Pioneering HE-based algorithm \\

Yang~\etal, 2019~\cite{yang2019parallel} & \fullcirc[0.6ex] & \emptycirc[0.6ex] & \fullcirc[0.6ex] & \emptycirc[0.6ex] & \emptycirc[0.6ex] & \emptycirc[0.6ex] & \fullcirc[0.6ex] & \emptycirc[0.6ex] & \emptycirc[0.6ex] & \emptycirc[0.6ex] & Third-party coordinator removal \\

Yang~\etal, 2019~\cite{yang2019quasi} & \fullcirc[0.6ex] & \emptycirc[0.6ex] & \fullcirc[0.6ex] & \emptycirc[0.6ex] & \emptycirc[0.6ex] & \emptycirc[0.6ex] & \fullcirc[0.6ex] & \emptycirc[0.6ex] & \emptycirc[0.6ex] & \emptycirc[0.6ex] & Incorporating second-order information \\

Xu~\etal, 2019~\cite{xu2019achieving} & \fullcirc[0.6ex] & \emptycirc[0.6ex] & \fullcirc[0.6ex] & \emptycirc[0.6ex] & \emptycirc[0.6ex] & \emptycirc[0.6ex] & \fullcirc[0.6ex] & \emptycirc[0.6ex] & \fullcirc[0.6ex] & \emptycirc[0.6ex] & Pioneering DP-based algorithm \\

FDML, 2019~\cite{hu2019fdml} & \fullcirc[0.6ex] & \emptycirc[0.6ex] & \fullcirc[0.6ex] & \emptycirc[0.6ex] & \emptycirc[0.6ex] & \emptycirc[0.6ex] & \emptycirc[0.6ex] & \emptycirc[0.6ex] & \halfcirc[0.6ex] & \fullcirc[0.6ex] & Idea of ensemble learning \\

FedBCD~\cite{liu2019communication} & \fullcirc[0.6ex] & \emptycirc[0.6ex] & \fullcirc[0.6ex] & \emptycirc[0.6ex] & \emptycirc[0.6ex] & \emptycirc[0.6ex] & \fullcirc[0.6ex] & \emptycirc[0.6ex] & \emptycirc[0.6ex] & \emptycirc[0.6ex] & Multiple local updates in each iteration \\
\midrule
HDP-VFL, 2020~\cite{wang2020hybrid} & \fullcirc[0.6ex] & \emptycirc[0.6ex] & \fullcirc[0.6ex] & \emptycirc[0.6ex] & \emptycirc[0.6ex] & \emptycirc[0.6ex] & \emptycirc[0.6ex] & \emptycirc[0.6ex] & \fullcirc[0.6ex] & \emptycirc[0.6ex] & Analysis of privacy leakage of intermediate result \\
GP-AVFL, 2020~\cite{li2020efficient} & \fullcirc[0.6ex] & \emptycirc[0.6ex] & \fullcirc[0.6ex] & \emptycirc[0.6ex] & \emptycirc[0.6ex] & \emptycirc[0.6ex] & \fullcirc[0.6ex] & \emptycirc[0.6ex] & \emptycirc[0.6ex] & \emptycirc[0.6ex] & Gradient prediction~\& sparse compression \\
\midrule
Yang~\etal, 2021~\cite{yang2021model} & \fullcirc[0.6ex] & \emptycirc[0.6ex] & \fullcirc[0.6ex] & \emptycirc[0.6ex] & \emptycirc[0.6ex] & \emptycirc[0.6ex] & \fullcirc[0.6ex] & \emptycirc[0.6ex] & \emptycirc[0.6ex] & \emptycirc[0.6ex] & Gradient compression \\

Wei~\etal, 2021~\cite{wei2021privacy} & \fullcirc[0.6ex] & \emptycirc[0.6ex] & \fullcirc[0.6ex] & \emptycirc[0.6ex] & \emptycirc[0.6ex] & \emptycirc[0.6ex] & \fullcirc[0.6ex] & \emptycirc[0.6ex] & \emptycirc[0.6ex] & \emptycirc[0.6ex] & Asynchronous gradient sharing and CKKS \\

CAESAR, 2021~\cite{chen2021homomorphic} & \fullcirc[0.6ex] & \emptycirc[0.6ex] & \fullcirc[0.6ex] & \emptycirc[0.6ex] & \emptycirc[0.6ex] & \emptycirc[0.6ex] & \halfcirc[0.6ex] & \fullcirc[0.6ex] & \emptycirc[0.6ex] & \emptycirc[0.6ex] & Combining SS with HE to handle sparse features \\

Dalskov~\etal, 2021~\cite{dalskov2021fantastic} & \fullcirc[0.6ex] & \emptycirc[0.6ex] & \fullcirc[0.6ex] & \emptycirc[0.6ex] & \fullcirc[0.6ex] & \emptycirc[0.6ex] & \emptycirc[0.6ex] & \fullcirc[0.6ex] & \emptycirc[0.6ex] & \emptycirc[0.6ex] & Extension to four parties \\

VERTIGO-CI, 2021~\cite{kim2021vertical} & \fullcirc[0.6ex] & \emptycirc[0.6ex] & \fullcirc[0.6ex] & \emptycirc[0.6ex] & \emptycirc[0.6ex] & \emptycirc[0.6ex] & \emptycirc[0.6ex] & \emptycirc[0.6ex] & \emptycirc[0.6ex] & \fullcirc[0.6ex] & Further handling variance estimation \\

AsySQN, 2021~\cite{zhang2021asysqn} & \fullcirc[0.6ex] & \emptycirc[0.6ex] & \fullcirc[0.6ex] & \fullcirc[0.6ex] & \emptycirc[0.6ex] & \emptycirc[0.6ex] & \emptycirc[0.6ex] & \emptycirc[0.6ex] & \emptycirc[0.6ex] & \fullcirc[0.6ex] & Asynchronous computation \\

VF{\bf B}$^2$, 2021~\cite{zhang2021secure} & \fullcirc[0.6ex] & \emptycirc[0.6ex] & \fullcirc[0.6ex] & \fullcirc[0.6ex] & \emptycirc[0.6ex] & \emptycirc[0.6ex] & \emptycirc[0.6ex] & \emptycirc[0.6ex] & \emptycirc[0.6ex] & \fullcirc[0.6ex] & Secure aggregation with RM \\

AsyREVEL, 2021~\cite{zhang2021desirable} & \fullcirc[0.6ex] & \fullcirc[0.6ex] & \fullcirc[0.6ex] & \fullcirc[0.6ex] & \fullcirc[0.6ex] & \emptycirc[0.6ex] & \emptycirc[0.6ex] & \emptycirc[0.6ex] & \emptycirc[0.6ex] & \fullcirc[0.6ex] & Utilizing zeroth-order optimization \\
AFSGD-VP, 2021~\cite{gu2021privacy} & \fullcirc[0.6ex] & \emptycirc[0.6ex] & \fullcirc[0.6ex] & \emptycirc[0.6ex] & \emptycirc[0.6ex] & \emptycirc[0.6ex] & \emptycirc[0.6ex] & \emptycirc[0.6ex] & \emptycirc[0.6ex] & \fullcirc[0.6ex] & Tree-structured communication scheme \\
\midrule
Shi~\etal, 2022~\cite{sun2022privacy,shi2022mvfls} & \fullcirc[0.6ex] & \emptycirc[0.6ex] & \fullcirc[0.6ex] & \emptycirc[0.6ex] & \emptycirc[0.6ex] & \emptycirc[0.6ex] & \emptycirc[0.6ex] & \emptycirc[0.6ex] & \emptycirc[0.6ex] & \fullcirc[0.6ex] & Adoption of MPC-based secure aggregation protocol \\
ACCEL, 2022~\cite{zhao2022accel} & \fullcirc[0.6ex] & \emptycirc[0.6ex] & \fullcirc[0.6ex] & \emptycirc[0.6ex] & \emptycirc[0.6ex] & \emptycirc[0.6ex] & \fullcirc[0.6ex] & \emptycirc[0.6ex] & \emptycirc[0.6ex] & \fullcirc[0.6ex] & Adoption of symmetric homomorphic encryption \\
Zhang~\etal, 2022~\cite{zhang2022adaptive} & \fullcirc[0.6ex] & \emptycirc[0.6ex] & \fullcirc[0.6ex] & \emptycirc[0.6ex] & \emptycirc[0.6ex] & \emptycirc[0.6ex] & \emptycirc[0.6ex] & \emptycirc[0.6ex] & \emptycirc[0.6ex] & \fullcirc[0.6ex] & Impact of unbalanced features on convergence \\
Cai~\etal, 2022~\cite{cai2022efficient} & \fullcirc[0.6ex] & \emptycirc[0.6ex] & \fullcirc[0.6ex] & \fullcirc[0.6ex] & \emptycirc[0.6ex] & \fullcirc[0.6ex] & \halfcirc[0.6ex] & \emptycirc[0.6ex] & \emptycirc[0.6ex] & \fullcirc[0.6ex] & RM-based method with theoretical guarantee \\
$P^2$VCLR, 2022~\cite{yu2022privacy} & \fullcirc[0.6ex] & \emptycirc[0.6ex] & \fullcirc[0.6ex] & \emptycirc[0.6ex] & \emptycirc[0.6ex] & \emptycirc[0.6ex] & \fullcirc[0.6ex] & \emptycirc[0.6ex] & \emptycirc[0.6ex] & \emptycirc[0.6ex] & Encrypting the original data with CKKS \\
EFMVFL, 2022~\cite{huang2022efmvfl} & \fullcirc[0.6ex] & \emptycirc[0.6ex] & \fullcirc[0.6ex] & \emptycirc[0.6ex] & \emptycirc[0.6ex] & \emptycirc[0.6ex] & \fullcirc[0.6ex] & \fullcirc[0.6ex] & \emptycirc[0.6ex] & \emptycirc[0.6ex] & Hybrid protocol of HE and SS without a third party \\
Yuan~\etal, 2022~\cite{yuan2022byzantine} & \fullcirc[0.6ex] & \emptycirc[0.6ex] & \fullcirc[0.6ex] & \emptycirc[0.6ex] & \emptycirc[0.6ex] & \emptycirc[0.6ex] & \emptycirc[0.6ex] & \emptycirc[0.6ex] & \emptycirc[0.6ex] & \fullcirc[0.6ex] & Byzantine-resilient method \\
Dai~\etal, 2022~\cite{dai2022edge} & \fullcirc[0.6ex] & \emptycirc[0.6ex] & \fullcirc[0.6ex] & \emptycirc[0.6ex] & \emptycirc[0.6ex] & \emptycirc[0.6ex] & \fullcirc[0.6ex] & \emptycirc[0.6ex] & \emptycirc[0.6ex] & \emptycirc[0.6ex] & Distributed computing in each party \\
\midrule
VFLR, 2023~\cite{zhao2023vflr} & \fullcirc[0.6ex] & \emptycirc[0.6ex] & \fullcirc[0.6ex] & \emptycirc[0.6ex] & \emptycirc[0.6ex] & \emptycirc[0.6ex] & \fullcirc[0.6ex] & \emptycirc[0.6ex] & \emptycirc[0.6ex] & \fullcirc[0.6ex] & Secure vertical federated query \\
\bottomrule
\end{tabularx}
\caption{Summary of the existing secure vertical federated linear algorithms. The black dot represents ``yes'', the white dot stands for ``no'' and the half dot means ``taking a small part''.}
\vspace{-0.8cm}
\label{tab:summary_vertical_linear}
\end{table*}

Linear models take the general form of $f(\mathbf{X}|\bm{\theta}) = \bm{\theta}^T\mathbf{X}$. Due to their simplicity, they are widely used in practice to solve problems with large feature dimensions~\cite{maulud2020review}. Therefore, linear models are among the first to be studied in VFL, which we categorize in Tab.~\ref{tab:summary_vertical_linear}. 
 
\paragraph{\textbf{HE-Based Vertical Federated Linear Algorithm.}}

A typical HE-based vertical federated linear algorithm with~\ul{\textit{S1}} security model is proposed by~\cite{yang2019federated}, aiming to protect~\ul{\textit{O1}}.
 In this scenario, party A owns dataset $\mathcal{D}_A = \{ \bm{X}_A \in \mathbb{R}^{N \times d_A}, \bm{y} \in \mathbb{R}^{N \times 1} \}$, while party B owns dataset $\mathcal{D}_B = \{ \bm{X}_B \in \mathbb{R}^{N \times d_B} \}$. We assume the two datasets have already been aligned by the aforementioned secure entity alignment algorithm. The two parties want to collaboratively train models as: $\mathop{\arg\min}_{\bm{w}_A, \bm{w}_B} \sum_i \| y_i - \bm{w}^T_A \bm{x}_{A,i} - \bm{w}^T_B \bm{x}_{B,i} \|^2$, where $\bm{w}_A \in \mathbb{R}^{d_A \times 1}, \bm{w}_B \in \mathbb{R}^{d_B \times 1}$ are model weights and $y_i, \bm{x}_{A,i}, \bm{x}_{B,i}$ represent the $i$th training sample. 
The main idea of this algorithm is to encrypt the intermediate results before transmitting them and conducting cryptographic computation over the received ciphertext. The protocol is designed as: 1) coordinator C creates the key pair of HE and sends the public key pair to both party A and B; 2) party B computes $[[ \bm{w}^T_B \bm{x}_{B,i} ]]$ and sends it to party A; 3) party A computes $[[ d_i ]] = y_i - \bm{w}^T_A \bm{x}_{A,i} - [[ \bm{w}^T_B \bm{x}_{B,i} ]]$, sends it to party B and computes encrypted masked gradients $[[ d_i ]] \bm{x}_{A, i} + \bm{r}_A$, sends it to coordinator C; 4) party A computes encrypted masked gradients $[[ d_i ]] \bm{x}_{B, i} + \bm{r}_B$ and sends it to coordinator C; 5) coordinator C decrypts the received messages as $d_i {\bm x}_{A, i} + \bm{r}_A$, $d_i \bm{x}_{B, i} + \bm{r}_B$ and sends them back; 6) two parties remove the masks $r_A$, $r_B$ and update the parameters, with gradients $d_i {\bm x}_{A, i}$, $d_i \bm{x}_{B, i}$ respectively.
Besides HE, this algorithm also utilizes RM $\bm{r}_A$, $\bm{r}_B$ to prevent gradients from disclosing to Coordinator C. Since the adoption of HE introduces computational and communicational overheads, many works focus on alleviating the efficiency problem.
For example,~\cite{yang2021model,wei2021privacy,yang2019parallel} construct similar VFL secure protocols based on HE for linear models. However, the partial local estimate is disclosed for better efficiency.
~\cite{liu2019communication} conducts multiple local updates before communication to reduce communication rounds to improve efficiency. Also aiming to reduce communication rounds,~\cite{yang2019quasi} adopted the quasi-Newton method for fast convergence.~\cite{li2020efficient} designed gradient prediction and double-end sparse compression for acceleration. Furthermore,~\cite{wei2021privacy,yu2022privacy} utilize CKKS instead of Paillier, a common choice in VFL, to train VFL logistic regression. Their designs result in better efficiency due to the single instruction multiple data (SIMD) property of CKKS.~\cite{wei2021privacy} chooses to encrypt the intermediate results, while~\cite{yu2022privacy} encrypts the original data directly. Besides, ACCEL~\cite{zhao2022accel} utilized symmetric homomorphic encryption (SHE)~\cite{mahdikhani2020achieving} to implement a data aggregation matrix construction protocol for vertical federated logistic regression to improve training efficiency. Based on ACCEL, VFLR~\cite{zhao2023vflr} designed a novel vertical federated query scheme for inference.~\cite{dai2022edge} incorporated distributed computing (DC) into vertical federated linear regression, where each party contains multiple workers. In more detail, the authors utilized coded DC~\cite{lee2017speeding}, which could obtain the expected results with the alive workers, to alleviate the straggler problem.

For the scenarios with~\ul{\textit{S2}} security model,~\cite{weng2020privacy} proposed the reverse multiplication attack, which could result in the leakage of private data. Corrupted by adversaries, party A colludes with coordinator C to successfully conduct the attack. However, the attack is affected by the batch size and learning rate. We could set large batch size and small learning for defense.

\vspace{-0.2cm}

\paragraph{\textbf{SS-Based Vertical Federated Linear Algorithm.}}

The main idea of SS-based algorithms is sharing model weights (or even the original data,~\etc) over different parties and conducting computations in the form of secret shares. Privacy is further protected because there is no leakage of the intermediate results. An intuitive method is to secretly sharing local original data to other parties, and all parties collaboratively train models on data shares~\cite{mohassel2017secureml,demmler2015aby,mohassel2018aby3}. However, this kind of method suffers from high communication overhead. Therefore, there exist many works which do not directly share original data but intermediate results instead. For example, Gasc{\'o}n~\etal combined SS with GC and OT to design a hybrid protocol for vertical linear regression models~\cite{gascon2017privacy}. They generate additive shares of intermediate results for linear regression model via a secure inner product protocol and adopt GC to solve the distributed equation. The intermediate results are not revealed to any party. Only the final model weight is shown to parties. However, there exists a strong assumption that two additional parties (\ie, crypto service provider and evaluator introduced by GC) do not collude with each other. Besides, Chen~\etal also chose to secretly share model parameters to reduce the high communication overhead caused by high-dimensional sparse data~\cite{chen2021homomorphic}. In detail, they design a secure sparse matrix multiplication protocol combining SS and HE. Huang~\etal also proposed an HE and SS hybrid protocol for generalized linear models. They secretly share the intermediate results instead of model weights for better efficiency and party extension. However, SS-based algorithms have slight accuracy loss because of the approximate of non-linear parts. For~\ul{\textit{S3}} security model,~\cite{mohassel2018aby3,dalskov2021fantastic} provide variants of building blocks that could defend against malicious adversaries.


\vspace{-0.2cm}

\paragraph{\textbf{DP-Based Vertical Federated Linear Algorithm.}}

Besides the HE- and SS-based protocols, DP-based protocols could further protect the privacy of federated prediction results~\cite{xu2019achieving,wang2020hybrid}. Many VFL attacks are conducted utilizing model predictions. For example, Luo~\etal designed the feature inference attack for vertical linear algorithms~\cite{luo2021feature}. The active party performs the attack and tries to infer the features owned by the passive party. However, the authors strongly assume that the active party knows the model parameters of the passive party, which is difficult to achieve in real-world scenarios. Furthermore, Jiang~\etal proposed a more effective attack based on the same setting~\cite{jiang2022comprehensive}, which relieves the assumption from the leakage of passive party's model parameters to the prior knowledge of a small set of auxiliary samples. DP could be used to defend against these attacks, but it also results in accuracy loss. DP could be a complementation for HE- and SS-based protocols. Based on the functional mechanism of DP,~\cite{xu2019achieving} divides the objective function into multiple parties and injects noises separately.~\cite{wang2020hybrid} adds noises to the intermediate results.



\vspace{-0.2cm}


\paragraph{\textbf{Vertical Federated Linear Algorithm with Other Protocols.}}

Besides the above three kinds of protocols, there exist many other methods for better efficiency while maintaining accuracy. For example,~\cite{li2016vertical,kim2021vertical} proposed to exchange the plain-text local gram matrix instead of the original data during the vertical federated logistic regression training process, which is claimed to be secure as long as there are enough local covariates. In addition, FDML utilizes an ensemble method to design the secure protocol, where each party conducts partial inference, and all the plaintext predictions are aggregated as the final results~\cite{hu2019fdml}. Worthing to note, the scheme of FMDL could also be applied to other VFL algorithms.~\cite{zhang2021asysqn,zhang2021secure,gu2021privacy} utilize RM to protect the plain-text intermediate results. The final value could be recovered by removing all masks. Similarly, Cai~\etal proposed an efficient vertical federated ridge regression algorithm over large-scale data based on RM~\cite{cai2022efficient}. They could directly use the masked data for computation and remove the masks to obtain the expected results. Moreover, Sun~\etal and Shi~\etal adopted the MPC-based secure aggregation protocol~\cite{bonawitz2017practical} to improve the efficiency of VFL~\cite{sun2022privacy,shi2022mvfls}, which can be easily extended to scenarios with more participants. The authors stated that this framework could also suit other more complicated machine learning models. Besides,~\cite{gu2021privacy} utilizes tree-structured communication schemes for better efficiency. In addition, Nathan~\etal propose first-order dual optimization methods for large-scale feature distributed optimization~\cite{nathan2017optimization}. First-order methods explicitly compute the first-order derivatives and perform iterative optimization. Due to their versatility, first-order methods are extensively studied in VFL.~\cite{zhang2021asysqn} incorporates approximate second-order information to reduce the number of communication rounds.~\cite{zhang2021desirable} utilizes the zeroth-order optimization to design a novel vertical linear algorithm.

For the scenarios with~\ul{\textit{S2}} security model,~\cite{zhang2021asysqn,zhang2021secure,zhang2021desirable} claimed  their algorithms are secure. Even though the RM could be removed by colluding,~\cite{zhang2021asysqn,zhang2021secure} indicated that exactly revealing the original data and model weights is still impossible.~\cite{zhang2021desirable} also claimed to be secure even for~\ul{\textit{S2}} security model.~\cite{cai2022efficient} also provided security analysis for~\ul{\textit{S1}},~\ul{\textit{S2}} and~\ul{\textit{S4}} assumption. For~\ul{\textit{O2}}, the backdoor attack in VFL could harm the utility of federated models via injecting triggers in gradients~\cite{liu2020backdoor}.~\cite{zhang2021desirable} proposed only to transmit the function values. Therefore, the corrupted parties could not access the intermediate gradients to conduct such attacks. Besides,~\cite{yuan2022byzantine} showed that Byzantine attack exists in VFL and proposed a dual subgradient method for the linear algorithm.

\underline{Asynchronization.} 
Although the commonly utilized synchronous optimization method ensures that VFL is lossless, it requires all parties to compute and communicate intermediate results for every iteration, which suffers from two drawbacks: 1) it incurs a high communication cost; 2) synchronization among parties is prone to stragglers. Asynchronous methods are proposed to break the synchronization barrier.~\cite{liu2019communication,wei2021privacy} suggested conducting multiple updates locally at each party before exchanging stale (\ie, several iterations behind) information. However, Li~\etal found this simple implementation of the asynchronized mechanism performs poorly in most cases~\cite{li2020efficient}. The reason is that the ``stale result problem'' is much more severe in VFL, whose participants have different computational power, data amount, and network environment. Therefore, this work is further designed to narrow the gap between the received and current intermediate results. FDML also proposed a similar intuitive asynchronized method~\cite{hu2019fdml}. And~\cite{gu2021privacy,zhang2021secure} extended FDML in two ways. First, they present asynchronous versions of variance reduction methods and theoretically analyze their convergence. Second, they do not require shared labels.~\cite{zhang2022adaptive} found that unbalanced features could further harm the convergence of asynchronous methods. The authors proposed a novel optimization method that adaptively chooses the number of local updates for each party.






\begin{table*}[h]
\centering
\footnotesize
\begin{tabularx}{\linewidth}{ccc|cccc|ccccl}
\toprule
\multirow{2}{*}{\bf Algorithm}
& \multicolumn{2}{c}{\makecell[c]{{\bf Protection} \\ {\bf objective (O)}}}
& \multicolumn{4}{c}{\bf Security model (S)}
& \multicolumn{4}{c}{\makecell[c]{{\bf Privacy-preserving} \\ {\bf protocol (P)}}} 
& \multirow{2}{*}{\bf Main innovation and comments} \\

& O1 & O2 & S1 & S2 & S3 & S4 & P1 & P2 & P3 & P4 & \\
\midrule
Vaidya~\etal, 2008~\cite{vaidya2008privacy} & \fullcirc[0.6ex] & \emptycirc[0.6ex] & \fullcirc[0.6ex] & \fullcirc[0.6ex] & \emptycirc[0.6ex] & \emptycirc[0.6ex] & \emptycirc[0.6ex] & \fullcirc[0.6ex] & \emptycirc[0.6ex] & \emptycirc[0.6ex] & Secure multiparty dot product \\
\midrule
Sheikhalishahi~\etal, 2017~\cite{sheikhalishahi2017privacy} & \fullcirc[0.6ex] & \emptycirc[0.6ex] & \fullcirc[0.6ex] & \emptycirc[0.6ex] & \emptycirc[0.6ex] & \emptycirc[0.6ex] & \emptycirc[0.6ex] & \emptycirc[0.6ex] & \emptycirc[0.6ex] & \fullcirc[0.6ex] & Secure numbers comparison protocol \\
\midrule
Khodaparast~\etal, 2018~\cite{khodaparast2018privacy} & \fullcirc[0.6ex] & \emptycirc[0.6ex] & \fullcirc[0.6ex] & \emptycirc[0.6ex] & \emptycirc[0.6ex] & \emptycirc[0.6ex] & \emptycirc[0.6ex] & \emptycirc[0.6ex] & \emptycirc[0.6ex] & \fullcirc[0.6ex] & Secure weighted average protocol \\
\midrule
SecureGBM, 2019,~\cite{feng2019securegbm} & \fullcirc[0.6ex] & \emptycirc[0.6ex] & \fullcirc[0.6ex] & \emptycirc[0.6ex] & \emptycirc[0.6ex] & \emptycirc[0.6ex] & \fullcirc[0.6ex] & \emptycirc[0.6ex] & \emptycirc[0.6ex] & \emptycirc[0.6ex] & Secure splits search with domain isolation \\
FDML, 2019~\cite{hu2019fdml} & \fullcirc[0.6ex] & \emptycirc[0.6ex] & \fullcirc[0.6ex] & \emptycirc[0.6ex] & \emptycirc[0.6ex] & \emptycirc[0.6ex] & \emptycirc[0.6ex] & \emptycirc[0.6ex] & \halfcirc[0.6ex] & \fullcirc[0.6ex] & Idea of ensemble learning \\
\midrule
Pivot, 2020~\cite{wu2020privacy} & \fullcirc[0.6ex] & \emptycirc[0.6ex] & \fullcirc[0.6ex] & \emptycirc[0.6ex] & \emptycirc[0.6ex] & \emptycirc[0.6ex] & \fullcirc[0.6ex] & \halfcirc[0.6ex] & \emptycirc[0.6ex] & \emptycirc[0.6ex] & No leakage of intermediate information \\
\midrule
SecureBoost, 2021~\cite{cheng2021secureboost} & \fullcirc[0.6ex] & \emptycirc[0.6ex] & \fullcirc[0.6ex] & \emptycirc[0.6ex] & \emptycirc[0.6ex] & \emptycirc[0.6ex] & \fullcirc[0.6ex] & \emptycirc[0.6ex] & \emptycirc[0.6ex] & \emptycirc[0.6ex] & Pioneering HE-based algorithm \\

VF2Boost, 2021~\cite{fu2021vf2boost} & \fullcirc[0.6ex] & \emptycirc[0.6ex] & \fullcirc[0.6ex] & \emptycirc[0.6ex] & \emptycirc[0.6ex] & \emptycirc[0.6ex] & \fullcirc[0.6ex] & \emptycirc[0.6ex] & \emptycirc[0.6ex] & \emptycirc[0.6ex] & System optimization,~\eg, re-scheduling \\

PIVODL, 2021~\cite{zhu2021pivodl} & \fullcirc[0.6ex] & \emptycirc[0.6ex] & \fullcirc[0.6ex] & \emptycirc[0.6ex] & \emptycirc[0.6ex] & \emptycirc[0.6ex] & \fullcirc[0.6ex] & \emptycirc[0.6ex] & \fullcirc[0.6ex] & \emptycirc[0.6ex] & New scenario with distributed labels \\

Fang~\etal, 2021~\cite{fang2021large} & \fullcirc[0.6ex] & \emptycirc[0.6ex] & \fullcirc[0.6ex] & \emptycirc[0.6ex] & \emptycirc[0.6ex] & \emptycirc[0.6ex] & \halfcirc[0.6ex] & \fullcirc[0.6ex] & \emptycirc[0.6ex] & \emptycirc[0.6ex] & HE-based secure permutation method \\

FedXGBoost, 2021~\cite{le2021fedxgboost} & \fullcirc[0.6ex] & \emptycirc[0.6ex] & \fullcirc[0.6ex] & \emptycirc[0.6ex] & \emptycirc[0.6ex] & \emptycirc[0.6ex] & \emptycirc[0.6ex] & \emptycirc[0.6ex] & \fullcirc[0.6ex] & \fullcirc[0.6ex] & Adoption of both DP and MPC \\
\midrule
CryptoBoost, 2022~\cite{jin2022towards} & \fullcirc[0.6ex] & \emptycirc[0.6ex] & \fullcirc[0.6ex] & \emptycirc[0.6ex] & \fullcirc[0.6ex] & \emptycirc[0.6ex] & \fullcirc[0.6ex] & \fullcirc[0.6ex] & \emptycirc[0.6ex] & \emptycirc[0.6ex] & Utilization of MHE  \\
FEVERLESS, 2022 ~\cite{wang2022feverless} & \fullcirc[0.6ex] & \emptycirc[0.6ex] & \fullcirc[0.6ex] & \emptycirc[0.6ex] & \fullcirc[0.6ex] & \emptycirc[0.6ex] & \emptycirc[0.6ex] & \emptycirc[0.6ex] & \fullcirc[0.6ex] & \fullcirc[0.6ex] & Global DP with noise leader \\
SGBoost, 2022~\cite{zhao2022sgboost} & \fullcirc[0.6ex] & \emptycirc[0.6ex] & \fullcirc[0.6ex] & \emptycirc[0.6ex] & \emptycirc[0.6ex] & \emptycirc[0.6ex] & \fullcirc[0.6ex] & \emptycirc[0.6ex] & \emptycirc[0.6ex] & \emptycirc[0.6ex] & Design of oblivious query \\
PSO‐EVFFS, 2022~\cite{zhang2022embedded} & \fullcirc[0.6ex] & \emptycirc[0.6ex] & \fullcirc[0.6ex] & \emptycirc[0.6ex] & \emptycirc[0.6ex] & \emptycirc[0.6ex] & \fullcirc[0.6ex] & \emptycirc[0.6ex] & \emptycirc[0.6ex] & \emptycirc[0.6ex] & Feature selection based on evolutionary method \\
Xia~\etal, 2022~\cite{xia2022privacy} & \fullcirc[0.6ex] & \emptycirc[0.6ex] & \fullcirc[0.6ex] & \emptycirc[0.6ex] & \emptycirc[0.6ex] & \emptycirc[0.6ex] & \fullcirc[0.6ex] & \emptycirc[0.6ex] & \emptycirc[0.6ex] & \emptycirc[0.6ex] & Using autoencoder for better feature utilization \\
\midrule
VF-CART, 2023~\cite{xu2023vf} & \fullcirc[0.6ex] & \emptycirc[0.6ex] & \fullcirc[0.6ex] & \fullcirc[0.6ex] & \emptycirc[0.6ex] & \emptycirc[0.6ex] & \fullcirc[0.6ex] & \emptycirc[0.6ex] & \emptycirc[0.6ex] & \emptycirc[0.6ex] & Conversion from feature to bin values  \\
Mao~\etal, 2023~\cite{mao2023full} & \fullcirc[0.6ex] & \emptycirc[0.6ex] & \fullcirc[0.6ex] & \emptycirc[0.6ex] & \emptycirc[0.6ex] & \emptycirc[0.6ex] & \fullcirc[0.6ex] & \emptycirc[0.6ex] & \emptycirc[0.6ex] & \emptycirc[0.6ex] & Application to the power grid area \\
\bottomrule
\end{tabularx}
\caption{Summary of the existing secure vertical federated tree algorithms. The black dot represents ``yes'' and the white dot stands for ``no''.}
\vspace{-1cm}
\label{tab:summary_vertical_tree}
\end{table*}

\parab{Vertical Federated Tree Algorithm}


Tree models are also widely used to model tabular data. A single decision tree consists of a series of decision rules, where each sample is classified by traversing from the root and following the decision rules to the leaf nodes. In practice, it is common to build ensembles of trees, including gradient boosting decision trees (GBDT)~\cite{ke2017lightgbm}, random forests (RF)~\cite{ho1995random}, and XgBoost~\cite{chen2016xgboost}. Vertical federated tree algorithm is among researchers' most popular VFL algorithms because of its effectiveness. We show the categorization of current vertical federated tree algorithms in Tab.~\ref{tab:summary_vertical_tree}.

\vspace{-0.3cm}

\paragraph{\textbf{HE-Based Vertical Federated Tree Algorithm.}}

\cite{cheng2021secureboost,feng2019securegbm} proposed vertical federated GBDT algorithms based on HE, with~\cite{cheng2021secureboost} discussing potential information leakage. They utilize HE to encrypt the statistical values transferred from active party to passive parties. Then, the active parties could perform addition operations on the encrypted values and send them back to the active party. 
Besides, SecureBoost~\cite{cheng2021secureboost} also tries to train the first tree locally in active party and treats its prediction as the mask to reduce information leakage, a similar idea to RM. Based on SecureBoost,~\cite{zhang2022embedded} designed a vertical federated feature selection algorithm with an evolutionary optimization technique called particle swarm optimization~\cite{wang2020adaptive}.~\cite{zhu2021pivodl} studied a different setting where labels are also distributed across parties, and they leveraged HE and DP to protect the distributed features and labels. Moreover, Mao~\etal applied the HE-based vertical federated tree model to the area of power grid~\cite{mao2023full}, showing the algorithm's effectiveness. To achieve a secure query,~\cite{zhao2022sgboost} proposed an oblivious query algorithm based on~\cite{mahdikhani2020achieving}. Users could utilize the well-trained vertical federated tree model without leaking their queries and results. Besides,~\cite{xia2022privacy} further adopted autoencoder to transform the high-dimension sparse original features into low‐dimension dense representation, which is more suitable for tree models. These HE-based methods generally incur large computation overhead. Thus, several works aim to improve the efficiency of tree ensembles in VFL. VF2Boost improves vertical federated GBDT in two ways: concurrent training to reduce idle waiting and customizing encryption for training~\cite{fu2021vf2boost}. In more detail, it VF2Boost re-scheduling method to sort the encrypted values to be accumulated according to the exponential terms to reduce the number of scaling operations and thus improve efficiency. Besides, VF-CART reduces communication costs by converting feature values to bin values and building histograms~\cite{xu2023vf}.


Furthermore, several works combine HE with SS to protect the intermediate computational results from leaking to any party (including the local party). For instance, Pivot combines SS and HE to guarantee that no intermediate information is disclosed when building general tree models, including RF and GBDT~\cite{wu2020privacy}. Considering the high communication complexity of SS, SS is only utilized in the operations,~\eg, comparison, which HE does not support. Each party conducts most of the local computations with HE. Besides, the prediction path can be protected via an HE-based protocol to prevent information leakage along the path. CryptoBoost utilized the multi-party homomorphic encryption (MHE) technique~\cite{mouchet2021multiparty} to design an end-to-end secure vertical federated XgBoost~\cite{jin2022towards}. MHE takes advantage of SS to reduce the considerable computational cost introduced by HE.

\vspace{-0.2cm}

\paragraph{\textbf{SS-Based Vertical Federated Tree Algorithm}}

There are fewer vertical federated tree algorithms designed mainly using SS techniques. The reason could be that utilizing SS to implement complicated algorithms needs large design efforts. Naive designs will result in severe communicational overheads. One of the SS-based vertical federated tree algorithms is~\cite{fang2021large}, which adopts SS to train secure GBDT with data vertically partitioned to prevent intermediate information during training from leaking. Furthermore, it also provides an HE-based secure permutation method to accelerate the training process.

\paragraph{\textbf{DP-Based Vertical Federated Tree Algorithm.}}

For the~\ul{\textit{S3}} security model,~\cite{weng2020privacy} designed the reverse sum attack where the malicious party encodes magic numbers in the ciphertext to steal the private data of other parties.~\cite{luo2021feature} also designed attacks based on individual and multiple predictions. DP could be utilized to alleviate these attacks. For instance,~\cite{le2021fedxgboost} designed the FedXGBoost-LDP based on the noise perturbation of local DP~\cite{cormode2018privacy}. Each party adds noises to every message before transmitting it. Therefore, the accumulated noises could severely affect model accuracy.~\cite{wang2022feverless} provided a solution with global DP~\cite{wei2020federated}, where a selected leader adds noises. Furthermore, DP could also be leveraged to protect the prediction update leakage as a supplement to HE~\cite{zhu2021pivodl}.

\paragraph{\textbf{Vertical Federated Tree Algorithm with Other Protocols.}} Besides FDML~\cite{hu2019fdml}, which could be used to build vertical federated tree models without HE, SS or, DP,~\cite{wang2022feverless} provided another method with an MPC-based secure aggregation protocol, which is similar to~\cite{bonawitz2017practical}. And~\cite{le2021fedxgboost} proposed the FedXGBoost-SMM algorithm based on the MPC-based secure matrix multiplication protocol~\cite{karr2009privacy}.~\cite{sheikhalishahi2017privacy} proposed a privacy-preserving categorical clustering tree algorithm over vertically-partitioned data. More detailedly, this MPC-based work proposed the secure weighted averaging and secure comparison protocols to compute and compare the entropy of clusters at each leaf, thus deciding the splitting strategy.~\cite{khodaparast2018privacy} also proposed a VFL method to train random decision trees via the same secure weighted average protocol. Besides,~\cite{vaidya2008privacy} provides a novel vertical federated decision tree algorithm for~\ul{\textit{S2}} adversaries.


\parab{Vertical Federated Neural Network Algorithm}

\begin{table*}[t]
\centering
\footnotesize
\begin{tabularx}{\linewidth}{ccc|cccc|ccccl}
\toprule
\multirow{2}{*}{\bf Algorithm}
& \multicolumn{2}{c}{\makecell[c]{{\bf Protection} \\ {\bf objective (O)}}}
& \multicolumn{4}{c}{\bf Security model (S)}
& \multicolumn{4}{c}{\makecell[c]{{\bf Privacy-preserving} \\ {\bf protocol (P)}}} 
& \multirow{2}{*}{\bf Main innovation and comments} \\

& O1 & O2 & S1 & S2 & S3 & S4 & P1 & P2 & P3 & P4 & \\
\midrule
ABY, 2015~\cite{demmler2015aby} & \fullcirc[0.6ex] & \emptycirc[0.6ex] & \fullcirc[0.6ex] & \emptycirc[0.6ex] & \emptycirc[0.6ex] & \emptycirc[0.6ex] & \emptycirc[0.6ex] & \fullcirc[0.6ex] & \emptycirc[0.6ex] & \emptycirc[0.6ex] & Pioneering SS-based framework \\
\midrule
SecureML, 2017~\cite{mohassel2017secureml} & \fullcirc[0.6ex] & \emptycirc[0.6ex] & \fullcirc[0.6ex] & \emptycirc[0.6ex] & \emptycirc[0.6ex] & \emptycirc[0.6ex] & \emptycirc[0.6ex] & \fullcirc[0.6ex] & \emptycirc[0.6ex] & \emptycirc[0.6ex] & Support of secure arithmetic operations \\
\midrule
SplitNN, 2018~\cite{vepakomma2018split} & \fullcirc[0.6ex] & \emptycirc[0.6ex] & \fullcirc[0.6ex] & \emptycirc[0.6ex] & \emptycirc[0.6ex] & \emptycirc[0.6ex] & \emptycirc[0.6ex] & \emptycirc[0.6ex] & \emptycirc[0.6ex] & \fullcirc[0.6ex] & Mainstream learning scheme \\
Gupta~\etal, 2018~\cite{gupta2018distributed} & \fullcirc[0.6ex] & \emptycirc[0.6ex] & \fullcirc[0.6ex] & \emptycirc[0.6ex] & \emptycirc[0.6ex] & \emptycirc[0.6ex] & \emptycirc[0.6ex] & \emptycirc[0.6ex] & \emptycirc[0.6ex] & \fullcirc[0.6ex] & Incorporating semi-supervised learning \\
ABY$^3$, 2018~\cite{mohassel2018aby3} & \fullcirc[0.6ex] & \emptycirc[0.6ex] & \fullcirc[0.6ex] & \emptycirc[0.6ex] & \fullcirc[0.6ex] & \emptycirc[0.6ex] & \emptycirc[0.6ex] & \fullcirc[0.6ex] & \emptycirc[0.6ex] & \emptycirc[0.6ex] & Extension to three parties \\
\midrule
FDML, 2019~\cite{hu2019fdml} & \fullcirc[0.6ex] & \emptycirc[0.6ex] & \fullcirc[0.6ex] & \emptycirc[0.6ex] & \emptycirc[0.6ex] & \emptycirc[0.6ex] & \emptycirc[0.6ex] & \emptycirc[0.6ex] & \emptycirc[0.6ex] & \fullcirc[0.6ex] & Idea of ensemble learning \\
\midrule
aimNet, 2020~\cite{liu2020federated} & \fullcirc[0.6ex] & \emptycirc[0.6ex] & \fullcirc[0.6ex] & \emptycirc[0.6ex] & \emptycirc[0.6ex] & \emptycirc[0.6ex] & \emptycirc[0.6ex] & \emptycirc[0.6ex] & \emptycirc[0.6ex] & \fullcirc[0.6ex] & Fusing representations from different tasks \\
Zhang~\etal, 2020~\cite{zhang2020additively} & \fullcirc[0.6ex] & \emptycirc[0.6ex] & \fullcirc[0.6ex] & \emptycirc[0.6ex] & \emptycirc[0.6ex] & \emptycirc[0.6ex] & \fullcirc[0.6ex] & \emptycirc[0.6ex] & \emptycirc[0.6ex] & \halfcirc[0.6ex] & Combining HE and RM for further security \\
VFGNN, 2020~\cite{zhou2020vertically} & \fullcirc[0.6ex] & \emptycirc[0.6ex] & \fullcirc[0.6ex] & \emptycirc[0.6ex] & \emptycirc[0.6ex] & \emptycirc[0.6ex] & \emptycirc[0.6ex] & \fullcirc[0.6ex] & \halfcirc[0.6ex] & \emptycirc[0.6ex] & VFL GNN algorithm combining SS and DP\\
VAFL, 2020~\cite{chen2020vafl} & \fullcirc[0.6ex] & \emptycirc[0.6ex] & \fullcirc[0.6ex] & \emptycirc[0.6ex] & \emptycirc[0.6ex] & \emptycirc[0.6ex] & \emptycirc[0.6ex] & \emptycirc[0.6ex] & \fullcirc[0.6ex] & \emptycirc[0.6ex] & Asynchronous LSTM-based method \\
\midrule
Cha, 2021~\cite{cha2021implementing} & \fullcirc[0.6ex] & \emptycirc[0.6ex] & \fullcirc[0.6ex] & \emptycirc[0.6ex] & \emptycirc[0.6ex] & \emptycirc[0.6ex] & \emptycirc[0.6ex] & \emptycirc[0.6ex] & \emptycirc[0.6ex] & \fullcirc[0.6ex] & Adoption of autoencoder \\
SS-VFNAS, 2021~\cite{liang2021self} & \fullcirc[0.6ex] & \emptycirc[0.6ex] & \fullcirc[0.6ex] & \emptycirc[0.6ex] & \emptycirc[0.6ex] & \emptycirc[0.6ex] & \emptycirc[0.6ex] & \emptycirc[0.6ex] & \halfcirc[0.6ex] & \fullcirc[0.6ex] & A self-supervised solution for VFL NAS \\
AsyREVEL, 2021~\cite{zhang2021desirable} & \fullcirc[0.6ex] & \fullcirc[0.6ex] & \fullcirc[0.6ex] & \fullcirc[0.6ex] & \fullcirc[0.6ex] & \emptycirc[0.6ex] & \emptycirc[0.6ex] & \emptycirc[0.6ex] & \emptycirc[0.6ex] & \fullcirc[0.6ex] & Utilizing zeroth-order optimization \\
Li~\etal, 2021~\cite{li2021label} & \fullcirc[0.6ex] & \emptycirc[0.6ex] & \fullcirc[0.6ex] & \emptycirc[0.6ex] & \emptycirc[0.6ex] & \emptycirc[0.6ex] & \emptycirc[0.6ex] & \emptycirc[0.6ex] & \emptycirc[0.6ex] & \fullcirc[0.6ex] & Randomly permuting gradients for security \\
DRAVL, 2021~\cite{sun2021defending} & \fullcirc[0.6ex] & \emptycirc[0.6ex] & \fullcirc[0.6ex] & \emptycirc[0.6ex] & \emptycirc[0.6ex] & \emptycirc[0.6ex] & \emptycirc[0.6ex] & \emptycirc[0.6ex] & \emptycirc[0.6ex] & \fullcirc[0.6ex] & Adoption of adversarial training \\
Dalskov~\etal, 2021~\cite{dalskov2021fantastic} & \fullcirc[0.6ex] & \emptycirc[0.6ex] & \fullcirc[0.6ex] & \emptycirc[0.6ex] & \fullcirc[0.6ex] & \emptycirc[0.6ex] & \emptycirc[0.6ex] & \fullcirc[0.6ex] & \emptycirc[0.6ex] & \emptycirc[0.6ex] & Extension to four parties \\
FedSim, 2021~\cite{wu2021exploiting} & \fullcirc[0.6ex] & \emptycirc[0.6ex] & \fullcirc[0.6ex] & \emptycirc[0.6ex] & \emptycirc[0.6ex] & \emptycirc[0.6ex] & \emptycirc[0.6ex] & \emptycirc[0.6ex] & \emptycirc[0.6ex] & \fullcirc[0.6ex] & PPRL's similarity results as input \\
FKGE, 2021~\cite{peng2021differentially} & \fullcirc[0.6ex] & \emptycirc[0.6ex] & \fullcirc[0.6ex] & \emptycirc[0.6ex] & \emptycirc[0.6ex] & \emptycirc[0.6ex] & \emptycirc[0.6ex] & \emptycirc[0.6ex] & \halfcirc[0.6ex] & \fullcirc[0.6ex] & Federated knowledge graphs embedding \\
STFL, 2021~\cite{chu2021privacy} & \fullcirc[0.6ex] & \emptycirc[0.6ex] & \fullcirc[0.6ex] & \emptycirc[0.6ex] & \emptycirc[0.6ex] & \emptycirc[0.6ex] & \emptycirc[0.6ex] & \emptycirc[0.6ex] & \emptycirc[0.6ex] & \fullcirc[0.6ex] & Adoption of self-taught learning \\
Kholod~\etal, 2021~\cite{kholod2021parallelization} & \fullcirc[0.6ex] & \emptycirc[0.6ex] & \fullcirc[0.6ex] & \emptycirc[0.6ex] & \emptycirc[0.6ex] & \emptycirc[0.6ex] & \emptycirc[0.6ex] & \emptycirc[0.6ex] & \emptycirc[0.6ex] & \fullcirc[0.6ex] & Utilizing self-organizing map model \\
FedSGC, 2021~\cite{cheung2021fedsgc} & \fullcirc[0.6ex] & \emptycirc[0.6ex] & \fullcirc[0.6ex] & \emptycirc[0.6ex] & \emptycirc[0.6ex] & \emptycirc[0.6ex] & \fullcirc[0.6ex] & \emptycirc[0.6ex] & \emptycirc[0.6ex] & \emptycirc[0.6ex] & HE-based vertical federated GNN algorithm \\
FedVGCN, 2021~\cite{ni2021vertical} & \fullcirc[0.6ex] & \emptycirc[0.6ex] & \fullcirc[0.6ex] & \emptycirc[0.6ex] & \emptycirc[0.6ex] & \emptycirc[0.6ex] & \fullcirc[0.6ex] & \emptycirc[0.6ex] & \emptycirc[0.6ex] & \emptycirc[0.6ex] & Fully unsupervised setting \\
\midrule
Sun~\etal, 2022~\cite{sun2022label} & \fullcirc[0.6ex] & \emptycirc[0.6ex] & \fullcirc[0.6ex] & \emptycirc[0.6ex] & \emptycirc[0.6ex] & \emptycirc[0.6ex] & \emptycirc[0.6ex] & \emptycirc[0.6ex] & \emptycirc[0.6ex] & \fullcirc[0.6ex] & Preventing leakage from forward embedding \\
SFA, 2022~\cite{cai2022secure} & \fullcirc[0.6ex] & \emptycirc[0.6ex] & \fullcirc[0.6ex] & \emptycirc[0.6ex] & \emptycirc[0.6ex] & \emptycirc[0.6ex] & \emptycirc[0.6ex] & \emptycirc[0.6ex] & \emptycirc[0.6ex] & \fullcirc[0.6ex] & Trade-off between privacy and accuracy \\
CoAE, 2022~\cite{liu2022batch} & \fullcirc[0.6ex] & \emptycirc[0.6ex] & \fullcirc[0.6ex] & \emptycirc[0.6ex] & \fullcirc[0.6ex] & \emptycirc[0.6ex] & \halfcirc[0.6ex] & \emptycirc[0.6ex] & \emptycirc[0.6ex] & \fullcirc[0.6ex] & Utilization of autorencoder to hide labels \\
TPSL, 2022~\cite{yang2022differentially} & \fullcirc[0.6ex] & \emptycirc[0.6ex] & \fullcirc[0.6ex] & \emptycirc[0.6ex] & \emptycirc[0.6ex] & \emptycirc[0.6ex] & \emptycirc[0.6ex] & \emptycirc[0.6ex] & \halfcirc[0.6ex] & \fullcirc[0.6ex] & Permuting both weights and transmitted messages \\
Cheetah, 2022~\cite{huang2022cheetah} & \fullcirc[0.6ex] & \emptycirc[0.6ex] & \fullcirc[0.6ex] & \emptycirc[0.6ex] & \emptycirc[0.6ex] & \emptycirc[0.6ex] & \emptycirc[0.6ex] & \fullcirc[0.6ex] & \emptycirc[0.6ex] & \emptycirc[0.6ex] & Two-party secure inference \\
FedCVT, 2022~\cite{kang2022fedcvt} & \fullcirc[0.6ex] & \emptycirc[0.6ex] & \fullcirc[0.6ex] & \emptycirc[0.6ex] & \emptycirc[0.6ex] & \emptycirc[0.6ex] & \emptycirc[0.6ex] & \emptycirc[0.6ex] & \emptycirc[0.6ex] & \fullcirc[0.6ex] & Multi-view learning to utilize unaligned entities \\
FedDA, 2022~\cite{zhang2022data} & \fullcirc[0.6ex] & \emptycirc[0.6ex] & \fullcirc[0.6ex] & \emptycirc[0.6ex] & \emptycirc[0.6ex] & \emptycirc[0.6ex] & \emptycirc[0.6ex] & \emptycirc[0.6ex] & \emptycirc[0.6ex] & \fullcirc[0.6ex] & Utilizing GAN to solve the shortage of overlappings \\
Ren~\etal, 2022~\cite{ren2022improving} & \fullcirc[0.6ex] & \emptycirc[0.6ex] & \fullcirc[0.6ex] & \emptycirc[0.6ex] & \emptycirc[0.6ex] & \emptycirc[0.6ex] & \emptycirc[0.6ex] & \emptycirc[0.6ex] & \emptycirc[0.6ex] & \fullcirc[0.6ex] & Adoption of knowledge distillation and OT \\
Li~\etal, 2022~\cite{li2022nearest} & \fullcirc[0.6ex] & \emptycirc[0.6ex] & \fullcirc[0.6ex] & \emptycirc[0.6ex] & \emptycirc[0.6ex] & \emptycirc[0.6ex] & \fullcirc[0.6ex] & \emptycirc[0.6ex] & \emptycirc[0.6ex] & \emptycirc[0.6ex] & Sampling strategy of imbalanced data for GNN \\
VFLFS, 2022~\cite{feng2022vertical} & \fullcirc[0.6ex] & \emptycirc[0.6ex] & \fullcirc[0.6ex] & \emptycirc[0.6ex] & \emptycirc[0.6ex] & \emptycirc[0.6ex] & \emptycirc[0.6ex] & \emptycirc[0.6ex] & \emptycirc[0.6ex] & \fullcirc[0.6ex] & Vertical federated feature selection \\
BlindFL, 2022~\cite{fu2022blindfl} & \fullcirc[0.6ex] & \emptycirc[0.6ex] & \fullcirc[0.6ex] & \emptycirc[0.6ex] & \emptycirc[0.6ex] & \emptycirc[0.6ex] & \fullcirc[0.6ex] & \fullcirc[0.6ex] & \emptycirc[0.6ex] & \emptycirc[0.6ex] & Extension of CAESAR~\cite{chen2021homomorphic} \\
FedOnce, 2022~\cite{wu2022practical} & \fullcirc[0.6ex] & \emptycirc[0.6ex] & \fullcirc[0.6ex] & \emptycirc[0.6ex] & \emptycirc[0.6ex] & \emptycirc[0.6ex] & \emptycirc[0.6ex] & \emptycirc[0.6ex] & \fullcirc[0.6ex] & \emptycirc[0.6ex] & Reducing inter-party privacy loss \\
Zou~\etal, 2022~\cite{zou2022defending} & \fullcirc[0.6ex] & \emptycirc[0.6ex] & \fullcirc[0.6ex] & \emptycirc[0.6ex] & \fullcirc[0.6ex] & \emptycirc[0.6ex] & \halfcirc[0.6ex] & \emptycirc[0.6ex] & \emptycirc[0.6ex] & \fullcirc[0.6ex] & Privacy leakage from batch-level intermediate results \\
FLFHNN, 2022~\cite{sun2022flfhnn} & \fullcirc[0.6ex] & \emptycirc[0.6ex] & \fullcirc[0.6ex] & \emptycirc[0.6ex] & \emptycirc[0.6ex] & \emptycirc[0.6ex] & \fullcirc[0.6ex] & \emptycirc[0.6ex] & \emptycirc[0.6ex] & \fullcirc[0.6ex] & Reducing communication cost by FHE \\
Yan~\etal, 2022~\cite{yan2022multi} & \fullcirc[0.6ex] & \emptycirc[0.6ex] & \fullcirc[0.6ex] & \emptycirc[0.6ex] & \emptycirc[0.6ex] & \emptycirc[0.6ex] & \emptycirc[0.6ex] & \emptycirc[0.6ex] & \emptycirc[0.6ex] & \fullcirc[0.6ex] & Practical secure aggregation protocol based on ECDH \\
AdptVFedConv, 2022~\cite{li2022adaptive} & \fullcirc[0.6ex] & \emptycirc[0.6ex] & \fullcirc[0.6ex] & \emptycirc[0.6ex] & \emptycirc[0.6ex] & \emptycirc[0.6ex] & \emptycirc[0.6ex] & \emptycirc[0.6ex] & \emptycirc[0.6ex] & \fullcirc[0.6ex] & Vertical federated CNN model \\
Wang~\etal, 2022~\cite{wang2022vertical} & \fullcirc[0.6ex] & \emptycirc[0.6ex] & \fullcirc[0.6ex] & \emptycirc[0.6ex] & \emptycirc[0.6ex] & \emptycirc[0.6ex] & \emptycirc[0.6ex] & \emptycirc[0.6ex] & \emptycirc[0.6ex] & \fullcirc[0.6ex] & Knowledge distillation \\
\midrule
Yang~\etal, 2023~\cite{yang2023hybrid} & \fullcirc[0.6ex] & \emptycirc[0.6ex] & \fullcirc[0.6ex] & \emptycirc[0.6ex] & \emptycirc[0.6ex] & \emptycirc[0.6ex] & \emptycirc[0.6ex] & \fullcirc[0.6ex] & \emptycirc[0.6ex] & \fullcirc[0.6ex] & Hybrid protocol with SS and GC \\
LabelGuard, 2023~\cite{xia2023cascade} & \fullcirc[0.6ex] & \emptycirc[0.6ex] & \fullcirc[0.6ex] & \emptycirc[0.6ex] & \emptycirc[0.6ex] & \emptycirc[0.6ex] & \emptycirc[0.6ex] & \emptycirc[0.6ex] & \emptycirc[0.6ex] & \fullcirc[0.6ex] & Distributed labels and straggler mitigation \\
\bottomrule
\end{tabularx}
\caption{Summary of the existing secure vertical federated neural network algorithms. The black dot represents ``yes'' and the white dot stands for ``no''.}
\vspace{-1cm}
\label{tab:summary_vertical_nn}
\end{table*}

Deep neural network (DNN)~\cite{yegnanarayana2009artificial} has become increasingly popular with the advances in high-quality, large-scale data and powerful computation facilities. DNNs generally contain stacked neural network layers and non-linear activation functions. Vertical federated neural network also takes a large portion in VFL. How to split the model under the concern of security and accuracy turns out to be the core design problem. We summarize these works in Tab.~\ref{tab:summary_vertical_nn}.


\paragraph{\textbf{SplitNN-Based Vertical Federated Neural Network.}}

Existing vertical DNN models generally follow a split neural network (SplitNN) architecture~\cite{vepakomma2018split,gupta2018distributed}. In SplitNN, each passive party holds a local bottom model mapping features to latent embeddings, while the active party gathers the embeddings and combines them with its own features to complete the forward pass. Based on the SplitNN architecture, Liu~\etal proposed a specific VFL framework to solve visual-language-grounding problems such as image captioning~\cite{liu2020federated}. Furthermore,~\cite{cha2021implementing} utilizes an overcomplete autoencoder to generate embeddings in each party (task). Then, the latent representation from different parties is directly aggregated for a specific VFL task. The authors claim that the latent representation differs from the original data, thus, enabling security. In addition, Li~\etal utilized the SplitNN architecture to design a novel convolutional neural network (CNN)~\cite{li2021survey} algorithm for image data~\cite{li2022adaptive}. Each party trains a local feature extractor and sends the latent representation to the server as the input of the classifier model. For graph data,~\cite{zhang2021federated} defined the vertical federated graph neural network (GNN) algorithm~\cite{kipf2016semi,hamilton2017inductive,gilmer2017neural} as all parties having the same set of nodes but different sets of edges and features.


\underline{Attack and Defence.} The above works could achieve high-efficiency performance but contain severe privacy problems.~\cite{fu2022label} finds that SplitNN-based vertical federated neural network is vulnerable to label inference attack. Label inference attack is conducted by the passive party and aims to leak the labels owned by the active party. The authors proposed three kinds of label inference attack,~\ie, passive label inference attack, active label inference attack, and direct label inference attack, for the SplitNN-based vertical neural network. Adversaries could infer the labels of the active party from both the received plaintext gradients and obtained plaintext final model weights. Although the proposed attacks are very effective, they make a strong assumption that auxiliary labels are provided for the adversary, which could be unavailable in practice. Besides,~\cite{sun2022label} indicated that the forward embedding could also disclose privacy.~\cite{liu2022clustering} proposed a passive clustering inference attack to reveal the private labels.~\cite{liu2022batch} designed the inversion and replacement attacks to disclose private labels from batch-level messages. In addition, since GNN is vulnerable to adversarial attacks,~\cite{chen2022graph} designed a novel adversarial attack for vertical federated GNN model and found the attack is also effective. Besides,~\cite{qiu2022your} designed novel attacks to reveal the leakage of samples' relations.

The defense against these attacks is still an open problem, although many efforts have been made. For example, the authors of~\cite{sun2022label} proposed to alleviate leakage via reducing the correlation of the embedding and labels. Besides,~\cite{cai2022secure} found out that there exists a trade-off between vulnerability to inference attacks and model accuracy. It designed a novel security forward aggregation (SFA) protocol to protect the privacy of raw data. In addition,~\cite{li2021label} indicated that randomly perturbing the gradients before transmission effectively protects privacy.~\cite{sun2021defending} designed a framework (DRAVL) based on adversarial training, which could defend against reconstruction attacks. In addition,~\cite{kholod2021parallelization} studied the self-organizing map model for VFL. It proposes a parallelization method to reduce data transmission, thus decreasing the learning cost and privacy leakage risk. Besides, Zou~\etal found that batch-level intermediate results could also disclose the privacy of label~\cite{zou2022defending}. The authors defended this attack with an adjustable autoencoder to confuse the adversary without hurting model accuracy. Xia~\etal proposed LabelGuard to defend against label inference attack via a novel learning method~\cite{xia2023cascade}. LabelGuard minimizes the VFL task training loss and the performance of estimated adversary models.~\cite{liu2022batch} proposed a confusional autoencoder (CoAE) method to defend the label inference attack. Besides, Yan~\etal utilized a practical secure aggregation protocol based on ECDH~\cite{barker2017recommendation} to provide additional security guarantee~\cite{yan2022multi}.

\underline{Training Data Insufficiency.} Most VFL algorithms assume that labeled data is sufficient to train a high-quality model. However, it may not hold in VFL, as the training data are determined by the entity alignment of different parties. When only a limited number of entities co-exist among parties, we will face the problem of training data and label scarcity. Several methods aim to transfer knowledge among similar samples to complement missing features. For the scenarios where parties own sample IDs to conduct PSI, Kang~\etal proposed FedCVT as a multi-view learning framework to utilize non-overlapping entities for VFL via pseudo-feature and pseudo-label generation~\cite{kang2022fedcvt}. Consider the two-party case, the overlapping samples form a cross-party view, while non-overlapping samples at each party form unique intra-party views. For each party, two neural networks are learned to capture features for the cross-party and intra-party views. Pseudo-features and labels are generated by querying the intra-party view of the other party via dot-product attention. Besides,~\cite{zhang2022data} also utilizes GAN~\cite{goodfellow2020generative} to address the shortage problem of overlapping entities. The authors generate more training samples by learning from the feature correlation of limited aligned entities and abundant unaligned ones.~\cite{chu2021privacy} performs self-taught learning~\cite{raina2007self} over all local samples at each party to extract useful features and benefit the VFL training process.~\cite{wang2022vertical} applies knowledge distillation~\cite{hinton2015distilling,gou2021knowledge} on the federated representation of shared samples learned by~\cite{chai2021federated} and transfers the knowledge to enrich the representation of local samples.~\cite{ren2022improving} also provides a solution based on knowledge distillation. The well-trained VFL model is regarded as a teacher model to generate teacher labels for the aligned entities. Then, a local student model is trained with both local labels and teacher labels to realize knowledge transfer. Moreover,~\cite{feng2022vertical} proposed to utilize the non-overlapping samples to conduct vertical federated feature selection. Furthermore, when using PPRL instead of PSI for entity alignment in the scenarios like GPS location, the output of entity alignment is the similarity score but not binary matching result,~\ie, aligned or unaligned. For using such similarity results, we could turn them to binary results with a threshold. However, this intuitive method causes information loss. To address this problem, Wu~\etal consider an alternative case where parties have overlapping features, upon which top-$k$ fuzzy sample matching can be performed~\cite{wu2021exploiting}. Specifically, for each sample $i$ in party A, a secure top-$k$ nearest neighbor search is performed in party B. The top-$k$ similar samples of $i$ will be fed into the VFL model via a weighted sum to complement its features.






\vspace{-0.3cm}

\paragraph{\textbf{HE-Based Vertical Federated Neural Network.}}

Based on the architecture of SplitNN,~\cite{zhang2020additively} combined Paillier with multiplicative RM to protect the privacy of the aggregated embedding. To reduce the communication cost introduced by Paillier, FLFHNN~\cite{sun2022flfhnn} adopted the CKKS scheme as the privacy-preserving method. Besides, Cheung~\etal proposed an HE-based vertical federated GNN~\cite{cheung2021fedsgc}, designed without activation functions, thus being linear and well supported by HE. Furthermore, Ni~\etal also studied vertical GNN via HE, tackling the non-linear activation functions via approximation~\cite{ni2021vertical}. Moreover,~\cite{li2022nearest} proposed an HE-based sample strategy of imbalanced data for vertical federated graph embedding.

\paragraph{\textbf{SS-Based Vertical Federated Neural Network.}}

We could also adopt SS instead of SplitNN from the two-party setting~\cite{demmler2015aby,mohassel2017secureml,huang2022cheetah}, three-party setting~\cite{mohassel2018aby3} to four-party setting~\cite{dalskov2021fantastic}. The SS-based approaches allow more strict security models,~\eg,~\ul{\textit{S3}}, but always suffer from worse efficiency. Besides,~\cite{zhou2020vertically} takes a different approach to implement vertical GNN by distributing and gathering node embeddings from different parties using SS. DP is also adopted to enhance privacy. In addition, BlindFL combined HE and SS to design a vertical federated learning method, which is suitable for sparse and categorical features~\cite{fu2022blindfl}. We could regard BlindFL as an extension of CAESAR~\cite{chen2021homomorphic}. Yang~\etal also proposed a hybrid protocol for VFL~\cite{yang2023hybrid}, where SS and GC support linear and non-linear operations separately.

\paragraph{\textbf{DP-Based Vertical Federated Neural Network.}}

Vertical federated NN algorithms based on NN have also been studied. For example, TPSL adopted the DP technique to make obfuscation to defend against the label leakage attack~\cite{yang2022differentially}. It permutes both the model weights and the transmitted intermediate results to provide a provable guarantee.~\cite{liang2021self} proposed a self-supervised solution to neural architecture search (NAS)~\cite{zhu2021from} for VFL with differential privacy. In addition, Peng~\etal proposed a DP-based method to further improve the embeddings of non-overlapping entities in the vertical federated knowledge graphs embedding task~\cite{peng2021differentially}. Besides,~\cite{chen2020vafl} combined DP with the asynchronous LSTM-based method and analyzed its convergence. With the adoption of DP, FedOnce~\cite{wu2022practical} showed that privacy loss is accumulated across not only iterations but also parties. Therefore, FedOnce proposed a new DP mechanism to slow down inter-party privacy loss.


\begin{table*}[h]
\centering
\footnotesize
\begin{tabularx}{\linewidth}{ccc|cccc|ccccl}
\toprule
\multirow{2}{*}{\bf Algorithm}
& \multicolumn{2}{c}{\makecell[c]{{\bf Protection} \\ {\bf objective (O)}}}
& \multicolumn{4}{c}{\bf Security model (S)}
& \multicolumn{4}{c}{\makecell[c]{{\bf Privacy-preserving} \\ {\bf protocol (P)}}} 
& \multirow{2}{*}{\bf Main innovation and comments} \\

& O1 & O2 & S1 & S2 & S3 & S4 & P1 & P2 & P3 & P4 & \\
\midrule
Vaidya~\etal, 2002~\cite{vaidya2002privacy} & \fullcirc[0.6ex] & \emptycirc[0.6ex] & \fullcirc[0.6ex] & \emptycirc[0.6ex] & \emptycirc[0.6ex] & \emptycirc[0.6ex] & \emptycirc[0.6ex] & \emptycirc[0.6ex] & \emptycirc[0.6ex] & \fullcirc[0.6ex] & Privacy-preserving association rule mining \\
\midrule
Vaidya~\etal, 2003~\cite{vaidya2003privacy} & \fullcirc[0.6ex] & \emptycirc[0.6ex] & \fullcirc[0.6ex] & \fullcirc[0.6ex] & \fullcirc[0.6ex] & \emptycirc[0.6ex] & \fullcirc[0.6ex] & \emptycirc[0.6ex] & \emptycirc[0.6ex] & \fullcirc[0.6ex] & Discussion of more strict security models \\
\midrule
Yu~\etal, 2006~\cite{yu2006privacy} & \fullcirc[0.6ex] & \emptycirc[0.6ex] & \fullcirc[0.6ex] & \emptycirc[0.6ex] & \emptycirc[0.6ex] & \emptycirc[0.6ex] & \emptycirc[0.6ex] & \emptycirc[0.6ex] & \emptycirc[0.6ex] & \fullcirc[0.6ex] & Secure sum protocol \\
\midrule
PPSVM, 2009~\cite{yunhong2009privacy} & \fullcirc[0.6ex] & \emptycirc[0.6ex] & \fullcirc[0.6ex] & \emptycirc[0.6ex] & \emptycirc[0.6ex] & \emptycirc[0.6ex] & \emptycirc[0.6ex] & \emptycirc[0.6ex] & \emptycirc[0.6ex] & \fullcirc[0.6ex] & Protecting security by matrix factorization theory \\
\midrule
Hetero-EM, 2020~\cite{ou2020homomorphic} & \fullcirc[0.6ex] & \emptycirc[0.6ex] & \fullcirc[0.6ex] & \emptycirc[0.6ex] & \emptycirc[0.6ex] & \emptycirc[0.6ex] & \fullcirc[0.6ex] & \emptycirc[0.6ex] & \emptycirc[0.6ex] & \emptycirc[0.6ex] & Vertical federated expectation maximization \\
FDSKL, 2020~\cite{gu2020federated} & \fullcirc[0.6ex] & \emptycirc[0.6ex] & \fullcirc[0.6ex] & \emptycirc[0.6ex] & \emptycirc[0.6ex] & \emptycirc[0.6ex] & \emptycirc[0.6ex] & \emptycirc[0.6ex] & \emptycirc[0.6ex] & \fullcirc[0.6ex] & Considering both data~\& model privacy \\
VFKL, 2020~\cite{dang2020large} & \fullcirc[0.6ex] & \emptycirc[0.6ex] & \fullcirc[0.6ex] & \emptycirc[0.6ex] & \emptycirc[0.6ex] & \emptycirc[0.6ex] & \emptycirc[0.6ex] & \emptycirc[0.6ex] & \emptycirc[0.6ex] & \fullcirc[0.6ex] & Approximation with random feature technique \\
SeSoRec, 2020~\cite{chen2020secure} & \fullcirc[0.6ex] & \emptycirc[0.6ex] & \fullcirc[0.6ex] & \emptycirc[0.6ex] & \emptycirc[0.6ex] & \emptycirc[0.6ex] & \emptycirc[0.6ex] & \fullcirc[0.6ex] & \emptycirc[0.6ex] & \emptycirc[0.6ex] & Vertical federated social recommendation \\
\midrule
HOFMs, 2021~\cite{atarashi2021vertical} & \fullcirc[0.6ex] & \emptycirc[0.6ex] & \fullcirc[0.6ex] & \emptycirc[0.6ex] & \emptycirc[0.6ex] & \emptycirc[0.6ex] & \emptycirc[0.6ex] & \emptycirc[0.6ex] & \emptycirc[0.6ex] & \fullcirc[0.6ex] & High-order factorization machine \\
\midrule
VFedAKPCA, 2022~\cite{cheung2022vertical} & \fullcirc[0.6ex] & \emptycirc[0.6ex] & \fullcirc[0.6ex] & \emptycirc[0.6ex] & \emptycirc[0.6ex] & \emptycirc[0.6ex] & \emptycirc[0.6ex] & \emptycirc[0.6ex] & \emptycirc[0.6ex] & \fullcirc[0.6ex] & Principle component analysis \\
Zhang~\etal, 2022~\cite{zhang2022secure} & \fullcirc[0.6ex] & \emptycirc[0.6ex] & \fullcirc[0.6ex] & \emptycirc[0.6ex] & \emptycirc[0.6ex] & \emptycirc[0.6ex] & \halfcirc[0.6ex] & \fullcirc[0.6ex] & \emptycirc[0.6ex] & \emptycirc[0.6ex] & Hybrid protocol for feature selection \\
\bottomrule
\end{tabularx}
\caption{Summary of the other secure vertical federated algorithms besides vertical federated linear algorithm, vertical federated tree algorithm, and vertical federated neural network. The black dot represents ``yes'' and the white dot stands for ``no''.}
\vspace{-0.6cm}
\label{tab:summary_vertical_other}
\end{table*}

\parab{Other Vertical Federated Algorithms}


Besides vertical federated linear, tree, and neural network algorithms, other VFL algorithms exist, which are essential for real-world scenarios. We conclude them in Tab.~\ref{tab:summary_vertical_other}.

\paragraph{\textbf{Other Vertical Federated Algorithms Based on HE.}}

Ou~\etal studied Expectation maximization (EM) algorithms in the VFL scenario~\cite{ou2020homomorphic}. EM algorithms are commonly used to infer probabilistic generative models with latent variables, such as k-means, topic models, variational autoencoders,~\etc. The authors proposed an HE-based algorithm that performs E steps locally and M steps using aggregated parameters. HE-based protocol is used to encrypt local calculation results. Vaidya~\etal proposed a privacy-preserving K-means clustering model on vertically partitioned data~\cite{vaidya2003privacy}. The algorithm is developed based on a secure add-and-compare protocol and an HE-based secure permutation protocol to secretly assign each sample to the nearest cluster. The authors also discussed situations where the adversaries are~\ul{\textit{S2}} and~\ul{\textit{S3}}.
\vspace{-0.1cm}

\paragraph{\textbf{Other Vertical Federated Algorithms Based on SS.}}

~\cite{chen2020secure} proposed SeSoRec as a vertical federated recommendation framework based on matrix factorization with MPC-based protocol. SeSoRec could allow collaborative training between the user-item interaction and user social platforms. Besides,~\cite{zhang2022secure} utilized SS to design a feature selection method based on Gini impurity, which could improve the accuracy of down-stream learning task. Furthermore, the authors enhanced the protocol with the additional HE adoption to reduce communication overhead.

\vspace{-0.1cm}

\paragraph{\textbf{Other Vertical Federated Algorithms Based on Other Protocols.}}

~\cite{yu2006privacy} studied kernel methods in support vector machines (SVM) over vertically partitioned data. The kernel method is a broad class of non-linear machine learning models~\cite{hofmann2008kernel} commonly used in bioinformatics. The critical insight of~\cite{yu2006privacy} is to represent the kernel function as matrices computed locally and to recover the global kernel matrix via secure sum protocols. However,~\cite{yu2006privacy} suffers from communication overhead caused by merging local kernel matrices. Therefore,~\cite{yunhong2009privacy} designed a more efficient protection method for SVM with matrix factorization theory. Besides, Cheung~\etal studied the problem of principle component analysis in VFL~\cite{cheung2022vertical}. The authors leverage power iteration locally to compute eigenvalues and eigenvectors of local covariance matrices before merging the local eigenvectors to form the global eigenvector. Moreover, the authors propose a fully decentralized version to perform PCA in VFL, with eigenvectors communication between immediate neighbors. They stated that this work fulfills the standard security,~\ie, only intermediate results are transmitted while raw data is maintained locally.~\cite{vaidya2002privacy} provided a vertical federated association rule mining algorithm, which designed an RM-based secure scalar product method. This method only allows computation over private data but discloses no privacy. In addition,~\cite{dang2020large} proposed to leverage a random feature approximation technique to decouple non-linear kernels into linearly separable computations. In addition, Atarashi~\etal proposed to learn high-order factorization machines~\cite{blondel2016higher}, a commonly used model in recommender system, under the VFL setting~\cite{atarashi2021vertical}. They use ANOVA kernels' recursion and can be applied with HE to enhance privacy.~\cite{gu2020federated} proposed a vertical federated kernel algorithm,~\ie, FDSKL, which could handle linearly non-separable data. This RM-based method considers both data privacy and model privacy. FDSKL extended the synchronous gradient descent algorithm~\cite{yang2019federated} to non-linear kernel methods.


{\bf (D4) Compiler of vertical federated machine learning algorithm.} Currently, various  vertical federated machine learning algorithms need different hand-crafted designs, while HFL has an unified architecture. This is why SOTA machine learning algorithms,~\eg, large language models~\cite{brown2020language}, are always hard to be adopted in VFL. Therefore, the VFL compiler is of urgent need to automatically design federated algorithms given arbitary machine learning models and privacy-preserving primitives.

{\bf (D5) Hybrid privacy-preserving protocol.} We observe that different security primitives have their own pros and cons. For example, SS-based protocols own large communication overheads but scarcely affect accuracy, while DP-based protocols could maintain good efficiency but suffer from severe accuracy loss. In the future, there will be more hybrid protocols to balance different costs.






\section{Layer Four: System Design of VFL}
\label{sec:framework_design}
Based on different vertical federated algorithms, the VFL system is designed in various aspects, such as incentive mechanism, interpretability, efficiency optimization, robustness enhancement, and verification.

\subsection{Incentive Mechanism} 
\label{sec:incentive_mechanism}

\begin{figure}[h]
	\centering
	\includegraphics[width=0.5\linewidth]{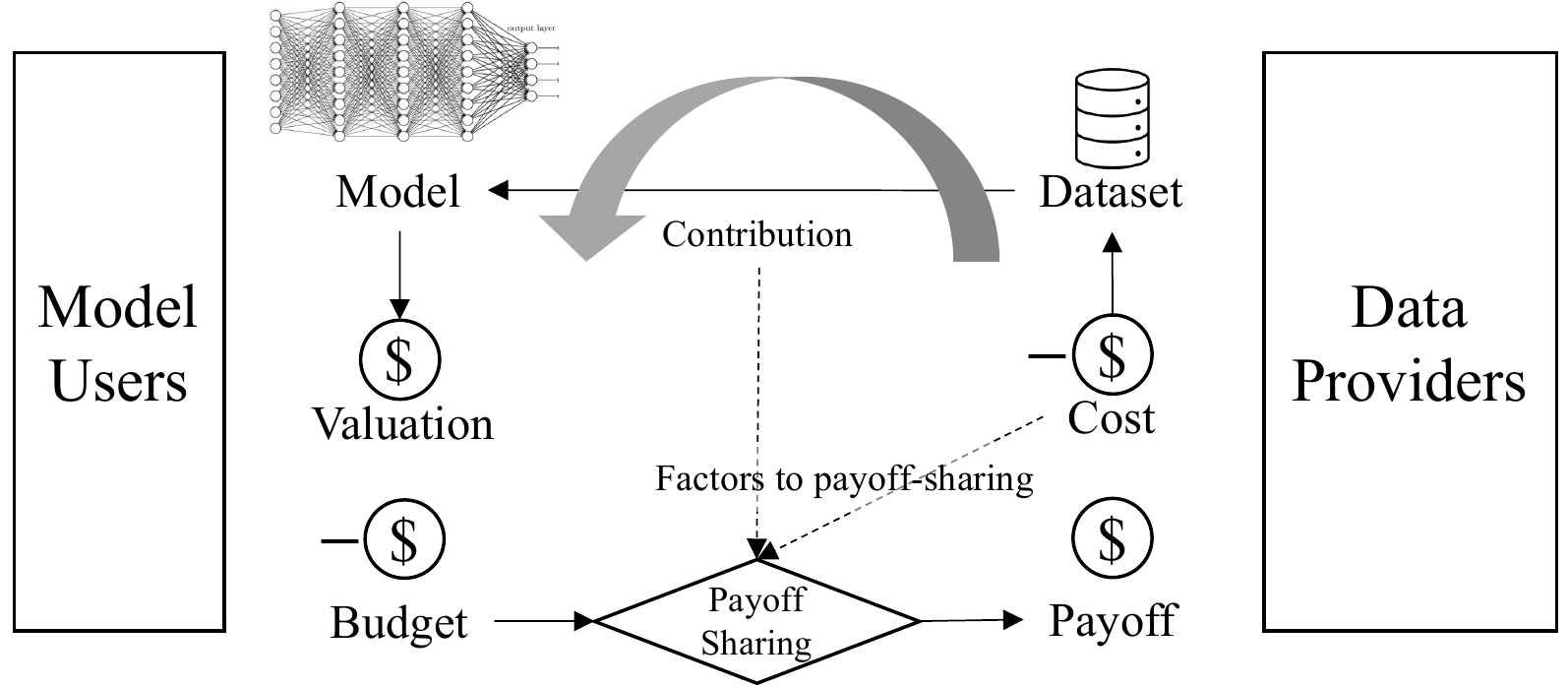}
	\caption{Design of incentive mechanism in vertical federated learning.}
	\vspace{-0.4cm}
	\label{fig:monetary-incentive}
\end{figure}

The incentive mechanism plays an essential role in designing the VFL framework for commercial usage, which could promote the truth-telling of the participants and fair profit allocation, consequently encouraging more parties to join the VFL community. In this section, we introduce a common monetary incentive mechanism design where participants purchase in a monetary way. The design contains two types of roles,~\ie, model user and data provider, shown in Fig.~\ref{fig:monetary-incentive}. Data providers contribute data with cost,~\eg, data collection and preprocessing. Model users utilize the provided dataset in a privacy-preserving manner to build a machine learning model, which brings value,~\eg, model performance improvement. Consequently, the model users must transfer a certain amount of payoff to the data providers. Specifically, each model user holds a certain budget amount, which will be subsequently distributed to the data providers as the payoff. The VFL monetary incentive mechanism design mainly contains three aspects: 1) the contribution evaluation of each data provider; 2) the quantification of supply-side data cost and demand-side model value; 3) the design of payoff sharing schema,~\ie, determining the demand-side budget and supply-side payoff.


\subsubsection{Contribution Evaluation.} 
Datasets from different providers should have non-identical contributions to the demand side's model. Intuitively, providers with high contributions deserve a better payoff. An accurate evaluation of individual providers' contributions can promote fairness in FL. Recent work has been proposed to use the Shapley value in evaluating data's contribution~\cite{liu2020fedcoin,wei2020efficient,wang2020principled,wang2019measure}. However, Shapley value has an exponential time complexity, thus very inefficient.~\cite{wang2022efficient} proposed to directly discard part of the computation of Shapley value, which could cause an error within 5\%. Besides, other approximation methods,~\eg, Monte-Carlo sampling, are designed to improve the computation efficiency~\cite{wei2020efficient, wang2019measure, wang2020principled}. In real-world applications, the federation or model users could run a sandbox simulation to calculate each data provider's contribution. Besides, the contribution evaluation methods are supposed to be robust against the cheating behaviors from the data providers. For example, if the data providers know the contribution evaluating rules,~\eg, the corresponding test dataset, they could generate data with high contribution value using low cost, and the data is useless for the modeling task.

\subsubsection{Evaluation of Cost and Value.}
The accurate assessments of the supply-side data cost and the demand-size model value are essential for the incentive mechanism design. There could be many ways for calculation. A straightforward method is building a computational model based on market research, which might be complicated because it is hard to design one computational model that fits all VFL scenarios. Apart from the computational model design, another solution is the self-report when the cost and value information are privately known. An auction-based self-report could be used in evaluating the supply-side data cost. Specifically, the federation can ask each data owner to request payment for the data contribution and then select which data owner shall join the federation. An example of the auction-based self-report on the data owner side is using the PVCG model~\cite{cong2020game}. The truth-telling self-report of demand-side model value needs the accomplishment of incentive compatibility, such that the model users could get the best outcome if they report the true model value. To achieve incentive compatibility, the authors proposed to find the Nash equilibrium, which is similar to~\cite{lu2022truthful}. Besides,~\cite{zhang2022data} proposed to abstract the data transaction process of VFL to a Stackelberg price competition model. Then, we could conduct a reverse analysis to obtain the optimal data pricing strategy.

\subsubsection{Payoff Sharing Schema Design.}
The payoff sharing schema design determines how much the model users should pay to the data providers. The pay-off sharing schema is non-trivial and challenging because it has many objectives: 1)~\textbf{Incentive compatibility}. A VFL incentive mechanism should ensure that  participants can achieve the best or at least not the worst outcome by reporting the truth regarding the cost and value; 2)~\textbf{Individual rationality}. The mechanism should not make any party worse off than if he quits the federation. In other words, each data provider's payoff is greater than the cost, and each model user's model value is greater than the budget; 3)~\textbf{Budget balance}. Model users' budget could cover the payoff of data providers. A mechanism is called strong budget balanced if budget equals payoff; 4)~\textbf{Social surplus optimization}. Social surplus is the sum of consumer surplus and producer surplus,~\ie, data cost and model value. A mechanism is called social surplus optimized when the social surplus is maximized; 5)~\textbf{Fairness}. The received payoff should be fair among the data providers. Specifically, data with higher contribution to the model or higher cost should be rewarded more payoff. Besides, the fairness of mechanism should also be robust to cheating from the data provider side; 6)~\textbf{Punishing the attacks}. A mechanism should be able to punish malicious parties if any attack is detected; 7)~\textbf{Eliminating payment delay}. A mechanism should try to eliminate payment delay incurred by model training time to prevent budget deficit. For example,  data owner A's data is contributed with high cost but has few improvements for model user B's task. If B pays A until the whole training process is finished, B will likely meet the budget deficit.

Besides, there exist trade-offs among the above objectives. Thus, building one incentive mechanism to optimize all objectives is unrealistic. For example, more optimizing targets usually result in worse social optimization~\cite{cong2020game}. In practice, we could select a subset of targets according to scenarios. Cong~\etal proposed a demand-side Cremer-McLean mechanism to optimize the incentive compatibility and individual rationality together~\cite{cong2020game}.

{\bf (D6) Fair incentive mechanism.} As a technique highly related to business, VFL needs a fair incentive mechanism to ensure equitable payment for each party contributing to the training process, encouraging more new users to join the community. However, the incentive mechanism in VFL is still an open problem.

\vspace{-0.2cm}

\subsection{Interpretability}
VFL is commonly applied in areas (\eg, financial scenarios) that severely need interpretability. Interpretability could improve data quality and help to understand the decision-making process. Therefore, enhancing the interpretability of VFL is also an essential problem. For example, Kang~\etal proposed an explainable VFL approach to help the domain adaptation process~\cite{kang2021privacy}. The features of an auxiliary party to guide the adaptation are split into groups using expertise knowledge in advance. They assign a separate feature extractor for each feature group to obtain interpretable transferable features.~\cite{chen2022evfl} designed an explainable VFL framework based on counterfactual explanation~\cite{mothilal2020explaining} to evaluate feature importance for VFL. 

However, there exist conflicts between interpretability and privacy in VFL. For instance, anonymous features are often utilized in VFL to avoid privacy leakage, which harms the explainability of model. To address this problem,~\cite{chen2021fed} proposed an interpretable inference framework for vertical tree ensemble models. It discloses the feature meanings to other parties but encrypts the decision path for security.

\vspace{-0.2cm}

\subsection{Efficiency Optimization}
Based on the architecture proposed by Yang~\etal~\cite{yang2019federated}, we introduce several innovative works that improve efficiency and effectiveness. They could be summarized as pipeline and compression.
        
\subsubsection{Pipeline.}
Current VFL design suffers from ``Liebig's Law''~\footnote{https://en.wikipedia.org/wiki/Liebig\%27s\_law\_of\_the\_minimum}, that the overall efficiency is limited by slow participants. Besides the asynchronized designs in~\S\ref{subsubsec:categorization}, there exist other pipeline methods to alleviate the efficiency problem. For example, VF2Boost addressed the mutual-waiting problem in VFL via an optimistic node-splitting method~\cite{fu2021vf2boost}. The active party directly splits tree nodes according to his own data without waiting for information from the passive party. A roll-back operation is designed for correcting wrong splits. 
               
\subsubsection{Compression.}
Another method to improve communication efficiency in VFL is compression. For example, Li~\etal proposed a double-end sparse compression method to reduce the additional intermediate results transmission~\cite{li2020efficient}. On both the active and passive parties, only the results with large changes will be transmitted. Moreover, the changes in the results which are not transmitted will be accumulated locally until large enough. Similarly, Yang~\etal adopted the same compression method, which only transmits important updates larger than a given threshold and accumulates the remaining updates locally~\cite{yang2021model}. Besides,~\cite{khan2022communication} adopted PCA and autoencoder to reduce the dimension of transmitted message. From another point of view, according to the properties of HE, VF2Boost designed a polynomial-based packing method to conduct compression. They find that the encoded integer of each value locates in a narrow range, therefore packing multiple ciphers into a single one to reduce the total transmission amount of intermediate results~\cite{fu2021vf2boost}. 
        
%
%
        
\subsection{Robustness Enhancement}
Robustness is another important problem for VFL. We introduce the robustness enhancement approaches in VFL considering three aspects,~\ie, multiple party scenario, dynamic party handling, and multi-tiered network structure.

\subsubsection{Multiple Party Scenario.}
Current researchers mainly consider the two-party VFL setting because it is more common in practice~\cite{chen2021homomorphic}. However, with the development of privacy-preserving techniques~\cite{chauhan2020survey}, the multiple party scenario of VFL is becoming an important topic. Among the existing works,~\cite{sun2022privacy,shi2022mvfls} proposed to utilize the secure aggregation method~\cite{bonawitz2017practical} for intermediate results communication, which could easily suit the multiple party setting. The HE and SS hybrid protocol proposed by~\cite{huang2022efmvfl} also supports flexible expansion to multiple parties.~\cite{feng2020multi} adopts a multi-participant framework based on multi-view learning~\cite{xu2013survey}. Furthermore,~\cite{mugunthan2021multi} introduces a scenario where labels also distribute among multiple parties.~\cite{li2022vfl} proposed a ring architecture for multi-party communication to avoid the complexity of protocol and reduce the effect third party. Besides,~\cite{zeng2022over} proposed a fast global aggregation method based on the over-the-air computation (AirComp) technology~\cite{yang2020federated} for scenarios with a large number of participants. AirComp could utilize the superposition property of wireless multiple access channels to concurrently transmit local intermediate results.

\subsubsection{Dynamic Party Handling.}
The dynamic party is much more common in HFL than in VFL. However, the handling of dynamic parties is also essential when the number of participants in VFL increases. For example,~\cite{zhang2022low} provided a scheduling policy based on the truncation threshold to silence the parties with a low transmission rate, which could reduce communication latency while maintaining convergence. Several works have already invesgated on this topic. Zhou~\etal proposed to utilize SS to solve the participant dropout problem~\cite{zhou2021privacy}. The aggregated messages will only be decrypted if they come from enough participants. Besides, Hou~\etal designed additional handling for both dynamic parties join and dropout problems~\cite{hou2021verifiable}. The secure aggregation method~\cite{bonawitz2017practical} adopted by~\cite{sun2022privacy,shi2022mvfls} could also solve the dropout problems by utilizing SS.

\subsubsection{Hierarchical Network Structure.}
Most existing VFL works only consider a one-tiered network, where the participants maintain data locally. However, in reality, the network structure of VFL could be hierarchical. The participants of VFL also consist of a group of data silos. For instance, two companies want to conduct VFL, and the data of each company is maintained distributedly in sub-companies. Das~\etal designed a novel VFL algorithm for a two-tiered scenario where each party (\ie, hub) in VFL contains a group of clients with horizontally-partitioned data~\cite{das2021multi}. For one iteration, firstly, each hub aggregates model parameters and intermediate results with the corresponding clients. Then, all hubs communicate intermediate results with each other for computing model updates. Finally, the hub broadcast updates to its clients. Besides,~\cite{kang2021privacy} also introduces a novel hierarchical structure where two parties first extend their feature space from an auxiliary party via VFL, and then those two parties could conduct domain adaptation. Valdeira~\etal applied the parallel Markov chain coordinate descent algorithm in a semi-decentralized setting, combining both client-client and client-server communication, to improve efficiency~\cite{valdeira2022multi}. TDCD~\cite{das2022cross} and VHFL~\cite{wan2022global} are proposed for the two-tiered network scenario, where the data is horizontally and vertically partitioned sequentially.




\subsection{Verification}
In real-world scenarios, participants are malicious with a high probability, which conflicts with the semi-honest assumption of most VFL papers. A simple and effective method is to arm the system with verification and audit functions. There are some pilot works on this topic. For instance, Hou~\etal adopted a verifiable computation method proposed by~\cite{backes2013verifiable} to check the data integrity~\cite{hou2021verifiable}. This method could verify if the results generated by messages that parties transmit to the server have been modified. Besides, Zhang~\etal designed a verification framework whose clients are deployed separately at all participants to monitor the transferred messages~\cite{zhang2021aegis}. They define VFL tasks as finite state machines uniformly, therefore the framework could adapt to different VFL algorithms. Furthermore, blockchain~\cite{zheng2018blockchain} could also be used to ensure a trustworthy federated training process~\cite{ur2020towards,dias2023blocklearning,cheng2022vflchain}.

{\bf (D7) Reliable verification.} The malicious adversary in the real world is one of the largest obstacles to adopt VFL. Verification is the easiest and most effective method for defending malicious adversaries. However, there is no uniform solution to verification.

\section{Layer Five: Application Design of VFL}
\label{sec:application_design}
\begin{figure}[h]
	\centering
	\includegraphics[width=0.25\linewidth]{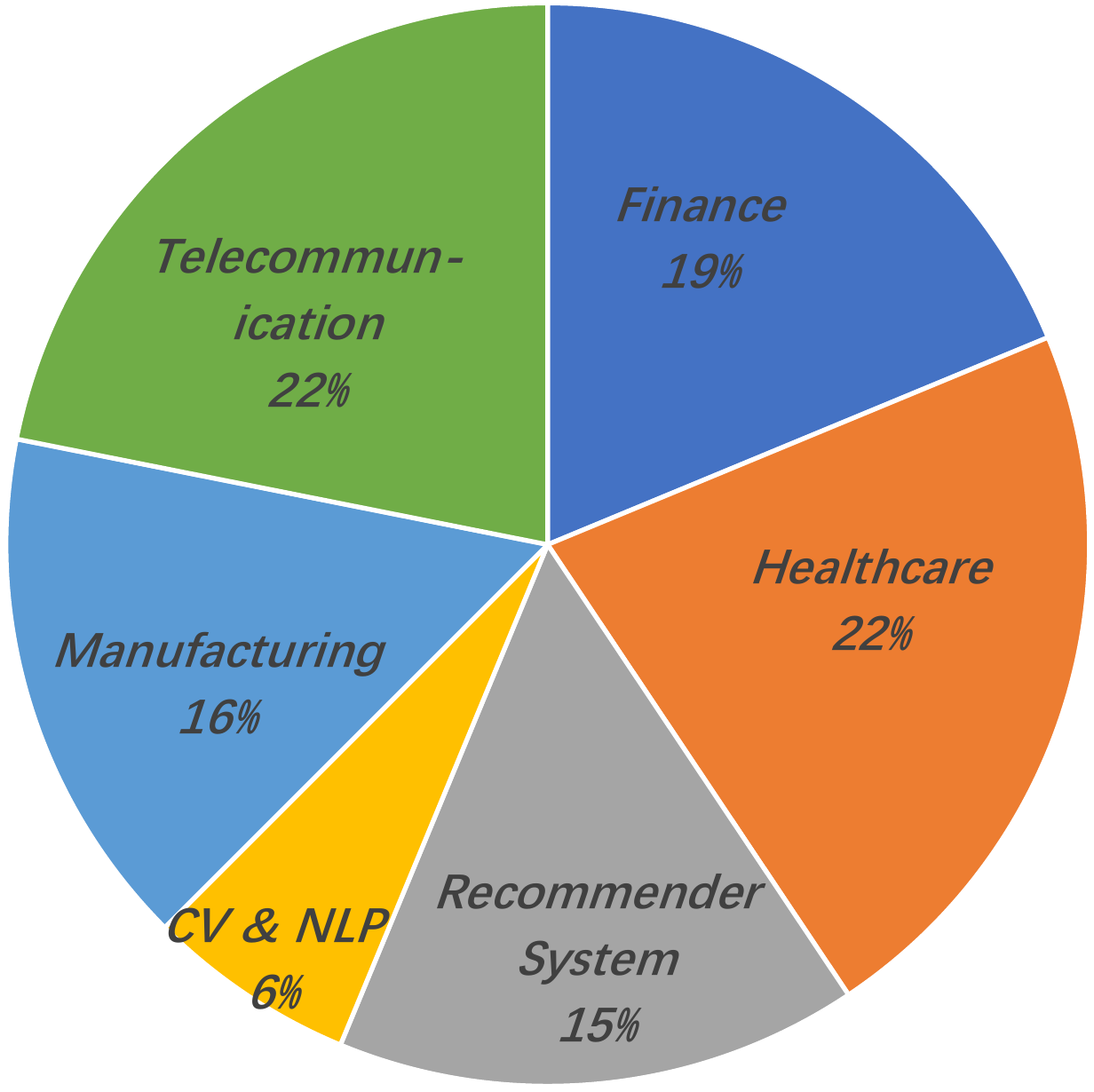}
	\caption{The portions of different application areas of vertical federated learning based on the papers we have investigated. Healthcare and telecommunication are the dominant application areas (22\% for each), while computer vision and language processing area takes the smallest part (6\%).}
	\vspace{-0.6cm}
	\label{fig:application-portion}
\end{figure}

VFL is typically an application-driven technique recently adopted in finance, healthcare, recommender system,~\etc. The portion of each application area is shown in Fig.~\ref{fig:application-portion}, based on the papers we have investigated. In this section, we introduce these applications in detail, including the background, problem, and solution.


\subsection{Finance}
Finance is an important application area of VFL~\cite{gupta2018distributed,ou2020homomorphic,yang2019federated,cheng2021secureboost,li2022nearest,hardy2017private}. In finance, many scenarios,~\eg, risk management~\cite{leo2019machine} and marketing~\cite{ma2020machine}, are driven by big data and machine learning algorithms. In detail, risk management~\cite{leo2019machine} is to make judgments on whether one user can be trusted to issue a loan or credit card. The risk becomes high if the loan is given to people who cannot pay it back. Financial marketing~\cite{ma2020machine} is the process of selecting candidate customers from a large group of people such that the selected candidates are more likely to purchase the related financial products. Machine learning models,~\eg, linear regression~\cite{maulud2020review}, gradient boosting tree~\cite{ke2017lightgbm}, and neural network~\cite{yegnanarayana2009artificial}, are commonly used in these scenarios. Financial institutions usually lack enough user portraits for model training. Therefore, VFL could provide a secure way to utilize rich data features from other companies,~\eg, Internet company.

Among these applications,~\cite{yang2019federated} and~\cite{hardy2017private} proposed vertical federated linear and logistic regression, respectively.~\cite{cheng2021secureboost} proposed vertical federated boosting tree model. They adopt HE to protect the transmitted intermediate results and achieve privacy preservation. More works about vertical linear, tree, and neural network models,~\eg,~\cite{cha2021implementing}, are investigated in Section~\ref{sec:protocol_design}. Besides,~\cite{gupta2018distributed} proposed a split-learning framework for financial applications when financial institutions are short of labeled data and computational resources.~\cite{li2022nearest} proposed to utilize vertical federated GNN in the financial area.~\cite{ou2020homomorphic} proposed a HE-based vertical federated Bayesian learning model for risk management.

\vspace{-0.4cm}

\subsection{Healthcare}
Many researchers investigated the application of VFL in the healthcare area, such as~\cite{pfitzner2021federated,gupta2018distributed,li2016vertical,kim2021vertical,lee2018privacy,errounda2022mobility,zhang2022secure},~\etc. Vertically partitioned data is quite common in healthcare~\cite{pfitzner2021federated}. Due to the unique characteristic of medical information, the leakage of medical data may bring severe issues like social discrimination. Therefore, VFL can be helpful to enrich the data features of patients. For example, multiple medical institutions,~\eg, hospitals, own different records for the same patients. They could securely train better models for healthcare tasks with VFL.

Among these applications,~\cite{li2016vertical} provided a new binary LR method to predict the mortality for breast cancer and myocardial infarction, using genome, electronic health records, and intensive care unit data.~\cite{kim2021vertical} further improved this method and verified its performance on a task that analyzes the association between coronary artery disease and cardiovascular risk factors. Besides,~\cite{gupta2018distributed} proposed a split-learning framework for healthcare applications.~\cite{lee2018privacy} designed a VFL patient similarity learning algorithm, which could be used to construct datasets for cross-silo observational studies, disease surveillance, and clinical trial recruitment. The experiments show that the proposed algorithm is effective in the prediction tasks of five diseases,~\ie, disorders of lipoid metabolism, hypertensive chronic kidney disease,  cardiac dysrhythmias, heart failure, and acute renal failure. Besides, MVFF~\cite{errounda2022mobility} is proposed as a mobility forecasting VFL method, which could be used to predict the spread of COVID-19~\cite{ilin2021public}.~\cite{zhang2022secure} designed the VFL feature selection algorithm to improve deep learning models' accuracy in electronic health (eHealth) systems~\cite{evans2016electronic}.

\subsection{Recommender System}
The recommender system (RecSys) is another popular application scenario of VFL~\cite{yang2020federated,atarashi2021vertical,tan2020federated,song2021federated,chen2020secure}. RecSys is widely adopted in people's daily life. For example, we can see our favorite items and articles when we shop and read news online via RecSys. The core of RecSys is to predict whether a user is interested in one specific item based on the side information of users and items as well as users' historical behaviors~\cite{lu2015recommender}. The prediction will have better performance if more training data is provided. Hence, VFL could be a feasible solution considering the privacy concerns and data silo problem.

The vertical federated recommender (FedRec) system refers to the scenarios where multiple parties hold the historical behaviors or side information for the same group of users~\cite{yang2020federated}. For instance, an online library service provider with users' reading history can cooperate with the online movie watching company, which contains watching history to enrich the data features and obtain better recommendation accuracy.~\cite{tan2020federated} proposed an open-source FedRec system, providing the vertical federared implementation of many recommendation algorithms,~\eg, matrix factorization~\cite{koren2009matrix}, factorization machine~\cite{rendle2010factorization}, and wide \& deep learning~\cite{cheng2016wide}. The system has been deployed for a real-world recommendation application and achieves performance improvement. Besides,~\cite{atarashi2021vertical} proposed a vertical higher-order factorization machine to capture high-order feature interactions among data silos.~\cite{song2021federated} provided a SecureBoost-based system to improve recommendation accuracy with data in mobile network operators and healthcare providers.~\cite{chen2020secure} designed an MPC-based vertical federated social recommendation framework.



%

\vspace{-0.3cm}        

\subsection{Computer Vision and Natural Language Processing}

Apart from the recommender system in the Internet area, VFL is also explored in the computer vision~\cite{forsyth2011computer} and natural language processing~\cite{chowdhary2020natural} area. For instance,~\cite{liu2020federated} proposed a federated learning framework to fuse various types of image representations from different tasks to form fine-grained image representations. Both visual and textual features are contained in these tasks. By generating fine-grained representations, their framework improves performance on various vision and language grounding problems without sharing downstream task data.~\cite{li2022adaptive} proposed a vertical federated CNN algorithm for the image classification task, where parties hold incomplete pieces for the same group of images. The accuracy of designed VFL method is close to the one of centralized method.


%

 
\vspace{-0.3cm}

\subsection{Manufacturing}

The applications of VFL in the field of manufacturing contain~\cite{aggour2019federated,ge2021failure,wu2022privacy,mao2023full,xia2022privacy}. The manufacturing process usually crosses different companies, workshops, or even production lines \cite{ge2021failure}. Due to the privacy-preserving regulations, manufacturing data could not be collected together to train models, which becomes the bottleneck of intelligent manufacturing. Ning~\etal proposed the vertical federated SVD and random forest algorithm for failure prediction in the production line~\cite{ge2021failure}. The proposed models are validated on public datasets in intelligent manufacturing. Xia~\etal further applied a vertical federated XGBoost algorithm in the task of failure prediction~\cite{xia2022privacy}. In addition,~\cite{aggour2019federated} designed a federated multimodal learning platform for additive manufacturing~\cite{frazier2014metal}. Wu~\etal proposed a federated frequent itemsets mining framework for industrial collaborative activities of the supply chain network~\cite{wu2022privacy}. Besides,~\cite{mao2023full} applied vertical federated XGBoost in the power grid area to power load forecasting, which is vital for the stability of power systems and reduction of cost.

        
\subsection{Telecommunication}
VFL has also been applied in the telecommunication area~\cite{subramanya2021centralized,zhang2020vertical,hashemi2021vertical,sacco2020federated,zhang2022low,zeng2022over,liu2022vertical}. In telecommunication, conventional centralized learning  requires data from multiple data sources, which could leak the private data of each party. Thus, VFL is adopted to support telecommunication in different aspects. For example, Tejas~\etal proposed a vertical autoscaling algorithm for intelligent end-to-end management and orchestration of network resources in the multi-domain 5G networks~\cite{subramanya2021centralized}.~\cite{zeng2022over} provided a communication-computation co-design for the 6G network~\cite{letaief2021edge}.~\cite{zhang2020vertical} designed a VFL-based cooperative sensing scheme for cognitive radio networks. Besides,~\cite{hashemi2021vertical} proposed a VFL algorithm for collaborative model training in disaggregated networks. The algorithm is used to classify the quality of transmission. In addition,~\cite{sacco2020federated} proposed a VFL architecture for routing packets in the challenged network with varying conditions and high-volumne traffic.~\cite{zhang2022low} designed a low-latency VFL algorithm to improve the accuracy of spectrum sensing~\cite{hussain2009spectrum}, which is used in the area of wireless communication. Besides,~\cite{liu2022vertical} applied VFL in the task of motion recognition to exploit the multi-view observation data from different wireless sensing devices.

\vspace{-0.3cm}


        


\section{Conclusion}
In this paper, we show the current development situation of vertical federated learning. From a layered perspective, we discuss the current works and explore the challenges of designing a vertical federated learning framework for all layers. Although the flourishing growth momentum has appeared in different application areas recently, the research of vertical federated learning is still in its infancy. As a technique full of commercial values, vertical federated learning is a topic that deserves to be studied.

\bibliographystyle{ACM-Reference-Format}
\bibliography{VFLSurvey}


\begin{thebibliography}{246}


\ifx \showCODEN    \undefined \def \showCODEN     #1{\unskip}     \fi
\ifx \showDOI      \undefined \def \showDOI       #1{#1}\fi
\ifx \showISBNx    \undefined \def \showISBNx     #1{\unskip}     \fi
\ifx \showISBNxiii \undefined \def \showISBNxiii  #1{\unskip}     \fi
\ifx \showISSN     \undefined \def \showISSN      #1{\unskip}     \fi
\ifx \showLCCN     \undefined \def \showLCCN      #1{\unskip}     \fi
\ifx \shownote     \undefined \def \shownote      #1{#1}          \fi
\ifx \showarticletitle \undefined \def \showarticletitle #1{#1}   \fi
\ifx \showURL      \undefined \def \showURL       {\relax}        \fi
\providecommand\bibfield[2]{#2}
\providecommand\bibinfo[2]{#2}
\providecommand\natexlab[1]{#1}
\providecommand\showeprint[2][]{arXiv:#2}

\bibitem[Acar et~al\mbox{.}(2018)]%
        {acar2018survey}
\bibfield{author}{\bibinfo{person}{Abbas Acar}, \bibinfo{person}{Hidayet Aksu},
  \bibinfo{person}{A~Selcuk Uluagac}, {and} \bibinfo{person}{Mauro Conti}.}
  \bibinfo{year}{2018}\natexlab{}.
\newblock \showarticletitle{A survey on homomorphic encryption schemes: Theory
  and implementation}.
\newblock \bibinfo{journal}{\emph{ACM Computing Surveys (CSUR)}}
  \bibinfo{volume}{51}, \bibinfo{number}{4} (\bibinfo{year}{2018}),
  \bibinfo{pages}{1--35}.
\newblock


\bibitem[Aggour et~al\mbox{.}(2019)]%
        {aggour2019federated}
\bibfield{author}{\bibinfo{person}{Kareem~S Aggour}, \bibinfo{person}{Vijay~S
  Kumar}, \bibinfo{person}{Paul Cuddihy}, \bibinfo{person}{Jenny~Weisenberg
  Williams}, \bibinfo{person}{Vipul Gupta}, \bibinfo{person}{Laura Dial},
  \bibinfo{person}{Tim Hanlon}, \bibinfo{person}{Justin Gambone}, {and}
  \bibinfo{person}{Joseph Vinciquerra}.} \bibinfo{year}{2019}\natexlab{}.
\newblock \showarticletitle{Federated multimodal big data storage \& analytics
  platform for additive manufacturing}. In \bibinfo{booktitle}{\emph{2019 IEEE
  international conference on big data (big data)}}. IEEE,
  \bibinfo{pages}{1729--1738}.
\newblock


\bibitem[Atarashi and Ishihata(2021)]%
        {atarashi2021vertical}
\bibfield{author}{\bibinfo{person}{Kyohei Atarashi} {and}
  \bibinfo{person}{Masakazu Ishihata}.} \bibinfo{year}{2021}\natexlab{}.
\newblock \showarticletitle{Vertical Federated Learning for Higher-Order
  Factorization Machines.}. In \bibinfo{booktitle}{\emph{PAKDD (2)}}.
  \bibinfo{pages}{346--357}.
\newblock


\bibitem[Backes et~al\mbox{.}(2013)]%
        {backes2013verifiable}
\bibfield{author}{\bibinfo{person}{Michael Backes}, \bibinfo{person}{Dario
  Fiore}, {and} \bibinfo{person}{Raphael~M Reischuk}.}
  \bibinfo{year}{2013}\natexlab{}.
\newblock \showarticletitle{Verifiable delegation of computation on outsourced
  data}. In \bibinfo{booktitle}{\emph{Proceedings of the 2013 ACM SIGSAC
  conference on Computer \& communications security}}.
  \bibinfo{pages}{863--874}.
\newblock


\bibitem[Barker et~al\mbox{.}(2017)]%
        {barker2017recommendation}
\bibfield{author}{\bibinfo{person}{Elaine Barker}, \bibinfo{person}{Lily Chen},
  \bibinfo{person}{Sharon Keller}, \bibinfo{person}{Allen Roginsky},
  \bibinfo{person}{Apostol Vassilev}, {and} \bibinfo{person}{Richard Davis}.}
  \bibinfo{year}{2017}\natexlab{}.
\newblock \bibinfo{booktitle}{\emph{Recommendation for pair-wise
  key-establishment schemes using discrete logarithm cryptography}}.
\newblock \bibinfo{type}{{T}echnical {R}eport}. \bibinfo{institution}{National
  Institute of Standards and Technology}.
\newblock


\bibitem[Beimel(2011)]%
        {beimel2011secret}
\bibfield{author}{\bibinfo{person}{Amos Beimel}.}
  \bibinfo{year}{2011}\natexlab{}.
\newblock \showarticletitle{Secret-sharing schemes: A survey}. In
  \bibinfo{booktitle}{\emph{International conference on coding and
  cryptology}}. Springer, \bibinfo{pages}{11--46}.
\newblock


\bibitem[Blondel et~al\mbox{.}(2016)]%
        {blondel2016higher}
\bibfield{author}{\bibinfo{person}{Mathieu Blondel}, \bibinfo{person}{Akinori
  Fujino}, \bibinfo{person}{Naonori Ueda}, {and} \bibinfo{person}{Masakazu
  Ishihata}.} \bibinfo{year}{2016}\natexlab{}.
\newblock \showarticletitle{Higher-order factorization machines}.
\newblock \bibinfo{journal}{\emph{Advances in Neural Information Processing
  Systems}}  \bibinfo{volume}{29} (\bibinfo{year}{2016}).
\newblock


\bibitem[Bonawitz et~al\mbox{.}(2017)]%
        {bonawitz2017practical}
\bibfield{author}{\bibinfo{person}{Keith Bonawitz}, \bibinfo{person}{Vladimir
  Ivanov}, \bibinfo{person}{Ben Kreuter}, \bibinfo{person}{Antonio Marcedone},
  \bibinfo{person}{H~Brendan McMahan}, \bibinfo{person}{Sarvar Patel},
  \bibinfo{person}{Daniel Ramage}, \bibinfo{person}{Aaron Segal}, {and}
  \bibinfo{person}{Karn Seth}.} \bibinfo{year}{2017}\natexlab{}.
\newblock \showarticletitle{Practical secure aggregation for privacy-preserving
  machine learning}. In \bibinfo{booktitle}{\emph{proceedings of the 2017 ACM
  SIGSAC Conference on Computer and Communications Security}}.
  \bibinfo{pages}{1175--1191}.
\newblock


\bibitem[Brown et~al\mbox{.}(2020)]%
        {brown2020language}
\bibfield{author}{\bibinfo{person}{Tom Brown}, \bibinfo{person}{Benjamin Mann},
  \bibinfo{person}{Nick Ryder}, \bibinfo{person}{Melanie Subbiah},
  \bibinfo{person}{Jared~D Kaplan}, \bibinfo{person}{Prafulla Dhariwal},
  \bibinfo{person}{Arvind Neelakantan}, \bibinfo{person}{Pranav Shyam},
  \bibinfo{person}{Girish Sastry}, \bibinfo{person}{Amanda Askell},
  {et~al\mbox{.}}} \bibinfo{year}{2020}\natexlab{}.
\newblock \showarticletitle{Language models are few-shot learners}.
\newblock \bibinfo{journal}{\emph{Advances in neural information processing
  systems}}  \bibinfo{volume}{33} (\bibinfo{year}{2020}),
  \bibinfo{pages}{1877--1901}.
\newblock


\bibitem[Cai et~al\mbox{.}(2022b)]%
        {cai2022efficient}
\bibfield{author}{\bibinfo{person}{Jianping Cai}, \bibinfo{person}{Ximeng Liu},
  \bibinfo{person}{Zhiyong Yu}, \bibinfo{person}{Kun Guo}, {and}
  \bibinfo{person}{Jiayin Li}.} \bibinfo{year}{2022}\natexlab{b}.
\newblock \showarticletitle{Efficient Vertical Federated Learning Method for
  Ridge Regression of Large-Scale Samples}.
\newblock \bibinfo{journal}{\emph{IEEE Transactions on Emerging Topics in
  Computing}} (\bibinfo{year}{2022}).
\newblock


\bibitem[Cai et~al\mbox{.}(2022a)]%
        {cai2022secure}
\bibfield{author}{\bibinfo{person}{Shuowei Cai}, \bibinfo{person}{Di Chai},
  \bibinfo{person}{Liu Yang}, \bibinfo{person}{Junxue Zhang},
  \bibinfo{person}{Yilun Jin}, \bibinfo{person}{Leye Wang},
  \bibinfo{person}{Kun Guo}, {and} \bibinfo{person}{Kai Chen}.}
  \bibinfo{year}{2022}\natexlab{a}.
\newblock \showarticletitle{Secure Forward Aggregation for Vertical Federated
  Neural Networks}.
\newblock \bibinfo{journal}{\emph{arXiv preprint arXiv:2207.00165}}
  (\bibinfo{year}{2022}).
\newblock


\bibitem[Cardwell et~al\mbox{.}(2017)]%
        {cardwell2017bbr}
\bibfield{author}{\bibinfo{person}{Neal Cardwell}, \bibinfo{person}{Yuchung
  Cheng}, \bibinfo{person}{C~Stephen Gunn}, \bibinfo{person}{Soheil~Hassas
  Yeganeh}, {and} \bibinfo{person}{Van Jacobson}.}
  \bibinfo{year}{2017}\natexlab{}.
\newblock \showarticletitle{BBR: congestion-based congestion control}.
\newblock \bibinfo{journal}{\emph{Commun. ACM}} \bibinfo{volume}{60},
  \bibinfo{number}{2} (\bibinfo{year}{2017}), \bibinfo{pages}{58--66}.
\newblock


\bibitem[Cha et~al\mbox{.}(2021)]%
        {cha2021implementing}
\bibfield{author}{\bibinfo{person}{Dongchul Cha}, \bibinfo{person}{MinDong
  Sung}, \bibinfo{person}{Yu-Rang Park}, {et~al\mbox{.}}}
  \bibinfo{year}{2021}\natexlab{}.
\newblock \showarticletitle{Implementing Vertical Federated Learning Using
  Autoencoders: Practical Application, Generalizability, and Utility Study}.
\newblock \bibinfo{journal}{\emph{JMIR Medical Informatics}}
  \bibinfo{volume}{9}, \bibinfo{number}{6} (\bibinfo{year}{2021}),
  \bibinfo{pages}{e26598}.
\newblock


\bibitem[Chai et~al\mbox{.}(2021)]%
        {chai2021federated}
\bibfield{author}{\bibinfo{person}{Di Chai}, \bibinfo{person}{Leye Wang},
  \bibinfo{person}{Lianzhi Fu}, \bibinfo{person}{Junxue Zhang},
  \bibinfo{person}{Kai Chen}, {and} \bibinfo{person}{Qiang Yang}.}
  \bibinfo{year}{2021}\natexlab{}.
\newblock \showarticletitle{Federated singular vector decomposition}.
\newblock \bibinfo{journal}{\emph{arXiv preprint arXiv:2105.08925}}
  (\bibinfo{year}{2021}).
\newblock


\bibitem[Chauhan et~al\mbox{.}(2020)]%
        {chauhan2020survey}
\bibfield{author}{\bibinfo{person}{Aman~Singh Chauhan},
  \bibinfo{person}{Dikshika Rani}, \bibinfo{person}{Akash Kumar},
  \bibinfo{person}{Rishabh Gupta}, {and} \bibinfo{person}{Ashutosh~Kumar
  Singh}.} \bibinfo{year}{2020}\natexlab{}.
\newblock \showarticletitle{A survey on privacy-preserving outsourced data on
  cloud with multiple data providers}. In \bibinfo{booktitle}{\emph{Proceedings
  of the International Conference on Innovative Computing \& Communications
  (ICICC)}}.
\newblock


\bibitem[Chen et~al\mbox{.}(2020b)]%
        {chen2020secure}
\bibfield{author}{\bibinfo{person}{Chaochao Chen}, \bibinfo{person}{Liang Li},
  \bibinfo{person}{Bingzhe Wu}, \bibinfo{person}{Cheng Hong},
  \bibinfo{person}{Li Wang}, {and} \bibinfo{person}{Jun Zhou}.}
  \bibinfo{year}{2020}\natexlab{b}.
\newblock \showarticletitle{Secure Social Recommendation Based on Secret
  Sharing}. In \bibinfo{booktitle}{\emph{24th European Conference on Artificial
  Intelligence (ECAI 20)}}. \bibinfo{publisher}{IOS Press},
  \bibinfo{pages}{506--512}.
\newblock


\bibitem[Chen et~al\mbox{.}(2021a)]%
        {chen2021homomorphic}
\bibfield{author}{\bibinfo{person}{Chaochao Chen}, \bibinfo{person}{Jun Zhou},
  \bibinfo{person}{Li Wang}, \bibinfo{person}{Xibin Wu},
  \bibinfo{person}{Wenjing Fang}, \bibinfo{person}{Jin Tan},
  \bibinfo{person}{Lei Wang}, \bibinfo{person}{Alex~X Liu},
  \bibinfo{person}{Hao Wang}, {and} \bibinfo{person}{Cheng Hong}.}
  \bibinfo{year}{2021}\natexlab{a}.
\newblock \showarticletitle{When homomorphic encryption marries secret sharing:
  Secure large-scale sparse logistic regression and applications in risk
  control}. In \bibinfo{booktitle}{\emph{Proceedings of the 27th ACM SIGKDD
  Conference on Knowledge Discovery \& Data Mining}}.
  \bibinfo{pages}{2652--2662}.
\newblock


\bibitem[Chen et~al\mbox{.}(2022b)]%
        {chen2022graph}
\bibfield{author}{\bibinfo{person}{Jinyin Chen}, \bibinfo{person}{Guohan
  Huang}, \bibinfo{person}{Haibin Zheng}, \bibinfo{person}{Shanqing Yu},
  \bibinfo{person}{Wenrong Jiang}, {and} \bibinfo{person}{Chen Cui}.}
  \bibinfo{year}{2022}\natexlab{b}.
\newblock \showarticletitle{Graph-Fraudster: Adversarial Attacks on Graph
  Neural Network Based Vertical Federated Learning}.
\newblock \bibinfo{journal}{\emph{IEEE Transactions on Computational Social
  Systems}} (\bibinfo{year}{2022}).
\newblock


\bibitem[Chen et~al\mbox{.}(2022a)]%
        {chen2022evfl}
\bibfield{author}{\bibinfo{person}{Peng Chen}, \bibinfo{person}{Xin Du},
  \bibinfo{person}{Zhihui Lu}, \bibinfo{person}{Jie Wu}, {and}
  \bibinfo{person}{Patrick~CK Hung}.} \bibinfo{year}{2022}\natexlab{a}.
\newblock \showarticletitle{EVFL: An explainable vertical federated learning
  for data-oriented Artificial Intelligence systems}.
\newblock \bibinfo{journal}{\emph{Journal of Systems Architecture}}
  (\bibinfo{year}{2022}), \bibinfo{pages}{102474}.
\newblock


\bibitem[Chen and Guestrin(2016)]%
        {chen2016xgboost}
\bibfield{author}{\bibinfo{person}{Tianqi Chen} {and} \bibinfo{person}{Carlos
  Guestrin}.} \bibinfo{year}{2016}\natexlab{}.
\newblock \showarticletitle{Xgboost: A scalable tree boosting system}. In
  \bibinfo{booktitle}{\emph{Proceedings of the 22nd acm sigkdd international
  conference on knowledge discovery and data mining}}.
  \bibinfo{pages}{785--794}.
\newblock


\bibitem[Chen et~al\mbox{.}(2020a)]%
        {chen2020vafl}
\bibfield{author}{\bibinfo{person}{Tianyi Chen}, \bibinfo{person}{Xiao Jin},
  \bibinfo{person}{Yuejiao Sun}, {and} \bibinfo{person}{Wotao Yin}.}
  \bibinfo{year}{2020}\natexlab{a}.
\newblock \showarticletitle{Vafl: a method of vertical asynchronous federated
  learning}. In \bibinfo{booktitle}{\emph{Proceedings of ICML Workshop on
  Federated Learning for User Privacy and Data Confidentiality}}.
\newblock


\bibitem[Chen et~al\mbox{.}(2021b)]%
        {chen2021fed}
\bibfield{author}{\bibinfo{person}{Xiaolin Chen}, \bibinfo{person}{Shuai Zhou},
  \bibinfo{person}{Kai Yang}, \bibinfo{person}{Hao Fan}, \bibinfo{person}{Zejin
  Feng}, \bibinfo{person}{Zhong Chen}, \bibinfo{person}{Hu Wang}, {and}
  \bibinfo{person}{Yongji Wang}.} \bibinfo{year}{2021}\natexlab{b}.
\newblock \showarticletitle{Fed-EINI: An Efficient and Interpretable Inference
  Framework for Decision Tree Ensembles in Federated Learning}.
\newblock \bibinfo{journal}{\emph{arXiv preprint arXiv:2105.09540}}
  (\bibinfo{year}{2021}).
\newblock


\bibitem[Cheng et~al\mbox{.}(2016)]%
        {cheng2016wide}
\bibfield{author}{\bibinfo{person}{Heng-Tze Cheng}, \bibinfo{person}{Levent
  Koc}, \bibinfo{person}{Jeremiah Harmsen}, \bibinfo{person}{Tal Shaked},
  \bibinfo{person}{Tushar Chandra}, \bibinfo{person}{Hrishi Aradhye},
  \bibinfo{person}{Glen Anderson}, \bibinfo{person}{Greg Corrado},
  \bibinfo{person}{Wei Chai}, \bibinfo{person}{Mustafa Ispir}, {et~al\mbox{.}}}
  \bibinfo{year}{2016}\natexlab{}.
\newblock \showarticletitle{Wide \& deep learning for recommender systems}. In
  \bibinfo{booktitle}{\emph{Proceedings of the 1st workshop on deep learning
  for recommender systems}}. \bibinfo{pages}{7--10}.
\newblock


\bibitem[Cheng et~al\mbox{.}(2021a)]%
        {cheng2021secureboost}
\bibfield{author}{\bibinfo{person}{Kewei Cheng}, \bibinfo{person}{Tao Fan},
  \bibinfo{person}{Yilun Jin}, \bibinfo{person}{Yang Liu},
  \bibinfo{person}{Tianjian Chen}, \bibinfo{person}{Dimitrios Papadopoulos},
  {and} \bibinfo{person}{Qiang Yang}.} \bibinfo{year}{2021}\natexlab{a}.
\newblock \showarticletitle{Secureboost: A lossless federated learning
  framework}.
\newblock \bibinfo{journal}{\emph{IEEE Intelligent Systems}}
  (\bibinfo{year}{2021}).
\newblock


\bibitem[Cheng et~al\mbox{.}(2021b)]%
        {cheng2021haflo}
\bibfield{author}{\bibinfo{person}{Xiaodian Cheng}, \bibinfo{person}{Wanhang
  Lu}, \bibinfo{person}{Xinyang Huang}, \bibinfo{person}{Shuihai Hu}, {and}
  \bibinfo{person}{Kai Chen}.} \bibinfo{year}{2021}\natexlab{b}.
\newblock \showarticletitle{HAFLO: GPU-Based Acceleration for Federated
  Logistic Regression}. In \bibinfo{booktitle}{\emph{International Workshop on
  Federated Learning for User Privacy and Data Confidentiality in Conjunction
  with IJCAI 2021 (FL-IJCAI'21)}}.
\newblock


\bibitem[Cheng et~al\mbox{.}(2022)]%
        {cheng2022vflchain}
\bibfield{author}{\bibinfo{person}{Ziwen Cheng}, \bibinfo{person}{Yongqi Pan},
  \bibinfo{person}{Yi Liu}, \bibinfo{person}{Bowen Wang}, \bibinfo{person}{Xin
  Deng}, {and} \bibinfo{person}{Cheng Zhu}.} \bibinfo{year}{2022}\natexlab{}.
\newblock \showarticletitle{VFLChain: Blockchain-enabled Vertical Federated
  Learning for Edge Network Data Sharing}. In \bibinfo{booktitle}{\emph{2022
  IEEE International Conference on Unmanned Systems (ICUS)}}. IEEE,
  \bibinfo{pages}{606--611}.
\newblock


\bibitem[Cheon et~al\mbox{.}(2017)]%
        {cheon2017homomorphic}
\bibfield{author}{\bibinfo{person}{Jung~Hee Cheon}, \bibinfo{person}{Andrey
  Kim}, \bibinfo{person}{Miran Kim}, {and} \bibinfo{person}{Yongsoo Song}.}
  \bibinfo{year}{2017}\natexlab{}.
\newblock \showarticletitle{Homomorphic encryption for arithmetic of
  approximate numbers}. In \bibinfo{booktitle}{\emph{International Conference
  on the Theory and Application of Cryptology and Information Security}}.
  Springer, \bibinfo{pages}{409--437}.
\newblock


\bibitem[Cheung et~al\mbox{.}(2021)]%
        {cheung2021fedsgc}
\bibfield{author}{\bibinfo{person}{Tsz-Him Cheung}, \bibinfo{person}{Weihang
  Dai}, {and} \bibinfo{person}{Shuhan Li}.} \bibinfo{year}{2021}\natexlab{}.
\newblock \showarticletitle{FedSGC: Federated Simple Graph Convolution for Node
  Classification}. In \bibinfo{booktitle}{\emph{IJCAI Workshop on Federated and
  Transfer Learning for Data Sparsity and Confidentiality}}.
\newblock


\bibitem[Cheung et~al\mbox{.}(2022)]%
        {cheung2022vertical}
\bibfield{author}{\bibinfo{person}{Yiu-ming Cheung}, \bibinfo{person}{Juyong
  Jiang}, \bibinfo{person}{Feng Yu}, {and} \bibinfo{person}{Jian Lou}.}
  \bibinfo{year}{2022}\natexlab{}.
\newblock \showarticletitle{Vertical Federated Principal Component Analysis and
  Its Kernel Extension on Feature-wise Distributed Data}.
\newblock \bibinfo{journal}{\emph{arXiv preprint arXiv:2203.01752}}
  (\bibinfo{year}{2022}).
\newblock


\bibitem[Chowdhary(2020)]%
        {chowdhary2020natural}
\bibfield{author}{\bibinfo{person}{KR1442 Chowdhary}.}
  \bibinfo{year}{2020}\natexlab{}.
\newblock \showarticletitle{Natural language processing}.
\newblock \bibinfo{journal}{\emph{Fundamentals of artificial intelligence}}
  (\bibinfo{year}{2020}), \bibinfo{pages}{603--649}.
\newblock


\bibitem[Chu and Zhang(2021)]%
        {chu2021privacy}
\bibfield{author}{\bibinfo{person}{Kai-Fung Chu} {and} \bibinfo{person}{Lintao
  Zhang}.} \bibinfo{year}{2021}\natexlab{}.
\newblock \showarticletitle{Privacy-Preserving Self-Taught Federated Learning
  for Heterogeneous Data}.
\newblock \bibinfo{journal}{\emph{arXiv preprint arXiv:2102.05883}}
  (\bibinfo{year}{2021}).
\newblock


\bibitem[Cong et~al\mbox{.}(2020)]%
        {cong2020game}
\bibfield{author}{\bibinfo{person}{Mingshu Cong}, \bibinfo{person}{Han Yu},
  \bibinfo{person}{Xi Weng}, {and} \bibinfo{person}{Siu~Ming Yiu}.}
  \bibinfo{year}{2020}\natexlab{}.
\newblock \showarticletitle{A Game-Theoretic Framework for Incentive Mechanism
  Design in Federated Learning}.
\newblock In \bibinfo{booktitle}{\emph{Federated Learning}}.
  \bibinfo{publisher}{Springer}, \bibinfo{pages}{205--222}.
\newblock


\bibitem[Cormode et~al\mbox{.}(2018)]%
        {cormode2018privacy}
\bibfield{author}{\bibinfo{person}{Graham Cormode}, \bibinfo{person}{Somesh
  Jha}, \bibinfo{person}{Tejas Kulkarni}, \bibinfo{person}{Ninghui Li},
  \bibinfo{person}{Divesh Srivastava}, {and} \bibinfo{person}{Tianhao Wang}.}
  \bibinfo{year}{2018}\natexlab{}.
\newblock \showarticletitle{Privacy at scale: Local differential privacy in
  practice}. In \bibinfo{booktitle}{\emph{Proceedings of the 2018 International
  Conference on Management of Data}}. \bibinfo{pages}{1655--1658}.
\newblock


\bibitem[Costan and Devadas(2016)]%
        {costan2016intel}
\bibfield{author}{\bibinfo{person}{Victor Costan} {and}
  \bibinfo{person}{Srinivas Devadas}.} \bibinfo{year}{2016}\natexlab{}.
\newblock \showarticletitle{Intel SGX explained}.
\newblock \bibinfo{journal}{\emph{Cryptology ePrint Archive}}
  (\bibinfo{year}{2016}).
\newblock


\bibitem[Covington et~al\mbox{.}(2016)]%
        {covington2016deep}
\bibfield{author}{\bibinfo{person}{Paul Covington}, \bibinfo{person}{Jay
  Adams}, {and} \bibinfo{person}{Emre Sargin}.}
  \bibinfo{year}{2016}\natexlab{}.
\newblock \showarticletitle{Deep neural networks for youtube recommendations}.
  In \bibinfo{booktitle}{\emph{Proceedings of the 10th ACM conference on
  recommender systems}}. \bibinfo{pages}{191--198}.
\newblock


\bibitem[Cramer et~al\mbox{.}(2015)]%
        {cramer2015secure}
\bibfield{author}{\bibinfo{person}{Ronald Cramer}, \bibinfo{person}{Ivan~Bjerre
  Damg{\aa}rd}, {et~al\mbox{.}}} \bibinfo{year}{2015}\natexlab{}.
\newblock \bibinfo{booktitle}{\emph{Secure multiparty computation}}.
\newblock \bibinfo{publisher}{Cambridge University Press}.
\newblock


\bibitem[Cristofaro and Tsudik(2010)]%
        {cristofaro2010practical}
\bibfield{author}{\bibinfo{person}{Emiliano~De Cristofaro} {and}
  \bibinfo{person}{Gene Tsudik}.} \bibinfo{year}{2010}\natexlab{}.
\newblock \showarticletitle{Practical private set intersection protocols with
  linear complexity}. In \bibinfo{booktitle}{\emph{International Conference on
  Financial Cryptography and Data Security}}. Springer,
  \bibinfo{pages}{143--159}.
\newblock


\bibitem[Dai et~al\mbox{.}(2022)]%
        {dai2022edge}
\bibfield{author}{\bibinfo{person}{Mingjun Dai}, \bibinfo{person}{Ziying
  Zheng}, \bibinfo{person}{Zhaoyan Hong}, \bibinfo{person}{Shengli Zhang},
  {and} \bibinfo{person}{Hui Wang}.} \bibinfo{year}{2022}\natexlab{}.
\newblock \showarticletitle{Edge Computing-Aided Coded Vertical Federated
  Linear Regression}.
\newblock \bibinfo{journal}{\emph{IEEE Transactions on Cognitive Communications
  and Networking}} \bibinfo{volume}{8}, \bibinfo{number}{3}
  (\bibinfo{year}{2022}), \bibinfo{pages}{1543--1551}.
\newblock


\bibitem[Dalskov et~al\mbox{.}(2021)]%
        {dalskov2021fantastic}
\bibfield{author}{\bibinfo{person}{Anders Dalskov}, \bibinfo{person}{Daniel
  Escudero}, {and} \bibinfo{person}{Marcel Keller}.}
  \bibinfo{year}{2021}\natexlab{}.
\newblock \showarticletitle{Fantastic
  Four:$\{$Honest-Majority$\}$$\{$Four-Party$\}$ Secure Computation With
  Malicious Security}. In \bibinfo{booktitle}{\emph{30th USENIX Security
  Symposium (USENIX Security 21)}}. \bibinfo{pages}{2183--2200}.
\newblock


\bibitem[Dang et~al\mbox{.}(2020)]%
        {dang2020large}
\bibfield{author}{\bibinfo{person}{Zhiyuan Dang}, \bibinfo{person}{Bin Gu},
  {and} \bibinfo{person}{Heng Huang}.} \bibinfo{year}{2020}\natexlab{}.
\newblock \showarticletitle{Large-Scale Kernel Method for Vertical Federated
  Learning}.
\newblock In \bibinfo{booktitle}{\emph{Federated Learning}}.
  \bibinfo{publisher}{Springer}, \bibinfo{pages}{66--80}.
\newblock


\bibitem[Das et~al\mbox{.}(2022)]%
        {das2022cross}
\bibfield{author}{\bibinfo{person}{Anirban Das}, \bibinfo{person}{Timothy
  Castiglia}, \bibinfo{person}{Shiqiang Wang}, {and} \bibinfo{person}{Stacy
  Patterson}.} \bibinfo{year}{2022}\natexlab{}.
\newblock \showarticletitle{Cross-silo federated learning for multi-tier
  networks with vertical and horizontal data partitioning}.
\newblock \bibinfo{journal}{\emph{ACM Transactions on Intelligent Systems and
  Technology (TIST)}} \bibinfo{volume}{13}, \bibinfo{number}{6}
  (\bibinfo{year}{2022}), \bibinfo{pages}{1--27}.
\newblock


\bibitem[Das and Patterson(2021)]%
        {das2021multi}
\bibfield{author}{\bibinfo{person}{Anirban Das} {and} \bibinfo{person}{Stacy
  Patterson}.} \bibinfo{year}{2021}\natexlab{}.
\newblock \showarticletitle{Multi-tier federated learning for vertically
  partitioned data}. In \bibinfo{booktitle}{\emph{ICASSP 2021-2021 IEEE
  International Conference on Acoustics, Speech and Signal Processing
  (ICASSP)}}. IEEE, \bibinfo{pages}{3100--3104}.
\newblock


\bibitem[Demmler et~al\mbox{.}(2015)]%
        {demmler2015aby}
\bibfield{author}{\bibinfo{person}{Daniel Demmler}, \bibinfo{person}{Thomas
  Schneider}, {and} \bibinfo{person}{Michael Zohner}.}
  \bibinfo{year}{2015}\natexlab{}.
\newblock \showarticletitle{ABY-A framework for efficient mixed-protocol secure
  two-party computation.}. In \bibinfo{booktitle}{\emph{NDSS}}.
\newblock


\bibitem[Dias and Meratnia(2023)]%
        {dias2023blocklearning}
\bibfield{author}{\bibinfo{person}{Henrique Dias} {and}
  \bibinfo{person}{Nirvana Meratnia}.} \bibinfo{year}{2023}\natexlab{}.
\newblock \showarticletitle{BlockLearning: A Modular Framework for
  Blockchain-Based Vertical Federated Learning}. In
  \bibinfo{booktitle}{\emph{Ubiquitous Security: Second International
  Conference, UbiSec 2022, Zhangjiajie, China, December 28--31, 2022, Revised
  Selected Papers}}. Springer, \bibinfo{pages}{319--333}.
\newblock


\bibitem[Dwork et~al\mbox{.}(2014)]%
        {dwork2014algorithmic}
\bibfield{author}{\bibinfo{person}{Cynthia Dwork}, \bibinfo{person}{Aaron
  Roth}, {et~al\mbox{.}}} \bibinfo{year}{2014}\natexlab{}.
\newblock \showarticletitle{The algorithmic foundations of differential
  privacy.}
\newblock \bibinfo{journal}{\emph{Foundations and Trends in Theoretical
  Computer Science}} \bibinfo{volume}{9}, \bibinfo{number}{3-4}
  (\bibinfo{year}{2014}), \bibinfo{pages}{211--407}.
\newblock


\bibitem[Einspruch(2012)]%
        {einspruch2012application}
\bibfield{author}{\bibinfo{person}{Norman Einspruch}.}
  \bibinfo{year}{2012}\natexlab{}.
\newblock \bibinfo{booktitle}{\emph{Application specific integrated circuit
  (ASIC) technology}}. Vol.~\bibinfo{volume}{23}.
\newblock \bibinfo{publisher}{Academic Press}.
\newblock


\bibitem[Errounda and Liu(2022)]%
        {errounda2022mobility}
\bibfield{author}{\bibinfo{person}{Fatima~Zahra Errounda} {and}
  \bibinfo{person}{Yan Liu}.} \bibinfo{year}{2022}\natexlab{}.
\newblock \showarticletitle{A Mobility Forecasting Framework with Vertical
  Federated Learning}. In \bibinfo{booktitle}{\emph{2022 IEEE 46th Annual
  Computers, Software, and Applications Conference (COMPSAC)}}. IEEE,
  \bibinfo{pages}{301--310}.
\newblock


\bibitem[Evans(2016)]%
        {evans2016electronic}
\bibfield{author}{\bibinfo{person}{R~Scott Evans}.}
  \bibinfo{year}{2016}\natexlab{}.
\newblock \showarticletitle{Electronic health records: then, now, and in the
  future}.
\newblock \bibinfo{journal}{\emph{Yearbook of medical informatics}}
  \bibinfo{volume}{25}, \bibinfo{number}{S 01} (\bibinfo{year}{2016}),
  \bibinfo{pages}{S48--S61}.
\newblock


\bibitem[Fang et~al\mbox{.}(2021)]%
        {fang2021large}
\bibfield{author}{\bibinfo{person}{Wenjing Fang}, \bibinfo{person}{Derun Zhao},
  \bibinfo{person}{Jin Tan}, \bibinfo{person}{Chaochao Chen},
  \bibinfo{person}{Chaofan Yu}, \bibinfo{person}{Li Wang}, \bibinfo{person}{Lei
  Wang}, \bibinfo{person}{Jun Zhou}, {and} \bibinfo{person}{Benyu Zhang}.}
  \bibinfo{year}{2021}\natexlab{}.
\newblock \showarticletitle{Large-scale Secure XGB for Vertical Federated
  Learning}. In \bibinfo{booktitle}{\emph{Proceedings of the 30th ACM
  International Conference on Information \& Knowledge Management}}.
  \bibinfo{pages}{443--452}.
\newblock


\bibitem[Feng(2022)]%
        {feng2022vertical}
\bibfield{author}{\bibinfo{person}{Siwei Feng}.}
  \bibinfo{year}{2022}\natexlab{}.
\newblock \showarticletitle{Vertical federated learning-based feature selection
  with non-overlapping sample utilization}.
\newblock \bibinfo{journal}{\emph{Expert Systems with Applications}}
  (\bibinfo{year}{2022}), \bibinfo{pages}{118097}.
\newblock


\bibitem[Feng and Yu(2020)]%
        {feng2020multi}
\bibfield{author}{\bibinfo{person}{Siwei Feng} {and} \bibinfo{person}{Han Yu}.}
  \bibinfo{year}{2020}\natexlab{}.
\newblock \showarticletitle{Multi-participant multi-class vertical federated
  learning}.
\newblock \bibinfo{journal}{\emph{arXiv preprint arXiv:2001.11154}}
  (\bibinfo{year}{2020}).
\newblock


\bibitem[Feng et~al\mbox{.}(2019)]%
        {feng2019securegbm}
\bibfield{author}{\bibinfo{person}{Zhi Feng}, \bibinfo{person}{Haoyi Xiong},
  \bibinfo{person}{Chuanyuan Song}, \bibinfo{person}{Sijia Yang},
  \bibinfo{person}{Baoxin Zhao}, \bibinfo{person}{Licheng Wang},
  \bibinfo{person}{Zeyu Chen}, \bibinfo{person}{Shengwen Yang},
  \bibinfo{person}{Liping Liu}, {and} \bibinfo{person}{Jun Huan}.}
  \bibinfo{year}{2019}\natexlab{}.
\newblock \showarticletitle{Securegbm: Secure multi-party gradient boosting}.
  In \bibinfo{booktitle}{\emph{2019 IEEE International Conference on Big Data
  (Big Data)}}. IEEE, \bibinfo{pages}{1312--1321}.
\newblock


\bibitem[Fink et~al\mbox{.}(2021)]%
        {fink2021artificial}
\bibfield{author}{\bibinfo{person}{Olga Fink}, \bibinfo{person}{Torbj{\o}rn
  Netland}, {and} \bibinfo{person}{Stefan Feuerriegelc}.}
  \bibinfo{year}{2021}\natexlab{}.
\newblock \showarticletitle{Artificial intelligence across company borders}.
\newblock \bibinfo{journal}{\emph{Commun. ACM}} \bibinfo{volume}{65},
  \bibinfo{number}{1} (\bibinfo{year}{2021}), \bibinfo{pages}{34--36}.
\newblock


\bibitem[Forsyth and Ponce(2011)]%
        {forsyth2011computer}
\bibfield{author}{\bibinfo{person}{David Forsyth} {and} \bibinfo{person}{Jean
  Ponce}.} \bibinfo{year}{2011}\natexlab{}.
\newblock \bibinfo{booktitle}{\emph{Computer vision: A modern approach.}}
\newblock \bibinfo{publisher}{Prentice hall}.
\newblock


\bibitem[Frazier(2014)]%
        {frazier2014metal}
\bibfield{author}{\bibinfo{person}{William~E Frazier}.}
  \bibinfo{year}{2014}\natexlab{}.
\newblock \showarticletitle{Metal additive manufacturing: a review}.
\newblock \bibinfo{journal}{\emph{Journal of Materials Engineering and
  performance}} \bibinfo{volume}{23}, \bibinfo{number}{6}
  (\bibinfo{year}{2014}), \bibinfo{pages}{1917--1928}.
\newblock


\bibitem[Fu et~al\mbox{.}(2022b)]%
        {fu2022label}
\bibfield{author}{\bibinfo{person}{Chong Fu}, \bibinfo{person}{Xuhong Zhang},
  \bibinfo{person}{Shouling Ji}, \bibinfo{person}{Jinyin Chen},
  \bibinfo{person}{Jingzheng Wu}, \bibinfo{person}{Shanqing Guo},
  \bibinfo{person}{Jun Zhou}, \bibinfo{person}{Alex~X. Liu}, {and}
  \bibinfo{person}{Ting Wang}.} \bibinfo{year}{2022}\natexlab{b}.
\newblock \showarticletitle{Label Inference Attacks Against Vertical Federated
  Learning}. In \bibinfo{booktitle}{\emph{31st USENIX Security Symposium
  (USENIX Security 22)}}. \bibinfo{publisher}{USENIX Association},
  \bibinfo{address}{Boston, MA}.
\newblock
\urldef\tempurl%
\url{https://www.usenix.org/conference/usenixsecurity22/presentation/fu}
\showURL{%
\tempurl}


\bibitem[Fu et~al\mbox{.}(2021)]%
        {fu2021vf2boost}
\bibfield{author}{\bibinfo{person}{Fangcheng Fu}, \bibinfo{person}{Yingxia
  Shao}, \bibinfo{person}{Lele Yu}, \bibinfo{person}{Jiawei Jiang},
  \bibinfo{person}{Huanran Xue}, \bibinfo{person}{Yangyu Tao}, {and}
  \bibinfo{person}{Bin Cui}.} \bibinfo{year}{2021}\natexlab{}.
\newblock \showarticletitle{VF2Boost: Very Fast Vertical Federated Gradient
  Boosting for Cross-Enterprise Learning}. In
  \bibinfo{booktitle}{\emph{Proceedings of the 2021 International Conference on
  Management of Data}}. \bibinfo{pages}{563--576}.
\newblock


\bibitem[Fu et~al\mbox{.}(2022a)]%
        {fu2022blindfl}
\bibfield{author}{\bibinfo{person}{Fangcheng Fu}, \bibinfo{person}{Huanran
  Xue}, \bibinfo{person}{Yong Cheng}, \bibinfo{person}{Yangyu Tao}, {and}
  \bibinfo{person}{Bin Cui}.} \bibinfo{year}{2022}\natexlab{a}.
\newblock \showarticletitle{Blindfl: Vertical federated machine learning
  without peeking into your data}. In \bibinfo{booktitle}{\emph{Proceedings of
  the 2022 International Conference on Management of Data}}.
  \bibinfo{pages}{1316--1330}.
\newblock


\bibitem[Gasc{\'o}n et~al\mbox{.}(2017)]%
        {gascon2017privacy}
\bibfield{author}{\bibinfo{person}{Adri{\`a} Gasc{\'o}n},
  \bibinfo{person}{Phillipp Schoppmann}, \bibinfo{person}{Borja Balle},
  \bibinfo{person}{Mariana Raykova}, \bibinfo{person}{Jack Doerner},
  \bibinfo{person}{Samee Zahur}, {and} \bibinfo{person}{David Evans}.}
  \bibinfo{year}{2017}\natexlab{}.
\newblock \showarticletitle{Privacy-Preserving Distributed Linear Regression on
  High-Dimensional Data.}
\newblock \bibinfo{journal}{\emph{Proc. Priv. Enhancing Technol.}}
  \bibinfo{volume}{2017}, \bibinfo{number}{4} (\bibinfo{year}{2017}),
  \bibinfo{pages}{345--364}.
\newblock


\bibitem[Ge et~al\mbox{.}(2021)]%
        {ge2021failure}
\bibfield{author}{\bibinfo{person}{Ning Ge}, \bibinfo{person}{Guanghao Li},
  \bibinfo{person}{Li Zhang}, {and} \bibinfo{person}{Yi Liu}.}
  \bibinfo{year}{2021}\natexlab{}.
\newblock \showarticletitle{Failure prediction in production line based on
  federated learning: an empirical study}.
\newblock \bibinfo{journal}{\emph{Journal of Intelligent Manufacturing}}
  (\bibinfo{year}{2021}), \bibinfo{pages}{1--18}.
\newblock


\bibitem[Gentry(2009)]%
        {gentry2009fully}
\bibfield{author}{\bibinfo{person}{Craig Gentry}.}
  \bibinfo{year}{2009}\natexlab{}.
\newblock \showarticletitle{Fully homomorphic encryption using ideal lattices}.
  In \bibinfo{booktitle}{\emph{Proceedings of the forty-first annual ACM
  symposium on Theory of computing}}. \bibinfo{pages}{169--178}.
\newblock


\bibitem[Gentry et~al\mbox{.}(2013)]%
        {gentry2013homomorphic}
\bibfield{author}{\bibinfo{person}{Craig Gentry}, \bibinfo{person}{Amit Sahai},
  {and} \bibinfo{person}{Brent Waters}.} \bibinfo{year}{2013}\natexlab{}.
\newblock \showarticletitle{Homomorphic encryption from learning with errors:
  Conceptually-simpler, asymptotically-faster, attribute-based}. In
  \bibinfo{booktitle}{\emph{Annual Cryptology Conference}}. Springer,
  \bibinfo{pages}{75--92}.
\newblock


\bibitem[Gilmer et~al\mbox{.}(2017)]%
        {gilmer2017neural}
\bibfield{author}{\bibinfo{person}{Justin Gilmer}, \bibinfo{person}{Samuel~S
  Schoenholz}, \bibinfo{person}{Patrick~F Riley}, \bibinfo{person}{Oriol
  Vinyals}, {and} \bibinfo{person}{George~E Dahl}.}
  \bibinfo{year}{2017}\natexlab{}.
\newblock \showarticletitle{Neural message passing for quantum chemistry}. In
  \bibinfo{booktitle}{\emph{International conference on machine learning}}.
  PMLR, \bibinfo{pages}{1263--1272}.
\newblock


\bibitem[Goldreich(2003)]%
        {goldreich2003cryptography}
\bibfield{author}{\bibinfo{person}{Oded Goldreich}.}
  \bibinfo{year}{2003}\natexlab{}.
\newblock \showarticletitle{Cryptography and cryptographic protocols}.
\newblock \bibinfo{journal}{\emph{Distributed Computing}} \bibinfo{volume}{16},
  \bibinfo{number}{2} (\bibinfo{year}{2003}), \bibinfo{pages}{177--199}.
\newblock


\bibitem[Goodfellow et~al\mbox{.}(2020)]%
        {goodfellow2020generative}
\bibfield{author}{\bibinfo{person}{Ian Goodfellow}, \bibinfo{person}{Jean
  Pouget-Abadie}, \bibinfo{person}{Mehdi Mirza}, \bibinfo{person}{Bing Xu},
  \bibinfo{person}{David Warde-Farley}, \bibinfo{person}{Sherjil Ozair},
  \bibinfo{person}{Aaron Courville}, {and} \bibinfo{person}{Yoshua Bengio}.}
  \bibinfo{year}{2020}\natexlab{}.
\newblock \showarticletitle{Generative adversarial networks}.
\newblock \bibinfo{journal}{\emph{Commun. ACM}} \bibinfo{volume}{63},
  \bibinfo{number}{11} (\bibinfo{year}{2020}), \bibinfo{pages}{139--144}.
\newblock


\bibitem[Gou et~al\mbox{.}(2021)]%
        {gou2021knowledge}
\bibfield{author}{\bibinfo{person}{Jianping Gou}, \bibinfo{person}{Baosheng
  Yu}, \bibinfo{person}{Stephen~J Maybank}, {and} \bibinfo{person}{Dacheng
  Tao}.} \bibinfo{year}{2021}\natexlab{}.
\newblock \showarticletitle{Knowledge distillation: A survey}.
\newblock \bibinfo{journal}{\emph{International Journal of Computer Vision}}
  \bibinfo{volume}{129}, \bibinfo{number}{6} (\bibinfo{year}{2021}),
  \bibinfo{pages}{1789--1819}.
\newblock


\bibitem[Gu et~al\mbox{.}(2020)]%
        {gu2020federated}
\bibfield{author}{\bibinfo{person}{Bin Gu}, \bibinfo{person}{Zhiyuan Dang},
  \bibinfo{person}{Xiang Li}, {and} \bibinfo{person}{Heng Huang}.}
  \bibinfo{year}{2020}\natexlab{}.
\newblock \showarticletitle{Federated doubly stochastic kernel learning for
  vertically partitioned data}. In \bibinfo{booktitle}{\emph{Proceedings of the
  26th ACM SIGKDD International Conference on Knowledge Discovery \& Data
  Mining}}. \bibinfo{pages}{2483--2493}.
\newblock


\bibitem[Gu et~al\mbox{.}(2021)]%
        {gu2021privacy}
\bibfield{author}{\bibinfo{person}{Bin Gu}, \bibinfo{person}{An Xu},
  \bibinfo{person}{Zhouyuan Huo}, \bibinfo{person}{Cheng Deng}, {and}
  \bibinfo{person}{Heng Huang}.} \bibinfo{year}{2021}\natexlab{}.
\newblock \showarticletitle{Privacy-Preserving Asynchronous Vertical Federated
  Learning Algorithms for Multiparty Collaborative Learning}.
\newblock \bibinfo{journal}{\emph{IEEE Transactions on Neural Networks and
  Learning Systems}} (\bibinfo{year}{2021}).
\newblock


\bibitem[Guo et~al\mbox{.}(2016)]%
        {guo2016rdma}
\bibfield{author}{\bibinfo{person}{Chuanxiong Guo}, \bibinfo{person}{Haitao
  Wu}, \bibinfo{person}{Zhong Deng}, \bibinfo{person}{Gaurav Soni},
  \bibinfo{person}{Jianxi Ye}, \bibinfo{person}{Jitu Padhye}, {and}
  \bibinfo{person}{Marina Lipshteyn}.} \bibinfo{year}{2016}\natexlab{}.
\newblock \showarticletitle{RDMA over commodity ethernet at scale}. In
  \bibinfo{booktitle}{\emph{Proceedings of the 2016 ACM SIGCOMM Conference}}.
  \bibinfo{pages}{202--215}.
\newblock


\bibitem[Gupta and Raskar(2018)]%
        {gupta2018distributed}
\bibfield{author}{\bibinfo{person}{Otkrist Gupta} {and} \bibinfo{person}{Ramesh
  Raskar}.} \bibinfo{year}{2018}\natexlab{}.
\newblock \showarticletitle{Distributed learning of deep neural network over
  multiple agents}.
\newblock \bibinfo{journal}{\emph{Journal of Network and Computer
  Applications}}  \bibinfo{volume}{116} (\bibinfo{year}{2018}),
  \bibinfo{pages}{1--8}.
\newblock


\bibitem[Hamilton et~al\mbox{.}(2017)]%
        {hamilton2017inductive}
\bibfield{author}{\bibinfo{person}{Will Hamilton}, \bibinfo{person}{Zhitao
  Ying}, {and} \bibinfo{person}{Jure Leskovec}.}
  \bibinfo{year}{2017}\natexlab{}.
\newblock \showarticletitle{Inductive representation learning on large graphs}.
\newblock \bibinfo{journal}{\emph{Advances in neural information processing
  systems}}  \bibinfo{volume}{30} (\bibinfo{year}{2017}).
\newblock


\bibitem[Hardy et~al\mbox{.}(2017)]%
        {hardy2017private}
\bibfield{author}{\bibinfo{person}{Stephen Hardy}, \bibinfo{person}{Wilko
  Henecka}, \bibinfo{person}{Hamish Ivey-Law}, \bibinfo{person}{Richard Nock},
  \bibinfo{person}{Giorgio Patrini}, \bibinfo{person}{Guillaume Smith}, {and}
  \bibinfo{person}{Brian Thorne}.} \bibinfo{year}{2017}\natexlab{}.
\newblock \showarticletitle{Private federated learning on vertically
  partitioned data via entity resolution and additively homomorphic
  encryption}.
\newblock \bibinfo{journal}{\emph{arXiv preprint arXiv:1711.10677}}
  (\bibinfo{year}{2017}).
\newblock


\bibitem[Hashemi et~al\mbox{.}(2021)]%
        {hashemi2021vertical}
\bibfield{author}{\bibinfo{person}{Nazila Hashemi}, \bibinfo{person}{Pooyan
  Safari}, \bibinfo{person}{Behnam Shariati}, {and}
  \bibinfo{person}{Johannes~Karl Fischer}.} \bibinfo{year}{2021}\natexlab{}.
\newblock \showarticletitle{Vertical federated learning for privacy-preserving
  ML model development in partially disaggregated networks}. In
  \bibinfo{booktitle}{\emph{2021 European Conference on Optical Communication
  (ECOC)}}. IEEE, \bibinfo{pages}{1--4}.
\newblock


\bibitem[Hinton et~al\mbox{.}(2015)]%
        {hinton2015distilling}
\bibfield{author}{\bibinfo{person}{Geoffrey Hinton}, \bibinfo{person}{Oriol
  Vinyals}, {and} \bibinfo{person}{Jeff Dean}.}
  \bibinfo{year}{2015}\natexlab{}.
\newblock \showarticletitle{Distilling the knowledge in a neural network}.
\newblock \bibinfo{journal}{\emph{arXiv preprint arXiv:1503.02531}}
  (\bibinfo{year}{2015}).
\newblock


\bibitem[Ho(1995)]%
        {ho1995random}
\bibfield{author}{\bibinfo{person}{Tin~Kam Ho}.}
  \bibinfo{year}{1995}\natexlab{}.
\newblock \showarticletitle{Random decision forests}. In
  \bibinfo{booktitle}{\emph{Proceedings of 3rd international conference on
  document analysis and recognition}}, Vol.~\bibinfo{volume}{1}. IEEE,
  \bibinfo{pages}{278--282}.
\newblock


\bibitem[Hofmann et~al\mbox{.}(2008)]%
        {hofmann2008kernel}
\bibfield{author}{\bibinfo{person}{Thomas Hofmann}, \bibinfo{person}{Bernhard
  Sch{\"o}lkopf}, {and} \bibinfo{person}{Alexander~J Smola}.}
  \bibinfo{year}{2008}\natexlab{}.
\newblock \showarticletitle{Kernel methods in machine learning}.
\newblock \bibinfo{journal}{\emph{The annals of statistics}}
  \bibinfo{volume}{36}, \bibinfo{number}{3} (\bibinfo{year}{2008}),
  \bibinfo{pages}{1171--1220}.
\newblock


\bibitem[Hou et~al\mbox{.}(2021)]%
        {hou2021verifiable}
\bibfield{author}{\bibinfo{person}{Jinpeng Hou}, \bibinfo{person}{Mang Su},
  \bibinfo{person}{Anmin Fu}, {and} \bibinfo{person}{Yan Yu}.}
  \bibinfo{year}{2021}\natexlab{}.
\newblock \showarticletitle{Verifiable Privacy-preserving Scheme based on
  Vertical Federated Random Forest}.
\newblock \bibinfo{journal}{\emph{IEEE Internet of Things Journal}}
  (\bibinfo{year}{2021}).
\newblock


\bibitem[Hu et~al\mbox{.}(2019)]%
        {hu2019fdml}
\bibfield{author}{\bibinfo{person}{Yaochen Hu}, \bibinfo{person}{Di Niu},
  \bibinfo{person}{Jianming Yang}, {and} \bibinfo{person}{Shengping Zhou}.}
  \bibinfo{year}{2019}\natexlab{}.
\newblock \showarticletitle{Fdml: A collaborative machine learning framework
  for distributed features}. In \bibinfo{booktitle}{\emph{Proceedings of the
  25th ACM SIGKDD International Conference on Knowledge Discovery \& Data
  Mining}}. \bibinfo{pages}{2232--2240}.
\newblock


\bibitem[Huang et~al\mbox{.}(2022a)]%
        {huang2022efmvfl}
\bibfield{author}{\bibinfo{person}{Yimin Huang}, \bibinfo{person}{Xinyu Feng},
  \bibinfo{person}{Wanwan Wang}, \bibinfo{person}{Hao He},
  \bibinfo{person}{Yukun Wang}, {and} \bibinfo{person}{Ming Yao}.}
  \bibinfo{year}{2022}\natexlab{a}.
\newblock \showarticletitle{EFMVFL: An Efficient and Flexible Multi-party
  Vertical Federated Learning without a Third Party}.
\newblock \bibinfo{journal}{\emph{arXiv preprint arXiv:2201.06244}}
  (\bibinfo{year}{2022}).
\newblock


\bibitem[Huang et~al\mbox{.}(2022b)]%
        {huang2022cheetah}
\bibfield{author}{\bibinfo{person}{Zhicong Huang}, \bibinfo{person}{Wen-jie
  Lu}, \bibinfo{person}{Cheng Hong}, {and} \bibinfo{person}{Jiansheng Ding}.}
  \bibinfo{year}{2022}\natexlab{b}.
\newblock \showarticletitle{Cheetah: Lean and Fast Secure Two-Party Deep Neural
  Network Inference}.
\newblock \bibinfo{journal}{\emph{Cryptology ePrint Archive}}
  (\bibinfo{year}{2022}).
\newblock


\bibitem[Hussain and Fernando(2009)]%
        {hussain2009spectrum}
\bibfield{author}{\bibinfo{person}{Sattar Hussain} {and}
  \bibinfo{person}{Xavier Fernando}.} \bibinfo{year}{2009}\natexlab{}.
\newblock \showarticletitle{Spectrum sensing in cognitive radio networks:
  Up-to-date techniques and future challenges}. In
  \bibinfo{booktitle}{\emph{2009 IEEE Toronto International Conference Science
  and Technology for Humanity (TIC-STH)}}. IEEE, \bibinfo{pages}{736--741}.
\newblock


\bibitem[Ilin et~al\mbox{.}(2021)]%
        {ilin2021public}
\bibfield{author}{\bibinfo{person}{Cornelia Ilin},
  \bibinfo{person}{S{\'e}bastien Annan-Phan}, \bibinfo{person}{Xiao~Hui Tai},
  \bibinfo{person}{Shikhar Mehra}, \bibinfo{person}{Solomon Hsiang}, {and}
  \bibinfo{person}{Joshua~E Blumenstock}.} \bibinfo{year}{2021}\natexlab{}.
\newblock \showarticletitle{Public mobility data enables COVID-19 forecasting
  and management at local and global scales}.
\newblock \bibinfo{journal}{\emph{Scientific reports}} \bibinfo{volume}{11},
  \bibinfo{number}{1} (\bibinfo{year}{2021}), \bibinfo{pages}{1--11}.
\newblock


\bibitem[Jiang et~al\mbox{.}(2022)]%
        {jiang2022comprehensive}
\bibfield{author}{\bibinfo{person}{Xue Jiang}, \bibinfo{person}{Xuebing Zhou},
  {and} \bibinfo{person}{Jens Grossklags}.} \bibinfo{year}{2022}\natexlab{}.
\newblock \showarticletitle{Comprehensive Analysis of Privacy Leakage in
  Vertical Federated Learning During Prediction}.
\newblock \bibinfo{journal}{\emph{Proceedings on Privacy Enhancing
  Technologies}}  \bibinfo{volume}{2} (\bibinfo{year}{2022}),
  \bibinfo{pages}{2022}.
\newblock


\bibitem[Jiang et~al\mbox{.}(2021)]%
        {jiang2021flashe}
\bibfield{author}{\bibinfo{person}{Zhifeng Jiang}, \bibinfo{person}{Wei Wang},
  {and} \bibinfo{person}{Yang Liu}.} \bibinfo{year}{2021}\natexlab{}.
\newblock \showarticletitle{Flashe: Additively symmetric homomorphic encryption
  for cross-silo federated learning}.
\newblock \bibinfo{journal}{\emph{arXiv preprint arXiv:2109.00675}}
  (\bibinfo{year}{2021}).
\newblock


\bibitem[Jin et~al\mbox{.}(2022)]%
        {jin2022towards}
\bibfield{author}{\bibinfo{person}{Chao Jin}, \bibinfo{person}{Jun Wang},
  \bibinfo{person}{Sin~G Teo}, \bibinfo{person}{Le Zhang}, \bibinfo{person}{CS
  Chan}, \bibinfo{person}{Qibin Hou}, {and} \bibinfo{person}{Khin Mi~Mi Aung}.}
  \bibinfo{year}{2022}\natexlab{}.
\newblock \showarticletitle{Towards End-to-End Secure and Efficient Federated
  Learning for XGBoost}.
\newblock  (\bibinfo{year}{2022}).
\newblock


\bibitem[Kalia et~al\mbox{.}(2016)]%
        {kalia2016design}
\bibfield{author}{\bibinfo{person}{Anuj Kalia}, \bibinfo{person}{Michael
  Kaminsky}, {and} \bibinfo{person}{David~G Andersen}.}
  \bibinfo{year}{2016}\natexlab{}.
\newblock \showarticletitle{Design guidelines for high performance $\{$RDMA$\}$
  systems}. In \bibinfo{booktitle}{\emph{2016 $\{$USENIX$\}$ Annual Technical
  Conference ($\{$USENIX$\}$$\{$ATC$\}$ 16)}}. \bibinfo{pages}{437--450}.
\newblock


\bibitem[Kang et~al\mbox{.}(2022)]%
        {kang2022fedcvt}
\bibfield{author}{\bibinfo{person}{Yan Kang}, \bibinfo{person}{Yang Liu}, {and}
  \bibinfo{person}{Xinle Liang}.} \bibinfo{year}{2022}\natexlab{}.
\newblock \showarticletitle{FedCVT: Semi-Supervised Vertical Federated Learning
  with Cross-View Training}.
\newblock \bibinfo{journal}{\emph{ACM Transactions on Intelligent Systems and
  Technology (TIST)}} (\bibinfo{year}{2022}).
\newblock


\bibitem[Kang et~al\mbox{.}(2021)]%
        {kang2021privacy}
\bibfield{author}{\bibinfo{person}{Yan Kang}, \bibinfo{person}{Yang Liu},
  \bibinfo{person}{Yuezhou Wu}, \bibinfo{person}{Guoqiang Ma}, {and}
  \bibinfo{person}{Qiang Yang}.} \bibinfo{year}{2021}\natexlab{}.
\newblock \showarticletitle{Privacy-preserving Federated Adversarial Domain
  Adaptation over Feature Groups for Interpretability}. In
  \bibinfo{booktitle}{\emph{International Workshop on Federated Learning for
  User Privacy and Data Confidentiality in Conjunction with IJCAI 2021
  (FL-IJCAI'21)}}.
\newblock


\bibitem[Karr et~al\mbox{.}(2009)]%
        {karr2009privacy}
\bibfield{author}{\bibinfo{person}{Alan~F Karr}, \bibinfo{person}{Xiaodong
  Lin}, \bibinfo{person}{Ashish~P Sanil}, {and} \bibinfo{person}{Jerome~P
  Reiter}.} \bibinfo{year}{2009}\natexlab{}.
\newblock \showarticletitle{Privacy-preserving analysis of vertically
  partitioned data using secure matrix products}.
\newblock \bibinfo{journal}{\emph{Journal of Official Statistics}}
  \bibinfo{volume}{25}, \bibinfo{number}{1} (\bibinfo{year}{2009}),
  \bibinfo{pages}{125}.
\newblock


\bibitem[Ke et~al\mbox{.}(2017)]%
        {ke2017lightgbm}
\bibfield{author}{\bibinfo{person}{Guolin Ke}, \bibinfo{person}{Qi Meng},
  \bibinfo{person}{Thomas Finley}, \bibinfo{person}{Taifeng Wang},
  \bibinfo{person}{Wei Chen}, \bibinfo{person}{Weidong Ma},
  \bibinfo{person}{Qiwei Ye}, {and} \bibinfo{person}{Tie-Yan Liu}.}
  \bibinfo{year}{2017}\natexlab{}.
\newblock \showarticletitle{Lightgbm: A highly efficient gradient boosting
  decision tree}.
\newblock \bibinfo{journal}{\emph{Advances in neural information processing
  systems}}  \bibinfo{volume}{30} (\bibinfo{year}{2017}).
\newblock


\bibitem[Khan et~al\mbox{.}(2022)]%
        {khan2022communication}
\bibfield{author}{\bibinfo{person}{Afsana Khan}, \bibinfo{person}{Marijn ten
  Thij}, {and} \bibinfo{person}{Anna Wilbik}.} \bibinfo{year}{2022}\natexlab{}.
\newblock \showarticletitle{Communication-efficient vertical federated
  learning}.
\newblock \bibinfo{journal}{\emph{Algorithms}} \bibinfo{volume}{15},
  \bibinfo{number}{8} (\bibinfo{year}{2022}), \bibinfo{pages}{273}.
\newblock


\bibitem[Khodaparast et~al\mbox{.}(2018)]%
        {khodaparast2018privacy}
\bibfield{author}{\bibinfo{person}{Fatemeh Khodaparast}, \bibinfo{person}{Mina
  Sheikhalishahi}, \bibinfo{person}{Hassan Haghighi}, {and}
  \bibinfo{person}{Fabio Martinelli}.} \bibinfo{year}{2018}\natexlab{}.
\newblock \showarticletitle{Privacy preserving random decision tree
  classification over horizontally and vertically partitioned data}. In
  \bibinfo{booktitle}{\emph{2018 IEEE 16th Intl Conf on Dependable, Autonomic
  and Secure Computing, 16th Intl Conf on Pervasive Intelligence and Computing,
  4th Intl Conf on Big Data Intelligence and Computing and Cyber Science and
  Technology Congress (DASC/PiCom/DataCom/CyberSciTech)}}. IEEE,
  \bibinfo{pages}{600--607}.
\newblock


\bibitem[Kholod et~al\mbox{.}(2021)]%
        {kholod2021parallelization}
\bibfield{author}{\bibinfo{person}{Ivan Kholod}, \bibinfo{person}{Andrey
  Rukavitsyn}, \bibinfo{person}{Alexey Paznikov}, {and} \bibinfo{person}{Sergei
  Gorlatch}.} \bibinfo{year}{2021}\natexlab{}.
\newblock \showarticletitle{Parallelization of the self-organized maps
  algorithm for federated learning on distributed sources}.
\newblock \bibinfo{journal}{\emph{The Journal of Supercomputing}}
  \bibinfo{volume}{77}, \bibinfo{number}{6} (\bibinfo{year}{2021}),
  \bibinfo{pages}{6197--6213}.
\newblock


\bibitem[Kim et~al\mbox{.}(2021)]%
        {kim2021vertical}
\bibfield{author}{\bibinfo{person}{Jihoon Kim}, \bibinfo{person}{Wentao Li},
  \bibinfo{person}{Tyler Bath}, \bibinfo{person}{Xiaoqian Jiang}, {and}
  \bibinfo{person}{Lucila Ohno-Machado}.} \bibinfo{year}{2021}\natexlab{}.
\newblock \showarticletitle{VERTIcal Grid lOgistic regression with Confidence
  Intervals (VERTIGO-CI)}. In \bibinfo{booktitle}{\emph{AMIA Annual Symposium
  Proceedings}}, Vol.~\bibinfo{volume}{2021}. American Medical Informatics
  Association, \bibinfo{pages}{355}.
\newblock


\bibitem[Kipf and Welling(2016)]%
        {kipf2016semi}
\bibfield{author}{\bibinfo{person}{Thomas~N Kipf} {and} \bibinfo{person}{Max
  Welling}.} \bibinfo{year}{2016}\natexlab{}.
\newblock \showarticletitle{Semi-supervised classification with graph
  convolutional networks}.
\newblock \bibinfo{journal}{\emph{arXiv preprint arXiv:1609.02907}}
  (\bibinfo{year}{2016}).
\newblock


\bibitem[Koren et~al\mbox{.}(2009)]%
        {koren2009matrix}
\bibfield{author}{\bibinfo{person}{Yehuda Koren}, \bibinfo{person}{Robert
  Bell}, {and} \bibinfo{person}{Chris Volinsky}.}
  \bibinfo{year}{2009}\natexlab{}.
\newblock \showarticletitle{Matrix factorization techniques for recommender
  systems}.
\newblock \bibinfo{journal}{\emph{Computer}} \bibinfo{volume}{42},
  \bibinfo{number}{8} (\bibinfo{year}{2009}), \bibinfo{pages}{30--37}.
\newblock


\bibitem[Kuon et~al\mbox{.}(2008)]%
        {kuon2008fpga}
\bibfield{author}{\bibinfo{person}{Ian Kuon}, \bibinfo{person}{Russell
  Tessier}, {and} \bibinfo{person}{Jonathan Rose}.}
  \bibinfo{year}{2008}\natexlab{}.
\newblock \bibinfo{booktitle}{\emph{FPGA architecture: Survey and challenges}}.
\newblock \bibinfo{publisher}{Now Publishers Inc}.
\newblock


\bibitem[Le et~al\mbox{.}(2021)]%
        {le2021fedxgboost}
\bibfield{author}{\bibinfo{person}{Nhan~Khanh Le}, \bibinfo{person}{Yang Liu},
  \bibinfo{person}{Quang~Minh Nguyen}, \bibinfo{person}{Qingchen Liu},
  \bibinfo{person}{Fangzhou Liu}, \bibinfo{person}{Quanwei Cai}, {and}
  \bibinfo{person}{Sandra Hirche}.} \bibinfo{year}{2021}\natexlab{}.
\newblock \showarticletitle{FedXGBoost: Privacy-Preserving XGBoost for
  Federated Learning}. In \bibinfo{booktitle}{\emph{International Workshop on
  Federated Learning for User Privacy and Data Confidentiality in Conjunction
  with IJCAI 2021 (FL-IJCAI'21)}}.
\newblock


\bibitem[Lee et~al\mbox{.}(2018)]%
        {lee2018privacy}
\bibfield{author}{\bibinfo{person}{Junghye Lee}, \bibinfo{person}{Jimeng Sun},
  \bibinfo{person}{Fei Wang}, \bibinfo{person}{Shuang Wang},
  \bibinfo{person}{Chi-Hyuck Jun}, {and} \bibinfo{person}{Xiaoqian Jiang}.}
  \bibinfo{year}{2018}\natexlab{}.
\newblock \showarticletitle{Privacy-preserving patient similarity learning in a
  federated environment: development and analysis}.
\newblock \bibinfo{journal}{\emph{JMIR medical informatics}}
  \bibinfo{volume}{6}, \bibinfo{number}{2} (\bibinfo{year}{2018}),
  \bibinfo{pages}{e7744}.
\newblock


\bibitem[Lee et~al\mbox{.}(2017)]%
        {lee2017speeding}
\bibfield{author}{\bibinfo{person}{Kangwook Lee}, \bibinfo{person}{Maximilian
  Lam}, \bibinfo{person}{Ramtin Pedarsani}, \bibinfo{person}{Dimitris
  Papailiopoulos}, {and} \bibinfo{person}{Kannan Ramchandran}.}
  \bibinfo{year}{2017}\natexlab{}.
\newblock \showarticletitle{Speeding up distributed machine learning using
  codes}.
\newblock \bibinfo{journal}{\emph{IEEE Transactions on Information Theory}}
  \bibinfo{volume}{64}, \bibinfo{number}{3} (\bibinfo{year}{2017}),
  \bibinfo{pages}{1514--1529}.
\newblock


\bibitem[Leo et~al\mbox{.}(2019)]%
        {leo2019machine}
\bibfield{author}{\bibinfo{person}{Martin Leo}, \bibinfo{person}{Suneel
  Sharma}, {and} \bibinfo{person}{Koilakuntla Maddulety}.}
  \bibinfo{year}{2019}\natexlab{}.
\newblock \showarticletitle{Machine learning in banking risk management: A
  literature review}.
\newblock \bibinfo{journal}{\emph{Risks}} \bibinfo{volume}{7},
  \bibinfo{number}{1} (\bibinfo{year}{2019}), \bibinfo{pages}{29}.
\newblock


\bibitem[Letaief et~al\mbox{.}(2021)]%
        {letaief2021edge}
\bibfield{author}{\bibinfo{person}{Khaled~B Letaief}, \bibinfo{person}{Yuanming
  Shi}, \bibinfo{person}{Jianmin Lu}, {and} \bibinfo{person}{Jianhua Lu}.}
  \bibinfo{year}{2021}\natexlab{}.
\newblock \showarticletitle{Edge artificial intelligence for 6G: Vision,
  enabling technologies, and applications}.
\newblock \bibinfo{journal}{\emph{IEEE Journal on Selected Areas in
  Communications}} \bibinfo{volume}{40}, \bibinfo{number}{1}
  (\bibinfo{year}{2021}), \bibinfo{pages}{5--36}.
\newblock


\bibitem[Li et~al\mbox{.}(2022b)]%
        {li2022nearest}
\bibfield{author}{\bibinfo{person}{Denghao Li}, \bibinfo{person}{Jianzong
  Wang}, \bibinfo{person}{Lingwei Kong}, \bibinfo{person}{Shijing Si},
  \bibinfo{person}{Zhangcheng Huang}, \bibinfo{person}{Chenyu Huang}, {and}
  \bibinfo{person}{Jing Xiao}.} \bibinfo{year}{2022}\natexlab{b}.
\newblock \showarticletitle{A Nearest Neighbor Under-sampling Strategy for
  Vertical Federated Learning in Financial Domain}. In
  \bibinfo{booktitle}{\emph{Proceedings of the 2022 ACM Workshop on Information
  Hiding and Multimedia Security}}. \bibinfo{pages}{123--128}.
\newblock


\bibitem[Li et~al\mbox{.}(2022c)]%
        {li2022vfl}
\bibfield{author}{\bibinfo{person}{Jialin Li}, \bibinfo{person}{Tongjiang Yan},
  {and} \bibinfo{person}{Pengcheng Ren}.} \bibinfo{year}{2022}\natexlab{c}.
\newblock \showarticletitle{VFL-R: a novel framework for multi-party in
  vertical federated learning}.
\newblock \bibinfo{journal}{\emph{Applied Intelligence}}
  (\bibinfo{year}{2022}), \bibinfo{pages}{1--17}.
\newblock


\bibitem[Li et~al\mbox{.}(2020)]%
        {li2020efficient}
\bibfield{author}{\bibinfo{person}{Ming Li}, \bibinfo{person}{Yiwei Chen},
  \bibinfo{person}{Yiqin Wang}, {and} \bibinfo{person}{Y Pan}.}
  \bibinfo{year}{2020}\natexlab{}.
\newblock \showarticletitle{Efficient Asynchronous Vertical Federated Learning
  via Gradient Prediction and Double-End Sparse Compression}. In
  \bibinfo{booktitle}{\emph{2020 16th International Conference on Control,
  Automation, Robotics and Vision (ICARCV)}}. IEEE, \bibinfo{pages}{291--296}.
\newblock


\bibitem[Li et~al\mbox{.}(2021b)]%
        {li2021label}
\bibfield{author}{\bibinfo{person}{Oscar Li}, \bibinfo{person}{Jiankai Sun},
  \bibinfo{person}{Xin Yang}, \bibinfo{person}{Weihao Gao},
  \bibinfo{person}{Hongyi Zhang}, \bibinfo{person}{Junyuan Xie},
  \bibinfo{person}{Virginia Smith}, {and} \bibinfo{person}{Chong Wang}.}
  \bibinfo{year}{2021}\natexlab{b}.
\newblock \showarticletitle{Label leakage and protection in two-party split
  learning}.
\newblock \bibinfo{journal}{\emph{arXiv preprint arXiv:2102.08504}}
  (\bibinfo{year}{2021}).
\newblock


\bibitem[Li et~al\mbox{.}(2016)]%
        {li2016vertical}
\bibfield{author}{\bibinfo{person}{Yong Li}, \bibinfo{person}{Xiaoqian Jiang},
  \bibinfo{person}{Shuang Wang}, \bibinfo{person}{Hongkai Xiong}, {and}
  \bibinfo{person}{Lucila Ohno-Machado}.} \bibinfo{year}{2016}\natexlab{}.
\newblock \showarticletitle{Vertical grid logistic regression (vertigo)}.
\newblock \bibinfo{journal}{\emph{Journal of the American Medical Informatics
  Association}} \bibinfo{volume}{23}, \bibinfo{number}{3}
  (\bibinfo{year}{2016}), \bibinfo{pages}{570--579}.
\newblock


\bibitem[Li et~al\mbox{.}(2022a)]%
        {li2022adaptive}
\bibfield{author}{\bibinfo{person}{Yuanzhang Li}, \bibinfo{person}{Tianchi
  Sha}, \bibinfo{person}{Thar Baker}, \bibinfo{person}{Xiao Yu},
  \bibinfo{person}{Zhiwei Shi}, {and} \bibinfo{person}{Sikang Hu}.}
  \bibinfo{year}{2022}\natexlab{a}.
\newblock \showarticletitle{Adaptive vertical federated learning via feature
  map transferring in mobile edge computing}.
\newblock \bibinfo{journal}{\emph{Computing}} (\bibinfo{year}{2022}),
  \bibinfo{pages}{1--17}.
\newblock


\bibitem[Li et~al\mbox{.}(2021a)]%
        {li2021survey}
\bibfield{author}{\bibinfo{person}{Zewen Li}, \bibinfo{person}{Fan Liu},
  \bibinfo{person}{Wenjie Yang}, \bibinfo{person}{Shouheng Peng}, {and}
  \bibinfo{person}{Jun Zhou}.} \bibinfo{year}{2021}\natexlab{a}.
\newblock \showarticletitle{A survey of convolutional neural networks:
  analysis, applications, and prospects}.
\newblock \bibinfo{journal}{\emph{IEEE transactions on neural networks and
  learning systems}} (\bibinfo{year}{2021}).
\newblock


\bibitem[Liang and Chawathe(2004)]%
        {liang2004privacy}
\bibfield{author}{\bibinfo{person}{Gang Liang} {and}
  \bibinfo{person}{Sudarshan~S Chawathe}.} \bibinfo{year}{2004}\natexlab{}.
\newblock \showarticletitle{Privacy-preserving inter-database operations}. In
  \bibinfo{booktitle}{\emph{International Conference on Intelligence and
  Security Informatics}}. Springer, \bibinfo{pages}{66--82}.
\newblock


\bibitem[Liang et~al\mbox{.}(2021)]%
        {liang2021self}
\bibfield{author}{\bibinfo{person}{Xinle Liang}, \bibinfo{person}{Yang Liu},
  \bibinfo{person}{Jiahuan Luo}, \bibinfo{person}{Yuanqin He},
  \bibinfo{person}{Tianjian Chen}, {and} \bibinfo{person}{Qiang Yang}.}
  \bibinfo{year}{2021}\natexlab{}.
\newblock \showarticletitle{Self-supervised cross-silo federated neural
  architecture search}.
\newblock \bibinfo{journal}{\emph{arXiv preprint arXiv:2101.11896}}
  (\bibinfo{year}{2021}).
\newblock


\bibitem[Liu(2020)]%
        {liu2020accelerating}
\bibfield{author}{\bibinfo{person}{Duowen Liu}.}
  \bibinfo{year}{2020}\natexlab{}.
\newblock \showarticletitle{Accelerating Intra-Party Communication in Vertical
  Federated Learning with RDMA}. In \bibinfo{booktitle}{\emph{Proceedings of
  the 1st Workshop on Distributed Machine Learning}}. \bibinfo{pages}{14--20}.
\newblock


\bibitem[Liu et~al\mbox{.}(2020b)]%
        {liu2020federated}
\bibfield{author}{\bibinfo{person}{Fenglin Liu}, \bibinfo{person}{Xian Wu},
  \bibinfo{person}{Shen Ge}, \bibinfo{person}{Wei Fan}, {and}
  \bibinfo{person}{Yuexian Zou}.} \bibinfo{year}{2020}\natexlab{b}.
\newblock \showarticletitle{Federated learning for vision-and-language
  grounding problems}. In \bibinfo{booktitle}{\emph{Proceedings of the AAAI
  Conference on Artificial Intelligence}}, Vol.~\bibinfo{volume}{34}.
  \bibinfo{pages}{11572--11579}.
\newblock


\bibitem[Liu et~al\mbox{.}(2021b)]%
        {liu2021distributed}
\bibfield{author}{\bibinfo{person}{Ji Liu}, \bibinfo{person}{Jizhou Huang},
  \bibinfo{person}{Yang Zhou}, \bibinfo{person}{Xuhong Li},
  \bibinfo{person}{Shilei Ji}, \bibinfo{person}{Haoyi Xiong}, {and}
  \bibinfo{person}{Dejing Dou}.} \bibinfo{year}{2021}\natexlab{b}.
\newblock \showarticletitle{From Distributed Machine Learning to Federated
  Learning: A Survey}.
\newblock \bibinfo{journal}{\emph{arXiv preprint arXiv:2104.14362}}
  (\bibinfo{year}{2021}).
\newblock


\bibitem[Liu and Lyu(2022)]%
        {liu2022clustering}
\bibfield{author}{\bibinfo{person}{Junlin Liu} {and} \bibinfo{person}{Xinchen
  Lyu}.} \bibinfo{year}{2022}\natexlab{}.
\newblock \showarticletitle{Clustering Label Inference Attack against Practical
  Split Learning}.
\newblock \bibinfo{journal}{\emph{arXiv preprint arXiv:2203.05222}}
  (\bibinfo{year}{2022}).
\newblock


\bibitem[Liu et~al\mbox{.}(2022)]%
        {liu2022vertical}
\bibfield{author}{\bibinfo{person}{Peixi Liu}, \bibinfo{person}{Guangxu Zhu},
  \bibinfo{person}{Wei Jiang}, \bibinfo{person}{Wu Luo}, \bibinfo{person}{Jie
  Xu}, {and} \bibinfo{person}{Shuguang Cui}.} \bibinfo{year}{2022}\natexlab{}.
\newblock \showarticletitle{Vertical federated edge learning with distributed
  integrated sensing and communication}.
\newblock \bibinfo{journal}{\emph{IEEE Communications Letters}}
  \bibinfo{volume}{26}, \bibinfo{number}{9} (\bibinfo{year}{2022}),
  \bibinfo{pages}{2091--2095}.
\newblock


\bibitem[Liu et~al\mbox{.}(2020a)]%
        {liu2020fedcoin}
\bibfield{author}{\bibinfo{person}{Yuan Liu}, \bibinfo{person}{Zhengpeng Ai},
  \bibinfo{person}{Shuai Sun}, \bibinfo{person}{Shuangfeng Zhang},
  \bibinfo{person}{Zelei Liu}, {and} \bibinfo{person}{Han Yu}.}
  \bibinfo{year}{2020}\natexlab{a}.
\newblock \showarticletitle{Fedcoin: A peer-to-peer payment system for
  federated learning}.
\newblock In \bibinfo{booktitle}{\emph{Federated Learning}}.
  \bibinfo{publisher}{Springer}, \bibinfo{pages}{125--138}.
\newblock


\bibitem[Liu et~al\mbox{.}(2021a)]%
        {liu2021fate}
\bibfield{author}{\bibinfo{person}{Yang Liu}, \bibinfo{person}{Tao Fan},
  \bibinfo{person}{Tianjian Chen}, \bibinfo{person}{Qian Xu}, {and}
  \bibinfo{person}{Qiang Yang}.} \bibinfo{year}{2021}\natexlab{a}.
\newblock \showarticletitle{FATE: An Industrial Grade Platform for
  Collaborative Learning With Data Protection}.
\newblock \bibinfo{journal}{\emph{Journal of Machine Learning Research}}
  \bibinfo{volume}{22}, \bibinfo{number}{226} (\bibinfo{year}{2021}),
  \bibinfo{pages}{1--6}.
\newblock


\bibitem[Liu et~al\mbox{.}(2019)]%
        {liu2019communication}
\bibfield{author}{\bibinfo{person}{Yang Liu}, \bibinfo{person}{Yan Kang},
  \bibinfo{person}{Xinwei Zhang}, \bibinfo{person}{Liping Li},
  \bibinfo{person}{Yong Cheng}, \bibinfo{person}{Tianjian Chen},
  \bibinfo{person}{Mingyi Hong}, {and} \bibinfo{person}{Qiang Yang}.}
  \bibinfo{year}{2019}\natexlab{}.
\newblock \showarticletitle{A communication efficient collaborative learning
  framework for distributed features}.
\newblock \bibinfo{journal}{\emph{arXiv preprint arXiv:1912.11187}}
  (\bibinfo{year}{2019}).
\newblock


\bibitem[Liu et~al\mbox{.}(2020c)]%
        {liu2020backdoor}
\bibfield{author}{\bibinfo{person}{Yang Liu}, \bibinfo{person}{Zhihao Yi},
  {and} \bibinfo{person}{Tianjian Chen}.} \bibinfo{year}{2020}\natexlab{c}.
\newblock \showarticletitle{Backdoor attacks and defenses in
  feature-partitioned collaborative learning}.
\newblock \bibinfo{journal}{\emph{arXiv preprint arXiv:2007.03608}}
  (\bibinfo{year}{2020}).
\newblock


\bibitem[Liu et~al\mbox{.}(2021c)]%
        {liu2022batch}
\bibfield{author}{\bibinfo{person}{Yang Liu}, \bibinfo{person}{Tianyuan Zou},
  \bibinfo{person}{Yan Kang}, \bibinfo{person}{Wenhan Liu},
  \bibinfo{person}{Yuanqin He}, \bibinfo{person}{Zhihao Yi}, {and}
  \bibinfo{person}{Qiang Yang}.} \bibinfo{year}{2021}\natexlab{c}.
\newblock \showarticletitle{Batch Label Inference and Replacement Attacks in
  Black-Boxed Vertical Federated Learning}.
\newblock \bibinfo{journal}{\emph{arXiv preprint arXiv:2112.05409v2}}
  (\bibinfo{year}{2021}).
\newblock


\bibitem[Lu et~al\mbox{.}(2022)]%
        {lu2022truthful}
\bibfield{author}{\bibinfo{person}{Jianfeng Lu}, \bibinfo{person}{Bangqi Pan},
  \bibinfo{person}{Abegaz~Mohammed Seid}, \bibinfo{person}{Bing Li},
  \bibinfo{person}{Gangqiang Hu}, {and} \bibinfo{person}{Shaohua Wan}.}
  \bibinfo{year}{2022}\natexlab{}.
\newblock \showarticletitle{Truthful Incentive Mechanism Design via
  Internalizing Externalities and LP Relaxation for Vertical Federated
  Learning}.
\newblock \bibinfo{journal}{\emph{IEEE Transactions on Computational Social
  Systems}} (\bibinfo{year}{2022}).
\newblock


\bibitem[Lu et~al\mbox{.}(2015)]%
        {lu2015recommender}
\bibfield{author}{\bibinfo{person}{Jie Lu}, \bibinfo{person}{Dianshuang Wu},
  \bibinfo{person}{Mingsong Mao}, \bibinfo{person}{Wei Wang}, {and}
  \bibinfo{person}{Guangquan Zhang}.} \bibinfo{year}{2015}\natexlab{}.
\newblock \showarticletitle{Recommender system application developments: a
  survey}.
\newblock \bibinfo{journal}{\emph{Decision Support Systems}}
  \bibinfo{volume}{74} (\bibinfo{year}{2015}), \bibinfo{pages}{12--32}.
\newblock


\bibitem[Lu and Ding(2020)]%
        {lu2020multi}
\bibfield{author}{\bibinfo{person}{Linpeng Lu} {and} \bibinfo{person}{Ning
  Ding}.} \bibinfo{year}{2020}\natexlab{}.
\newblock \showarticletitle{Multi-Party Private Set Intersection in Vertical
  Federated Learning}. In \bibinfo{booktitle}{\emph{2020 IEEE 19th
  International Conference on Trust, Security and Privacy in Computing and
  Communications (TrustCom)}}. IEEE, \bibinfo{pages}{707--714}.
\newblock


\bibitem[Luo et~al\mbox{.}(2021a)]%
        {luo2021secure}
\bibfield{author}{\bibinfo{person}{Qiyao Luo}, \bibinfo{person}{Yilei Wang},
  \bibinfo{person}{Zhenghang Ren}, \bibinfo{person}{Ke Yi},
  \bibinfo{person}{Kai Chen}, {and} \bibinfo{person}{Xiao Wang}.}
  \bibinfo{year}{2021}\natexlab{a}.
\newblock \showarticletitle{Secure Machine Learning over Relational Data}.
\newblock \bibinfo{journal}{\emph{arXiv preprint arXiv:2109.14806}}
  (\bibinfo{year}{2021}).
\newblock


\bibitem[Luo et~al\mbox{.}(2021b)]%
        {luo2021feature}
\bibfield{author}{\bibinfo{person}{Xinjian Luo}, \bibinfo{person}{Yuncheng Wu},
  \bibinfo{person}{Xiaokui Xiao}, {and} \bibinfo{person}{Beng~Chin Ooi}.}
  \bibinfo{year}{2021}\natexlab{b}.
\newblock \showarticletitle{Feature inference attack on model predictions in
  vertical federated learning}. In \bibinfo{booktitle}{\emph{2021 IEEE 37th
  International Conference on Data Engineering (ICDE)}}. IEEE,
  \bibinfo{pages}{181--192}.
\newblock


\bibitem[Ma and Sun(2020)]%
        {ma2020machine}
\bibfield{author}{\bibinfo{person}{Liye Ma} {and} \bibinfo{person}{Baohong
  Sun}.} \bibinfo{year}{2020}\natexlab{}.
\newblock \showarticletitle{Machine learning and AI in marketing--Connecting
  computing power to human insights}.
\newblock \bibinfo{journal}{\emph{International Journal of Research in
  Marketing}} \bibinfo{volume}{37}, \bibinfo{number}{3} (\bibinfo{year}{2020}),
  \bibinfo{pages}{481--504}.
\newblock


\bibitem[Mahdikhani et~al\mbox{.}(2020)]%
        {mahdikhani2020achieving}
\bibfield{author}{\bibinfo{person}{Hassan Mahdikhani},
  \bibinfo{person}{Rongxing Lu}, \bibinfo{person}{Yandong Zheng},
  \bibinfo{person}{Jun Shao}, {and} \bibinfo{person}{Ali~A Ghorbani}.}
  \bibinfo{year}{2020}\natexlab{}.
\newblock \showarticletitle{Achieving O (log$^3$n) communication-efficient
  privacy-preserving range query in fog-based IoT}.
\newblock \bibinfo{journal}{\emph{IEEE Internet of Things Journal}}
  \bibinfo{volume}{7}, \bibinfo{number}{6} (\bibinfo{year}{2020}),
  \bibinfo{pages}{5220--5232}.
\newblock


\bibitem[Mao et~al\mbox{.}(2023)]%
        {mao2023full}
\bibfield{author}{\bibinfo{person}{Zhengxiong Mao}, \bibinfo{person}{Hui Li},
  \bibinfo{person}{Zuyuan Huang}, \bibinfo{person}{Yuan Tian},
  \bibinfo{person}{Peng Zhao}, \bibinfo{person}{Yanan Li}, {et~al\mbox{.}}}
  \bibinfo{year}{2023}\natexlab{}.
\newblock \showarticletitle{Full Data-Processing Power Load Forecasting Based
  on Vertical Federated Learning}.
\newblock \bibinfo{journal}{\emph{Journal of Electrical and Computer
  Engineering}}  \bibinfo{volume}{2023} (\bibinfo{year}{2023}).
\newblock


\bibitem[Maulud and Abdulazeez(2020)]%
        {maulud2020review}
\bibfield{author}{\bibinfo{person}{Dastan Maulud} {and}
  \bibinfo{person}{Adnan~M Abdulazeez}.} \bibinfo{year}{2020}\natexlab{}.
\newblock \showarticletitle{A review on linear regression comprehensive in
  machine learning}.
\newblock \bibinfo{journal}{\emph{Journal of Applied Science and Technology
  Trends}} \bibinfo{volume}{1}, \bibinfo{number}{4} (\bibinfo{year}{2020}),
  \bibinfo{pages}{140--147}.
\newblock


\bibitem[McMahan et~al\mbox{.}(2017)]%
        {mcmahan2017communication}
\bibfield{author}{\bibinfo{person}{Brendan McMahan}, \bibinfo{person}{Eider
  Moore}, \bibinfo{person}{Daniel Ramage}, \bibinfo{person}{Seth Hampson},
  {and} \bibinfo{person}{Blaise~Aguera y Arcas}.}
  \bibinfo{year}{2017}\natexlab{}.
\newblock \showarticletitle{Communication-efficient learning of deep networks
  from decentralized data}. In \bibinfo{booktitle}{\emph{Artificial
  intelligence and statistics}}. PMLR, \bibinfo{pages}{1273--1282}.
\newblock


\bibitem[McMahan et~al\mbox{.}(2021)]%
        {mcmahan2021advances}
\bibfield{author}{\bibinfo{person}{H~Brendan McMahan} {et~al\mbox{.}}}
  \bibinfo{year}{2021}\natexlab{}.
\newblock \showarticletitle{Advances and open problems in federated learning}.
\newblock \bibinfo{journal}{\emph{Foundations and Trends{\textregistered} in
  Machine Learning}} \bibinfo{volume}{14}, \bibinfo{number}{1}
  (\bibinfo{year}{2021}).
\newblock


\bibitem[Mo et~al\mbox{.}(2021)]%
        {mo2021ppfl}
\bibfield{author}{\bibinfo{person}{Fan Mo}, \bibinfo{person}{Hamed Haddadi},
  \bibinfo{person}{Kleomenis Katevas}, \bibinfo{person}{Eduard Marin},
  \bibinfo{person}{Diego Perino}, {and} \bibinfo{person}{Nicolas Kourtellis}.}
  \bibinfo{year}{2021}\natexlab{}.
\newblock \showarticletitle{PPFL: privacy-preserving federated learning with
  trusted execution environments}. In \bibinfo{booktitle}{\emph{Proceedings of
  the 19th Annual International Conference on Mobile Systems, Applications, and
  Services}}. \bibinfo{pages}{94--108}.
\newblock


\bibitem[Mohassel and Rindal(2018)]%
        {mohassel2018aby3}
\bibfield{author}{\bibinfo{person}{Payman Mohassel} {and}
  \bibinfo{person}{Peter Rindal}.} \bibinfo{year}{2018}\natexlab{}.
\newblock \showarticletitle{ABY3: A mixed protocol framework for machine
  learning}. In \bibinfo{booktitle}{\emph{Proceedings of the 2018 ACM SIGSAC
  Conference on Computer and Communications Security}}.
  \bibinfo{pages}{35--52}.
\newblock


\bibitem[Mohassel and Zhang(2017)]%
        {mohassel2017secureml}
\bibfield{author}{\bibinfo{person}{Payman Mohassel} {and}
  \bibinfo{person}{Yupeng Zhang}.} \bibinfo{year}{2017}\natexlab{}.
\newblock \showarticletitle{Secureml: A system for scalable privacy-preserving
  machine learning}. In \bibinfo{booktitle}{\emph{2017 IEEE symposium on
  security and privacy (SP)}}. IEEE, \bibinfo{pages}{19--38}.
\newblock


\bibitem[Montgomery(1994)]%
        {montgomery1994survey}
\bibfield{author}{\bibinfo{person}{Peter~L Montgomery}.}
  \bibinfo{year}{1994}\natexlab{}.
\newblock \showarticletitle{A survey of modern integer factorization
  algorithms}.
\newblock \bibinfo{journal}{\emph{CWI quarterly}} \bibinfo{volume}{7},
  \bibinfo{number}{4} (\bibinfo{year}{1994}), \bibinfo{pages}{337--366}.
\newblock


\bibitem[Mothilal et~al\mbox{.}(2020)]%
        {mothilal2020explaining}
\bibfield{author}{\bibinfo{person}{Ramaravind~K Mothilal},
  \bibinfo{person}{Amit Sharma}, {and} \bibinfo{person}{Chenhao Tan}.}
  \bibinfo{year}{2020}\natexlab{}.
\newblock \showarticletitle{Explaining machine learning classifiers through
  diverse counterfactual explanations}. In
  \bibinfo{booktitle}{\emph{Proceedings of the 2020 Conference on Fairness,
  Accountability, and Transparency}}. \bibinfo{pages}{607--617}.
\newblock


\bibitem[Mouchet et~al\mbox{.}(2021)]%
        {mouchet2021multiparty}
\bibfield{author}{\bibinfo{person}{Christian Mouchet}, \bibinfo{person}{Juan
  Troncoso-Pastoriza}, \bibinfo{person}{Jean-Philippe Bossuat}, {and}
  \bibinfo{person}{Jean-Pierre Hubaux}.} \bibinfo{year}{2021}\natexlab{}.
\newblock \showarticletitle{Multiparty homomorphic encryption from
  ring-learning-with-errors}.
\newblock \bibinfo{journal}{\emph{Proceedings on Privacy Enhancing
  Technologies}} \bibinfo{volume}{2021}, \bibinfo{number}{CONF}
  (\bibinfo{year}{2021}), \bibinfo{pages}{291--311}.
\newblock


\bibitem[Mugunthan et~al\mbox{.}(2021)]%
        {mugunthan2021multi}
\bibfield{author}{\bibinfo{person}{Vaikkunth Mugunthan}, \bibinfo{person}{Pawan
  Goyal}, {and} \bibinfo{person}{Lalana Kagal}.}
  \bibinfo{year}{2021}\natexlab{}.
\newblock \showarticletitle{Multi-VFL: A Vertical Federated Learning System for
  Multiple Data and Label Owners}.
\newblock \bibinfo{journal}{\emph{arXiv preprint arXiv:2106.05468}}
  (\bibinfo{year}{2021}).
\newblock


\bibitem[Nathan and Klabjan(2017)]%
        {nathan2017optimization}
\bibfield{author}{\bibinfo{person}{Alexandros Nathan} {and}
  \bibinfo{person}{Diego Klabjan}.} \bibinfo{year}{2017}\natexlab{}.
\newblock \showarticletitle{Optimization for large-scale machine learning with
  distributed features and observations}. In
  \bibinfo{booktitle}{\emph{International Conference on Machine Learning and
  Data Mining in Pattern Recognition}}. Springer, \bibinfo{pages}{132--146}.
\newblock


\bibitem[Ni et~al\mbox{.}(2021)]%
        {ni2021vertical}
\bibfield{author}{\bibinfo{person}{Xiang Ni}, \bibinfo{person}{Xiaolong Xu},
  \bibinfo{person}{Lingjuan Lyu}, \bibinfo{person}{Changhua Meng}, {and}
  \bibinfo{person}{Weiqiang Wang}.} \bibinfo{year}{2021}\natexlab{}.
\newblock \showarticletitle{A Vertical Federated Learning Framework for Graph
  Convolutional Network}.
\newblock \bibinfo{journal}{\emph{arXiv preprint arXiv:2106.11593}}
  (\bibinfo{year}{2021}).
\newblock


\bibitem[Nock et~al\mbox{.}(2021)]%
        {nock2021impact}
\bibfield{author}{\bibinfo{person}{Richard Nock}, \bibinfo{person}{Stephen
  Hardy}, \bibinfo{person}{Wilko Henecka}, \bibinfo{person}{Hamish Ivey-Law},
  \bibinfo{person}{Jakub Nabaglo}, \bibinfo{person}{Giorgio Patrini},
  \bibinfo{person}{Guillaume Smith}, {and} \bibinfo{person}{Brian Thorne}.}
  \bibinfo{year}{2021}\natexlab{}.
\newblock \showarticletitle{The Impact of Record Linkage on Learning from
  Feature Partitioned Data}. In \bibinfo{booktitle}{\emph{International
  Conference on Machine Learning}}. PMLR, \bibinfo{pages}{8216--8226}.
\newblock


\bibitem[Ou et~al\mbox{.}(2020)]%
        {ou2020homomorphic}
\bibfield{author}{\bibinfo{person}{Wei Ou}, \bibinfo{person}{Jianhuan Zeng},
  \bibinfo{person}{Zijun Guo}, \bibinfo{person}{Wanqin Yan},
  \bibinfo{person}{Dingwan Liu}, {and} \bibinfo{person}{Stelios Fuentes}.}
  \bibinfo{year}{2020}\natexlab{}.
\newblock \showarticletitle{A homomorphic-encryption-based vertical federated
  learning scheme for rick management}.
\newblock \bibinfo{journal}{\emph{Computer Science and Information Systems}}
  \bibinfo{number}{00} (\bibinfo{year}{2020}), \bibinfo{pages}{22--22}.
\newblock


\bibitem[Owens et~al\mbox{.}(2008)]%
        {owens2008gpu}
\bibfield{author}{\bibinfo{person}{John~D Owens}, \bibinfo{person}{Mike
  Houston}, \bibinfo{person}{David Luebke}, \bibinfo{person}{Simon Green},
  \bibinfo{person}{John~E Stone}, {and} \bibinfo{person}{James~C Phillips}.}
  \bibinfo{year}{2008}\natexlab{}.
\newblock \showarticletitle{GPU computing}.
\newblock \bibinfo{journal}{\emph{Proc. IEEE}} \bibinfo{volume}{96},
  \bibinfo{number}{5} (\bibinfo{year}{2008}), \bibinfo{pages}{879--899}.
\newblock


\bibitem[Paillier(1999)]%
        {paillier1999public}
\bibfield{author}{\bibinfo{person}{Pascal Paillier}.}
  \bibinfo{year}{1999}\natexlab{}.
\newblock \showarticletitle{Public-key cryptosystems based on composite degree
  residuosity classes}. In \bibinfo{booktitle}{\emph{International conference
  on the theory and applications of cryptographic techniques}}. Springer,
  \bibinfo{pages}{223--238}.
\newblock


\bibitem[Parkhi et~al\mbox{.}(2015)]%
        {parkhi2015deep}
\bibfield{author}{\bibinfo{person}{Omkar~M Parkhi}, \bibinfo{person}{Andrea
  Vedaldi}, {and} \bibinfo{person}{Andrew Zisserman}.}
  \bibinfo{year}{2015}\natexlab{}.
\newblock \showarticletitle{Deep face recognition}.
\newblock  (\bibinfo{year}{2015}).
\newblock


\bibitem[Peng et~al\mbox{.}(2021)]%
        {peng2021differentially}
\bibfield{author}{\bibinfo{person}{Hao Peng}, \bibinfo{person}{Haoran Li},
  \bibinfo{person}{Yangqiu Song}, \bibinfo{person}{Vincent Zheng}, {and}
  \bibinfo{person}{Jianxin Li}.} \bibinfo{year}{2021}\natexlab{}.
\newblock \showarticletitle{Differentially Private Federated Knowledge Graphs
  Embedding}. In \bibinfo{booktitle}{\emph{Proceedings of the 30th ACM
  International Conference on Information \& Knowledge Management}}.
  \bibinfo{pages}{1416--1425}.
\newblock


\bibitem[Pfitzner et~al\mbox{.}(2021)]%
        {pfitzner2021federated}
\bibfield{author}{\bibinfo{person}{Bjarne Pfitzner}, \bibinfo{person}{Nico
  Steckhan}, {and} \bibinfo{person}{Bert Arnrich}.}
  \bibinfo{year}{2021}\natexlab{}.
\newblock \showarticletitle{Federated Learning in a Medical Context: A
  Systematic Literature Review}.
\newblock \bibinfo{journal}{\emph{ACM Transactions on Internet Technology
  (TOIT)}} \bibinfo{volume}{21}, \bibinfo{number}{2} (\bibinfo{year}{2021}),
  \bibinfo{pages}{1--31}.
\newblock


\bibitem[Pinkas et~al\mbox{.}(2019)]%
        {pinkas2019efficient}
\bibfield{author}{\bibinfo{person}{Benny Pinkas}, \bibinfo{person}{Thomas
  Schneider}, \bibinfo{person}{Oleksandr Tkachenko}, {and}
  \bibinfo{person}{Avishay Yanai}.} \bibinfo{year}{2019}\natexlab{}.
\newblock \showarticletitle{Efficient circuit-based PSI with linear
  communication}. In \bibinfo{booktitle}{\emph{Annual International Conference
  on the Theory and Applications of Cryptographic Techniques}}. Springer,
  \bibinfo{pages}{122--153}.
\newblock


\bibitem[Pinkas et~al\mbox{.}(2014)]%
        {pinkas2014faster}
\bibfield{author}{\bibinfo{person}{Benny Pinkas}, \bibinfo{person}{Thomas
  Schneider}, {and} \bibinfo{person}{Michael Zohner}.}
  \bibinfo{year}{2014}\natexlab{}.
\newblock \showarticletitle{Faster private set intersection based on $\{$OT$\}$
  extension}. In \bibinfo{booktitle}{\emph{23rd USENIX Security Symposium
  (USENIX Security 14)}}. \bibinfo{pages}{797--812}.
\newblock


\bibitem[Pinkas et~al\mbox{.}(2018)]%
        {pinkas2018scalable}
\bibfield{author}{\bibinfo{person}{Benny Pinkas}, \bibinfo{person}{Thomas
  Schneider}, {and} \bibinfo{person}{Michael Zohner}.}
  \bibinfo{year}{2018}\natexlab{}.
\newblock \showarticletitle{Scalable private set intersection based on OT
  extension}.
\newblock \bibinfo{journal}{\emph{ACM Transactions on Privacy and Security
  (TOPS)}} \bibinfo{volume}{21}, \bibinfo{number}{2} (\bibinfo{year}{2018}),
  \bibinfo{pages}{1--35}.
\newblock


\bibitem[Qiu et~al\mbox{.}(2022)]%
        {qiu2022your}
\bibfield{author}{\bibinfo{person}{Pengyu Qiu}, \bibinfo{person}{Xuhong Zhang},
  \bibinfo{person}{Shouling Ji}, \bibinfo{person}{Tianyu Du},
  \bibinfo{person}{Yuwen Pu}, \bibinfo{person}{Jun Zhou}, {and}
  \bibinfo{person}{Ting Wang}.} \bibinfo{year}{2022}\natexlab{}.
\newblock \showarticletitle{Your labels are selling you out: Relation leaks in
  vertical federated learning}.
\newblock \bibinfo{journal}{\emph{IEEE Transactions on Dependable and Secure
  Computing}} (\bibinfo{year}{2022}).
\newblock


\bibitem[Raina et~al\mbox{.}(2007)]%
        {raina2007self}
\bibfield{author}{\bibinfo{person}{Rajat Raina}, \bibinfo{person}{Alexis
  Battle}, \bibinfo{person}{Honglak Lee}, \bibinfo{person}{Benjamin Packer},
  {and} \bibinfo{person}{Andrew~Y Ng}.} \bibinfo{year}{2007}\natexlab{}.
\newblock \showarticletitle{Self-taught learning: transfer learning from
  unlabeled data}. In \bibinfo{booktitle}{\emph{Proceedings of the 24th
  international conference on Machine learning}}. \bibinfo{pages}{759--766}.
\newblock


\bibitem[Ren et~al\mbox{.}({[n.\,d.]})]%
        {ren2022improving}
\bibfield{author}{\bibinfo{person}{Zhenghang Ren}, \bibinfo{person}{Liu Yang},
  {and} \bibinfo{person}{Kai Chen}.} \bibinfo{year}{[n.\,d.]}\natexlab{}.
\newblock \showarticletitle{Improving Availability of Vertical Federated
  Learning: Relaxing Inference on Non-overlapping Data}.
\newblock \bibinfo{journal}{\emph{ACM Transactions on Intelligent Systems and
  Technology (TIST)}} (\bibinfo{year}{[n.\,d.]}).
\newblock


\bibitem[Rendle(2010)]%
        {rendle2010factorization}
\bibfield{author}{\bibinfo{person}{Steffen Rendle}.}
  \bibinfo{year}{2010}\natexlab{}.
\newblock \showarticletitle{Factorization machines}. In
  \bibinfo{booktitle}{\emph{2010 IEEE International conference on data
  mining}}. IEEE, \bibinfo{pages}{995--1000}.
\newblock


\bibitem[Rivest et~al\mbox{.}(2019)]%
        {rivest2019method}
\bibfield{author}{\bibinfo{person}{Ronald~L Rivest}, \bibinfo{person}{Adi
  Shamir}, {and} \bibinfo{person}{Leonard~M Adleman}.}
  \bibinfo{year}{2019}\natexlab{}.
\newblock \bibinfo{booktitle}{\emph{A method for obtaining digital signatures
  and public key cryptosystems}}.
\newblock \bibinfo{publisher}{Routledge}.
\newblock


\bibitem[Sacco et~al\mbox{.}(2020)]%
        {sacco2020federated}
\bibfield{author}{\bibinfo{person}{Alessio Sacco}, \bibinfo{person}{Flavio
  Esposito}, {and} \bibinfo{person}{Guido Marchetto}.}
  \bibinfo{year}{2020}\natexlab{}.
\newblock \showarticletitle{A federated learning approach to routing in
  challenged sdn-enabled edge networks}. In \bibinfo{booktitle}{\emph{2020 6th
  IEEE Conference on Network Softwarization (NetSoft)}}. IEEE,
  \bibinfo{pages}{150--154}.
\newblock


\bibitem[San et~al\mbox{.}(2016)]%
        {san2016efficient}
\bibfield{author}{\bibinfo{person}{Ismail San}, \bibinfo{person}{Nuray At},
  \bibinfo{person}{Ibrahim Yakut}, {and} \bibinfo{person}{Huseyin Polat}.}
  \bibinfo{year}{2016}\natexlab{}.
\newblock \showarticletitle{Efficient paillier cryptoprocessor for
  privacy-preserving data mining}.
\newblock \bibinfo{journal}{\emph{Security and communication networks}}
  \bibinfo{volume}{9}, \bibinfo{number}{11} (\bibinfo{year}{2016}),
  \bibinfo{pages}{1535--1546}.
\newblock


\bibitem[Sav et~al\mbox{.}(2021)]%
        {sav2021poseidon}
\bibfield{author}{\bibinfo{person}{Sinem Sav}, \bibinfo{person}{Apostolos
  Pyrgelis}, \bibinfo{person}{Juan~R Troncoso-Pastoriza},
  \bibinfo{person}{David Froelicher}, \bibinfo{person}{Jean-Philippe Bossuat},
  \bibinfo{person}{Joao~Sa Sousa}, {and} \bibinfo{person}{Jean-Pierre Hubaux}.}
  \bibinfo{year}{2021}\natexlab{}.
\newblock \showarticletitle{POSEIDON: privacy-preserving federated neural
  network learning}. In \bibinfo{booktitle}{\emph{28th Annual Network and
  Distributed System Security Symposium}}.
\newblock


\bibitem[Schlegel(2015)]%
        {schlegel2015deep}
\bibfield{author}{\bibinfo{person}{Daniel Schlegel}.}
  \bibinfo{year}{2015}\natexlab{}.
\newblock \showarticletitle{Deep machine learning on Gpu}.
\newblock \bibinfo{journal}{\emph{University of Heidelber-Ziti}}
  \bibinfo{volume}{12} (\bibinfo{year}{2015}).
\newblock


\bibitem[Sheikhalishahi and Martinelli(2017)]%
        {sheikhalishahi2017privacy}
\bibfield{author}{\bibinfo{person}{Mina Sheikhalishahi} {and}
  \bibinfo{person}{Fabio Martinelli}.} \bibinfo{year}{2017}\natexlab{}.
\newblock \showarticletitle{Privacy preserving clustering over horizontal and
  vertical partitioned data}. In \bibinfo{booktitle}{\emph{2017 IEEE Symposium
  on Computers and Communications (ISCC)}}. IEEE, \bibinfo{pages}{1237--1244}.
\newblock


\bibitem[Shen et~al\mbox{.}(2020)]%
        {shen2020distributed}
\bibfield{author}{\bibinfo{person}{Sheng Shen}, \bibinfo{person}{Tianqing Zhu},
  \bibinfo{person}{Di Wu}, \bibinfo{person}{Wei Wang}, {and}
  \bibinfo{person}{Wanlei Zhou}.} \bibinfo{year}{2020}\natexlab{}.
\newblock \showarticletitle{From distributed machine learning to federated
  learning: In the view of data privacy and security}.
\newblock \bibinfo{journal}{\emph{Concurrency and Computation: Practice and
  Experience}} (\bibinfo{year}{2020}).
\newblock


\bibitem[Shi et~al\mbox{.}(2022)]%
        {shi2022mvfls}
\bibfield{author}{\bibinfo{person}{Haoran Shi}, \bibinfo{person}{Yali Jiang},
  \bibinfo{person}{Han Yu}, \bibinfo{person}{Yonghui Xu}, {and}
  \bibinfo{person}{Lizhen Cui}.} \bibinfo{year}{2022}\natexlab{}.
\newblock \showarticletitle{MVFLS: Multi-participant Vertical Federated
  Learning based on Secret Sharing}. In \bibinfo{booktitle}{\emph{AAAI Workshop
  on Trustable, Verifiable and Auditable Federated Learning}}.
\newblock


\bibitem[Song et~al\mbox{.}(2021)]%
        {song2021federated}
\bibfield{author}{\bibinfo{person}{Yong Song}, \bibinfo{person}{Yuchen Xie},
  \bibinfo{person}{Hongwei Zhang}, \bibinfo{person}{Yuxin Liang},
  \bibinfo{person}{Xiaozhou Ye}, \bibinfo{person}{Aidong Yang}, {and}
  \bibinfo{person}{Ye Ouyang}.} \bibinfo{year}{2021}\natexlab{}.
\newblock \showarticletitle{Federated Learning Application on
  Telecommunication-Joint Healthcare Recommendation}. In
  \bibinfo{booktitle}{\emph{2021 IEEE 21st International Conference on
  Communication Technology (ICCT)}}. IEEE, \bibinfo{pages}{1443--1448}.
\newblock


\bibitem[Subramanya and Riggio(2021)]%
        {subramanya2021centralized}
\bibfield{author}{\bibinfo{person}{Tejas Subramanya} {and}
  \bibinfo{person}{Roberto Riggio}.} \bibinfo{year}{2021}\natexlab{}.
\newblock \showarticletitle{Centralized and federated learning for predictive
  VNF autoscaling in multi-domain 5G networks and beyond}.
\newblock \bibinfo{journal}{\emph{IEEE Transactions on Network and Service
  Management}} \bibinfo{volume}{18}, \bibinfo{number}{1}
  (\bibinfo{year}{2021}), \bibinfo{pages}{63--78}.
\newblock


\bibitem[Sun et~al\mbox{.}(2022a)]%
        {sun2022privacy}
\bibfield{author}{\bibinfo{person}{Huizhong Sun}, \bibinfo{person}{Zhenya
  Wang}, \bibinfo{person}{Yuejia Huang}, {and} \bibinfo{person}{Junda Ye}.}
  \bibinfo{year}{2022}\natexlab{a}.
\newblock \showarticletitle{Privacy-preserving vertical federated logistic
  regression without trusted third-party coordinator}. In
  \bibinfo{booktitle}{\emph{2022 The 6th International Conference on Machine
  Learning and Soft Computing}}. \bibinfo{pages}{132--138}.
\newblock


\bibitem[Sun et~al\mbox{.}(2022c)]%
        {sun2022flfhnn}
\bibfield{author}{\bibinfo{person}{Han Sun}, \bibinfo{person}{Yan Zhang},
  \bibinfo{person}{Mingxuan Li}, {and} \bibinfo{person}{Zhen Xu}.}
  \bibinfo{year}{2022}\natexlab{c}.
\newblock \showarticletitle{FLFHNN: An Efficient and Flexible Vertical
  Federated Learning Framework for Heterogeneous Neural Network}. In
  \bibinfo{booktitle}{\emph{Wireless Algorithms, Systems, and Applications:
  17th International Conference, WASA 2022, Dalian, China, November 24--26,
  2022, Proceedings, Part I}}. Springer, \bibinfo{pages}{338--350}.
\newblock


\bibitem[Sun et~al\mbox{.}(2022b)]%
        {sun2022label}
\bibfield{author}{\bibinfo{person}{Jiankai Sun}, \bibinfo{person}{Xin Yang},
  \bibinfo{person}{Yuanshun Yao}, {and} \bibinfo{person}{Chong Wang}.}
  \bibinfo{year}{2022}\natexlab{b}.
\newblock \showarticletitle{Label Leakage and Protection from Forward Embedding
  in Vertical Federated Learning}.
\newblock \bibinfo{journal}{\emph{arXiv preprint arXiv:2203.01451}}
  (\bibinfo{year}{2022}).
\newblock


\bibitem[Sun et~al\mbox{.}(2021a)]%
        {sun2021vertical}
\bibfield{author}{\bibinfo{person}{Jiankai Sun}, \bibinfo{person}{Xin Yang},
  \bibinfo{person}{Yuanshun Yao}, \bibinfo{person}{Aonan Zhang},
  \bibinfo{person}{Weihao Gao}, \bibinfo{person}{Junyuan Xie}, {and}
  \bibinfo{person}{Chong Wang}.} \bibinfo{year}{2021}\natexlab{a}.
\newblock \showarticletitle{Vertical Federated Learning without Revealing
  Intersection Membership}.
\newblock \bibinfo{journal}{\emph{arXiv preprint arXiv:2106.05508}}
  (\bibinfo{year}{2021}).
\newblock


\bibitem[Sun et~al\mbox{.}(2021b)]%
        {sun2021defending}
\bibfield{author}{\bibinfo{person}{Jiankai Sun}, \bibinfo{person}{Yuanshun
  Yao}, \bibinfo{person}{Weihao Gao}, \bibinfo{person}{Junyuan Xie}, {and}
  \bibinfo{person}{Chong Wang}.} \bibinfo{year}{2021}\natexlab{b}.
\newblock \showarticletitle{Defending against Reconstruction Attack in Vertical
  Federated Learning}.
\newblock \bibinfo{journal}{\emph{arXiv preprint arXiv:2107.09898}}
  (\bibinfo{year}{2021}).
\newblock


\bibitem[Tan et~al\mbox{.}(2020)]%
        {tan2020federated}
\bibfield{author}{\bibinfo{person}{Ben Tan}, \bibinfo{person}{Bo Liu},
  \bibinfo{person}{Vincent Zheng}, {and} \bibinfo{person}{Qiang Yang}.}
  \bibinfo{year}{2020}\natexlab{}.
\newblock \showarticletitle{A federated recommender system for online
  services}. In \bibinfo{booktitle}{\emph{Fourteenth ACM Conference on
  Recommender Systems}}. \bibinfo{pages}{579--581}.
\newblock


\bibitem[ur~Rehman et~al\mbox{.}(2020)]%
        {ur2020towards}
\bibfield{author}{\bibinfo{person}{Muhammad~Habib ur Rehman},
  \bibinfo{person}{Khaled Salah}, \bibinfo{person}{Ernesto Damiani}, {and}
  \bibinfo{person}{Davor Svetinovic}.} \bibinfo{year}{2020}\natexlab{}.
\newblock \showarticletitle{Towards blockchain-based reputation-aware federated
  learning}. In \bibinfo{booktitle}{\emph{IEEE INFOCOM 2020-IEEE Conference on
  Computer Communications Workshops (INFOCOM WKSHPS)}}. IEEE,
  \bibinfo{pages}{183--188}.
\newblock


\bibitem[Vaidya and Clifton(2002)]%
        {vaidya2002privacy}
\bibfield{author}{\bibinfo{person}{Jaideep Vaidya} {and} \bibinfo{person}{Chris
  Clifton}.} \bibinfo{year}{2002}\natexlab{}.
\newblock \showarticletitle{Privacy preserving association rule mining in
  vertically partitioned data}. In \bibinfo{booktitle}{\emph{Proceedings of the
  eighth ACM SIGKDD international conference on Knowledge discovery and data
  mining}}. \bibinfo{pages}{639--644}.
\newblock


\bibitem[Vaidya and Clifton(2003)]%
        {vaidya2003privacy}
\bibfield{author}{\bibinfo{person}{Jaideep Vaidya} {and} \bibinfo{person}{Chris
  Clifton}.} \bibinfo{year}{2003}\natexlab{}.
\newblock \showarticletitle{Privacy-preserving k-means clustering over
  vertically partitioned data}. In \bibinfo{booktitle}{\emph{Proceedings of the
  ninth ACM SIGKDD international conference on Knowledge discovery and data
  mining}}. \bibinfo{pages}{206--215}.
\newblock


\bibitem[Vaidya et~al\mbox{.}(2008)]%
        {vaidya2008privacy}
\bibfield{author}{\bibinfo{person}{Jaideep Vaidya}, \bibinfo{person}{Chris
  Clifton}, \bibinfo{person}{Murat Kantarcioglu}, {and}
  \bibinfo{person}{A~Scott Patterson}.} \bibinfo{year}{2008}\natexlab{}.
\newblock \showarticletitle{Privacy-preserving decision trees over vertically
  partitioned data}.
\newblock \bibinfo{journal}{\emph{ACM Transactions on Knowledge Discovery from
  Data (TKDD)}} \bibinfo{volume}{2}, \bibinfo{number}{3}
  (\bibinfo{year}{2008}), \bibinfo{pages}{1--27}.
\newblock


\bibitem[Valdeira et~al\mbox{.}({[n.\,d.]})]%
        {valdeira2022multi}
\bibfield{author}{\bibinfo{person}{Pedro Valdeira}, \bibinfo{person}{Yuejie
  Chi}, \bibinfo{person}{Claudia Soares}, {and} \bibinfo{person}{Joao Xavier}.}
  \bibinfo{year}{[n.\,d.]}\natexlab{}.
\newblock \showarticletitle{A Multi-Token Coordinate Descent Method for
  Vertical Federated Learning}. In \bibinfo{booktitle}{\emph{Workshop on
  Federated Learning: Recent Advances and New Challenges (in Conjunction with
  NeurIPS 2022)}}.
\newblock


\bibitem[Vatsalan et~al\mbox{.}(2017)]%
        {vatsalan2017privacy}
\bibfield{author}{\bibinfo{person}{Dinusha Vatsalan}, \bibinfo{person}{Ziad
  Sehili}, \bibinfo{person}{Peter Christen}, {and} \bibinfo{person}{Erhard
  Rahm}.} \bibinfo{year}{2017}\natexlab{}.
\newblock \showarticletitle{Privacy-preserving record linkage for big data:
  Current approaches and research challenges}.
\newblock In \bibinfo{booktitle}{\emph{Handbook of big data technologies}}.
  \bibinfo{publisher}{Springer}, \bibinfo{pages}{851--895}.
\newblock


\bibitem[Vepakomma et~al\mbox{.}(2018)]%
        {vepakomma2018split}
\bibfield{author}{\bibinfo{person}{Praneeth Vepakomma},
  \bibinfo{person}{Otkrist Gupta}, \bibinfo{person}{Tristan Swedish}, {and}
  \bibinfo{person}{Ramesh Raskar}.} \bibinfo{year}{2018}\natexlab{}.
\newblock \showarticletitle{Split learning for health: Distributed deep
  learning without sharing raw patient data}.
\newblock \bibinfo{journal}{\emph{arXiv preprint arXiv:1812.00564}}
  (\bibinfo{year}{2018}).
\newblock


\bibitem[Vidanage et~al\mbox{.}(2020)]%
        {vidanage2020graph}
\bibfield{author}{\bibinfo{person}{Anushka Vidanage}, \bibinfo{person}{Peter
  Christen}, \bibinfo{person}{Thilina Ranbaduge}, {and} \bibinfo{person}{Rainer
  Schnell}.} \bibinfo{year}{2020}\natexlab{}.
\newblock \showarticletitle{A graph matching attack on privacy-preserving
  record linkage}. In \bibinfo{booktitle}{\emph{Proceedings of the 29th ACM
  International Conference on Information \& Knowledge Management}}.
  \bibinfo{pages}{1485--1494}.
\newblock


\bibitem[Wan et~al\mbox{.}(2022)]%
        {wan2022global}
\bibfield{author}{\bibinfo{person}{Shuo Wan}, \bibinfo{person}{Jiaxun Lu},
  \bibinfo{person}{Pingyi Fan}, \bibinfo{person}{Yunfeng Shao},
  \bibinfo{person}{Chenghui Peng}, {and} \bibinfo{person}{Khaled~B Letaief}.}
  \bibinfo{year}{2022}\natexlab{}.
\newblock \showarticletitle{How Global Observation embedding in
  Vertical-Horizontal Federated Learning}. In \bibinfo{booktitle}{\emph{2022
  International Wireless Communications and Mobile Computing (IWCMC)}}. IEEE,
  \bibinfo{pages}{12--17}.
\newblock


\bibitem[Wang et~al\mbox{.}(2020a)]%
        {wang2020hybrid}
\bibfield{author}{\bibinfo{person}{Chang Wang}, \bibinfo{person}{Jian Liang},
  \bibinfo{person}{Mingkai Huang}, \bibinfo{person}{Bing Bai},
  \bibinfo{person}{Kun Bai}, {and} \bibinfo{person}{Hao Li}.}
  \bibinfo{year}{2020}\natexlab{a}.
\newblock \showarticletitle{Hybrid Differentially Private Federated Learning on
  Vertically Partitioned Data}.
\newblock \bibinfo{journal}{\emph{arXiv preprint arXiv:2009.02763}}
  (\bibinfo{year}{2020}).
\newblock


\bibitem[Wang et~al\mbox{.}(2019)]%
        {wang2019measure}
\bibfield{author}{\bibinfo{person}{Guan Wang},
  \bibinfo{person}{Charlie~Xiaoqian Dang}, {and} \bibinfo{person}{Ziye Zhou}.}
  \bibinfo{year}{2019}\natexlab{}.
\newblock \showarticletitle{Measure contribution of participants in federated
  learning}. In \bibinfo{booktitle}{\emph{2019 IEEE International Conference on
  Big Data (Big Data)}}. IEEE, \bibinfo{pages}{2597--2604}.
\newblock


\bibitem[Wang et~al\mbox{.}(2022c)]%
        {wang2022efficient}
\bibfield{author}{\bibinfo{person}{Junhao Wang}, \bibinfo{person}{Lan Zhang},
  \bibinfo{person}{Anran Li}, \bibinfo{person}{Xuanke You}, {and}
  \bibinfo{person}{Haoran Cheng}.} \bibinfo{year}{2022}\natexlab{c}.
\newblock \showarticletitle{Efficient Participant Contribution Evaluation for
  Horizontal and Vertical Federated Learning}. In
  \bibinfo{booktitle}{\emph{2022 IEEE 38th International Conference on Data
  Engineering (ICDE)}}. IEEE, \bibinfo{pages}{911--923}.
\newblock


\bibitem[Wang et~al\mbox{.}(2022b)]%
        {wang2022vertical}
\bibfield{author}{\bibinfo{person}{Leye Wang}, \bibinfo{person}{Chongru Huang},
  {and} \bibinfo{person}{Xiao Han}.} \bibinfo{year}{2022}\natexlab{b}.
\newblock \showarticletitle{Vertical federated knowledge transfer via
  representation distillation}. In \bibinfo{booktitle}{\emph{FL-IJCAI
  workshop}}.
\newblock


\bibitem[Wang et~al\mbox{.}(2022a)]%
        {wang2022feverless}
\bibfield{author}{\bibinfo{person}{Rui Wang}, \bibinfo{person}{O{\u{g}}uzhan
  Ersoy}, \bibinfo{person}{Hangyu Zhu}, \bibinfo{person}{Yaochu Jin}, {and}
  \bibinfo{person}{Kaitai Liang}.} \bibinfo{year}{2022}\natexlab{a}.
\newblock \showarticletitle{Feverless: Fast and secure vertical federated
  learning based on xgboost for decentralized labels}.
\newblock \bibinfo{journal}{\emph{IEEE Transactions on Big Data}}
  (\bibinfo{year}{2022}).
\newblock


\bibitem[Wang et~al\mbox{.}(2020b)]%
        {wang2020principled}
\bibfield{author}{\bibinfo{person}{Tianhao Wang}, \bibinfo{person}{Johannes
  Rausch}, \bibinfo{person}{Ce Zhang}, \bibinfo{person}{Ruoxi Jia}, {and}
  \bibinfo{person}{Dawn Song}.} \bibinfo{year}{2020}\natexlab{b}.
\newblock \showarticletitle{A principled approach to data valuation for
  federated learning}.
\newblock In \bibinfo{booktitle}{\emph{Federated Learning}}.
  \bibinfo{publisher}{Springer}, \bibinfo{pages}{153--167}.
\newblock


\bibitem[Wang et~al\mbox{.}(2020c)]%
        {wang2020adaptive}
\bibfield{author}{\bibinfo{person}{Zi-Jia Wang}, \bibinfo{person}{Zhi-Hui
  Zhan}, \bibinfo{person}{Sam Kwong}, \bibinfo{person}{Hu Jin}, {and}
  \bibinfo{person}{Jun Zhang}.} \bibinfo{year}{2020}\natexlab{c}.
\newblock \showarticletitle{Adaptive granularity learning distributed particle
  swarm optimization for large-scale optimization}.
\newblock \bibinfo{journal}{\emph{IEEE transactions on cybernetics}}
  \bibinfo{volume}{51}, \bibinfo{number}{3} (\bibinfo{year}{2020}),
  \bibinfo{pages}{1175--1188}.
\newblock


\bibitem[Weed(2005)]%
        {weed2005weight}
\bibfield{author}{\bibinfo{person}{Douglas~L Weed}.}
  \bibinfo{year}{2005}\natexlab{}.
\newblock \showarticletitle{Weight of evidence: a review of concept and
  methods}.
\newblock \bibinfo{journal}{\emph{Risk Analysis: An International Journal}}
  \bibinfo{volume}{25}, \bibinfo{number}{6} (\bibinfo{year}{2005}),
  \bibinfo{pages}{1545--1557}.
\newblock


\bibitem[Wei et~al\mbox{.}(2020a)]%
        {wei2020federated}
\bibfield{author}{\bibinfo{person}{Kang Wei}, \bibinfo{person}{Jun Li},
  \bibinfo{person}{Ming Ding}, \bibinfo{person}{Chuan Ma},
  \bibinfo{person}{Howard~H Yang}, \bibinfo{person}{Farhad Farokhi},
  \bibinfo{person}{Shi Jin}, \bibinfo{person}{Tony~QS Quek}, {and}
  \bibinfo{person}{H~Vincent Poor}.} \bibinfo{year}{2020}\natexlab{a}.
\newblock \showarticletitle{Federated learning with differential privacy:
  Algorithms and performance analysis}.
\newblock \bibinfo{journal}{\emph{IEEE Transactions on Information Forensics
  and Security}}  \bibinfo{volume}{15} (\bibinfo{year}{2020}),
  \bibinfo{pages}{3454--3469}.
\newblock


\bibitem[Wei et~al\mbox{.}(2021)]%
        {wei2021privacy}
\bibfield{author}{\bibinfo{person}{Qianjun Wei}, \bibinfo{person}{Qiang Li},
  \bibinfo{person}{Zhipeng Zhou}, \bibinfo{person}{ZhengQiang Ge}, {and}
  \bibinfo{person}{Yonggang Zhang}.} \bibinfo{year}{2021}\natexlab{}.
\newblock \showarticletitle{Privacy-preserving two-parties logistic regression
  on vertically partitioned data using asynchronous gradient sharing}.
\newblock \bibinfo{journal}{\emph{Peer-to-Peer Networking and Applications}}
  \bibinfo{volume}{14}, \bibinfo{number}{3} (\bibinfo{year}{2021}),
  \bibinfo{pages}{1379--1387}.
\newblock


\bibitem[Wei et~al\mbox{.}(2020b)]%
        {wei2020efficient}
\bibfield{author}{\bibinfo{person}{Shuyue Wei}, \bibinfo{person}{Yongxin Tong},
  \bibinfo{person}{Zimu Zhou}, {and} \bibinfo{person}{Tianshu Song}.}
  \bibinfo{year}{2020}\natexlab{b}.
\newblock \showarticletitle{Efficient and Fair Data Valuation for Horizontal
  Federated Learning}.
\newblock In \bibinfo{booktitle}{\emph{Federated Learning}}.
  \bibinfo{publisher}{Springer}, \bibinfo{pages}{139--152}.
\newblock


\bibitem[Weng et~al\mbox{.}(2020)]%
        {weng2020privacy}
\bibfield{author}{\bibinfo{person}{Haiqin Weng}, \bibinfo{person}{Juntao
  Zhang}, \bibinfo{person}{Feng Xue}, \bibinfo{person}{Tao Wei},
  \bibinfo{person}{Shouling Ji}, {and} \bibinfo{person}{Zhiyuan Zong}.}
  \bibinfo{year}{2020}\natexlab{}.
\newblock \showarticletitle{Privacy Leakage of Real-World Vertical Federated
  Learning}.
\newblock \bibinfo{journal}{\emph{arXiv preprint arXiv:2011.09290}}
  (\bibinfo{year}{2020}).
\newblock


\bibitem[Wu et~al\mbox{.}(2022b)]%
        {wu2022privacy}
\bibfield{author}{\bibinfo{person}{Jimmy Ming-Tai Wu}, \bibinfo{person}{Qian
  Teng}, \bibinfo{person}{Shamsul Huda}, \bibinfo{person}{Yeh-Cheng Chen},
  {and} \bibinfo{person}{Chien-Ming Chen}.} \bibinfo{year}{2022}\natexlab{b}.
\newblock \showarticletitle{A Privacy Frequent Itemsets Mining Framework for
  Collaboration in IoT Using Federated Learning}.
\newblock \bibinfo{journal}{\emph{ACM Transactions on Sensor Networks (TOSN)}}
  (\bibinfo{year}{2022}).
\newblock


\bibitem[Wu et~al\mbox{.}(2020)]%
        {wu2020privacy}
\bibfield{author}{\bibinfo{person}{Yuncheng Wu}, \bibinfo{person}{Shaofeng
  Cai}, \bibinfo{person}{Xiaokui Xiao}, \bibinfo{person}{Gang Chen}, {and}
  \bibinfo{person}{Beng~Chin Ooi}.} \bibinfo{year}{2020}\natexlab{}.
\newblock \showarticletitle{Privacy preserving vertical federated learning for
  tree-based models}.
\newblock \bibinfo{journal}{\emph{Proc. {VLDB} Endow.}} (\bibinfo{year}{2020}).
\newblock


\bibitem[Wu et~al\mbox{.}(2021)]%
        {wu2021exploiting}
\bibfield{author}{\bibinfo{person}{Zhaomin Wu}, \bibinfo{person}{Qinbin Li},
  {and} \bibinfo{person}{Bingsheng He}.} \bibinfo{year}{2021}\natexlab{}.
\newblock \showarticletitle{Exploiting Record Similarity for Practical Vertical
  Federated Learning}.
\newblock \bibinfo{journal}{\emph{arXiv preprint arXiv:2106.06312}}
  (\bibinfo{year}{2021}).
\newblock


\bibitem[Wu et~al\mbox{.}(2022a)]%
        {wu2022practical}
\bibfield{author}{\bibinfo{person}{Zhaomin Wu}, \bibinfo{person}{Qinbin Li},
  {and} \bibinfo{person}{Bingsheng He}.} \bibinfo{year}{2022}\natexlab{a}.
\newblock \showarticletitle{Practical vertical federated learning with
  unsupervised representation learning}.
\newblock \bibinfo{journal}{\emph{IEEE Transactions on Big Data}}
  (\bibinfo{year}{2022}).
\newblock


\bibitem[Xia et~al\mbox{.}(2022)]%
        {xia2022privacy}
\bibfield{author}{\bibinfo{person}{Liqiao Xia}, \bibinfo{person}{Pai Zheng},
  \bibinfo{person}{Jinjie Li}, \bibinfo{person}{Wangchujun Tang}, {and}
  \bibinfo{person}{Xiangying Zhang}.} \bibinfo{year}{2022}\natexlab{}.
\newblock \showarticletitle{Privacy-preserving gradient boosting tree: Vertical
  federated learning for collaborative bearing fault diagnosis}.
\newblock \bibinfo{journal}{\emph{IET Collaborative Intelligent Manufacturing}}
  \bibinfo{volume}{4}, \bibinfo{number}{3} (\bibinfo{year}{2022}),
  \bibinfo{pages}{208--219}.
\newblock


\bibitem[Xia et~al\mbox{.}(2023)]%
        {xia2023cascade}
\bibfield{author}{\bibinfo{person}{Wensheng Xia}, \bibinfo{person}{Ying Li},
  \bibinfo{person}{Lan Zhang}, \bibinfo{person}{Zhonghai Wu}, {and}
  \bibinfo{person}{Xiaoyong Yuan}.} \bibinfo{year}{2023}\natexlab{}.
\newblock \showarticletitle{Cascade Vertical Federated Learning Towards
  Straggler Mitigation and Label Privacy over Distributed Labels}.
\newblock \bibinfo{journal}{\emph{IEEE Transactions on Big Data}}
  (\bibinfo{year}{2023}).
\newblock


\bibitem[Xu et~al\mbox{.}(2013)]%
        {xu2013survey}
\bibfield{author}{\bibinfo{person}{Chang Xu}, \bibinfo{person}{Dacheng Tao},
  {and} \bibinfo{person}{Chao Xu}.} \bibinfo{year}{2013}\natexlab{}.
\newblock \showarticletitle{A survey on multi-view learning}.
\newblock \bibinfo{journal}{\emph{arXiv preprint arXiv:1304.5634}}
  (\bibinfo{year}{2013}).
\newblock


\bibitem[Xu et~al\mbox{.}(2019)]%
        {xu2019achieving}
\bibfield{author}{\bibinfo{person}{Depeng Xu}, \bibinfo{person}{Shuhan Yuan},
  {and} \bibinfo{person}{Xintao Wu}.} \bibinfo{year}{2019}\natexlab{}.
\newblock \showarticletitle{Achieving Differential Privacy in Vertically
  Partitioned Multiparty Learning}.
\newblock \bibinfo{journal}{\emph{arXiv preprint arXiv:1911.04587}}
  (\bibinfo{year}{2019}).
\newblock


\bibitem[Xu et~al\mbox{.}(2023)]%
        {xu2023vf}
\bibfield{author}{\bibinfo{person}{Yang Xu}, \bibinfo{person}{Xuexian Hu},
  \bibinfo{person}{Jianghong Wei}, \bibinfo{person}{Hongjian Yang}, {and}
  \bibinfo{person}{Kejia Li}.} \bibinfo{year}{2023}\natexlab{}.
\newblock \showarticletitle{VF-CART: A communication-efficient vertical
  federated framework for the CART algorithm}.
\newblock \bibinfo{journal}{\emph{Journal of King Saud University-Computer and
  Information Sciences}} \bibinfo{volume}{35}, \bibinfo{number}{1}
  (\bibinfo{year}{2023}), \bibinfo{pages}{237--249}.
\newblock


\bibitem[Yadav et~al\mbox{.}(2021)]%
        {yadav2021survey}
\bibfield{author}{\bibinfo{person}{Vijay~Kumar Yadav}, \bibinfo{person}{Nitish
  Andola}, \bibinfo{person}{Shekhar Verma}, {and} \bibinfo{person}{S
  Venkatesan}.} \bibinfo{year}{2021}\natexlab{}.
\newblock \showarticletitle{A Survey of Oblivious Transfer Protocol}.
\newblock \bibinfo{journal}{\emph{ACM Computing Surveys (CSUR)}}
  (\bibinfo{year}{2021}).
\newblock


\bibitem[Yan et~al\mbox{.}(2022)]%
        {yan2022multi}
\bibfield{author}{\bibinfo{person}{Yang Yan}, \bibinfo{person}{Guozheng Yang},
  \bibinfo{person}{Yan Gao}, \bibinfo{person}{Cheng Zang},
  \bibinfo{person}{Jiajun Chen}, {and} \bibinfo{person}{Qiang Wang}.}
  \bibinfo{year}{2022}\natexlab{}.
\newblock \showarticletitle{Multi-Participant Vertical Federated Learning Based
  Time Series Prediction}. In \bibinfo{booktitle}{\emph{Proceedings of the 8th
  International Conference on Computing and Artificial Intelligence}}.
  \bibinfo{pages}{165--171}.
\newblock


\bibitem[Yang et~al\mbox{.}(2019a)]%
        {yang2019quasi}
\bibfield{author}{\bibinfo{person}{Kai Yang}, \bibinfo{person}{Tao Fan},
  \bibinfo{person}{Tianjian Chen}, \bibinfo{person}{Yuanming Shi}, {and}
  \bibinfo{person}{Qiang Yang}.} \bibinfo{year}{2019}\natexlab{a}.
\newblock \showarticletitle{A quasi-newton method based vertical federated
  learning framework for logistic regression}.
\newblock \bibinfo{journal}{\emph{arXiv preprint arXiv:1912.00513}}
  (\bibinfo{year}{2019}).
\newblock


\bibitem[Yang et~al\mbox{.}(2020b)]%
        {yang2020federated}
\bibfield{author}{\bibinfo{person}{Kai Yang}, \bibinfo{person}{Tao Jiang},
  \bibinfo{person}{Yuanming Shi}, {and} \bibinfo{person}{Zhi Ding}.}
  \bibinfo{year}{2020}\natexlab{b}.
\newblock \showarticletitle{Federated learning via over-the-air computation}.
\newblock \bibinfo{journal}{\emph{IEEE Transactions on Wireless
  Communications}} \bibinfo{volume}{19}, \bibinfo{number}{3}
  (\bibinfo{year}{2020}), \bibinfo{pages}{2022--2035}.
\newblock


\bibitem[Yang et~al\mbox{.}(2021)]%
        {yang2021model}
\bibfield{author}{\bibinfo{person}{Kuihe Yang}, \bibinfo{person}{Ziying Song},
  \bibinfo{person}{Yingchao Zhang}, \bibinfo{person}{Yufan Zhou},
  \bibinfo{person}{Xiaohan Sun}, {and} \bibinfo{person}{Jianxuan Wang}.}
  \bibinfo{year}{2021}\natexlab{}.
\newblock \showarticletitle{Model Optimization Method Based on Vertical
  Federated Learning}. In \bibinfo{booktitle}{\emph{2021 IEEE International
  Symposium on Circuits and Systems (ISCAS)}}. IEEE, \bibinfo{pages}{1--5}.
\newblock


\bibitem[Yang et~al\mbox{.}(2019b)]%
        {yang2019federated}
\bibfield{author}{\bibinfo{person}{Qiang Yang}, \bibinfo{person}{Yang Liu},
  \bibinfo{person}{Tianjian Chen}, {and} \bibinfo{person}{Yongxin Tong}.}
  \bibinfo{year}{2019}\natexlab{b}.
\newblock \showarticletitle{Federated machine learning: Concept and
  applications}.
\newblock \bibinfo{journal}{\emph{ACM Transactions on Intelligent Systems and
  Technology (TIST)}} \bibinfo{volume}{10}, \bibinfo{number}{2}
  (\bibinfo{year}{2019}), \bibinfo{pages}{1--19}.
\newblock


\bibitem[Yang et~al\mbox{.}(2019c)]%
        {yang2019parallel}
\bibfield{author}{\bibinfo{person}{Shengwen Yang}, \bibinfo{person}{Bing Ren},
  \bibinfo{person}{Xuhui Zhou}, {and} \bibinfo{person}{Liping Liu}.}
  \bibinfo{year}{2019}\natexlab{c}.
\newblock \showarticletitle{Parallel distributed logistic regression for
  vertical federated learning without third-party coordinator}.
\newblock \bibinfo{journal}{\emph{arXiv preprint arXiv:1911.09824}}
  (\bibinfo{year}{2019}).
\newblock


\bibitem[Yang et~al\mbox{.}(2023)]%
        {yang2023hybrid}
\bibfield{author}{\bibinfo{person}{Wenti Yang}, \bibinfo{person}{Zhaoyang He},
  \bibinfo{person}{Yalei Li}, \bibinfo{person}{Haiyan Zhang}, {and}
  \bibinfo{person}{Zhitao Guan}.} \bibinfo{year}{2023}\natexlab{}.
\newblock \showarticletitle{A Hybrid Secure Two-Party Protocol for Vertical
  Federated Learning}. In \bibinfo{booktitle}{\emph{Ubiquitous Security: Second
  International Conference, UbiSec 2022, Zhangjiajie, China, December 28--31,
  2022, Revised Selected Papers}}. Springer, \bibinfo{pages}{38--51}.
\newblock


\bibitem[Yang et~al\mbox{.}(2022)]%
        {yang2022differentially}
\bibfield{author}{\bibinfo{person}{Xin Yang}, \bibinfo{person}{Jiankai Sun},
  \bibinfo{person}{Yuanshun Yao}, \bibinfo{person}{Junyuan Xie}, {and}
  \bibinfo{person}{Chong Wang}.} \bibinfo{year}{2022}\natexlab{}.
\newblock \showarticletitle{Differentially Private Label Protection in Split
  Learning}.
\newblock \bibinfo{journal}{\emph{arXiv preprint arXiv:2203.02073}}
  (\bibinfo{year}{2022}).
\newblock


\bibitem[Yang et~al\mbox{.}(2020a)]%
        {yang2020fpga}
\bibfield{author}{\bibinfo{person}{Zhaoxiong Yang}, \bibinfo{person}{Shuihai
  Hu}, {and} \bibinfo{person}{Kai Chen}.} \bibinfo{year}{2020}\natexlab{a}.
\newblock \showarticletitle{FPGA-Based Hardware Accelerator of Homomorphic
  Encryption for Efficient Federated Learning}.
\newblock \bibinfo{journal}{\emph{arXiv preprint arXiv:2007.10560}}
  (\bibinfo{year}{2020}).
\newblock


\bibitem[Yao(1986)]%
        {yao1986generate}
\bibfield{author}{\bibinfo{person}{Andrew Chi-Chih Yao}.}
  \bibinfo{year}{1986}\natexlab{}.
\newblock \showarticletitle{How to generate and exchange secrets}. In
  \bibinfo{booktitle}{\emph{27th Annual Symposium on Foundations of Computer
  Science (sfcs 1986)}}. IEEE, \bibinfo{pages}{162--167}.
\newblock


\bibitem[Yegnanarayana(2009)]%
        {yegnanarayana2009artificial}
\bibfield{author}{\bibinfo{person}{Bayya Yegnanarayana}.}
  \bibinfo{year}{2009}\natexlab{}.
\newblock \bibinfo{booktitle}{\emph{Artificial neural networks}}.
\newblock \bibinfo{publisher}{PHI Learning Pvt. Ltd.}
\newblock


\bibitem[Yekhanin(2010)]%
        {yekhanin2010private}
\bibfield{author}{\bibinfo{person}{Sergey Yekhanin}.}
  \bibinfo{year}{2010}\natexlab{}.
\newblock \showarticletitle{Private information retrieval}.
\newblock In \bibinfo{booktitle}{\emph{Locally Decodable Codes and Private
  Information Retrieval Schemes}}. \bibinfo{publisher}{Springer},
  \bibinfo{pages}{61--74}.
\newblock


\bibitem[Yu et~al\mbox{.}(2020)]%
        {yu2020sustainable}
\bibfield{author}{\bibinfo{person}{Han Yu}, \bibinfo{person}{Zelei Liu},
  \bibinfo{person}{Yang Liu}, \bibinfo{person}{Tianjian Chen},
  \bibinfo{person}{Mingshu Cong}, \bibinfo{person}{Xi Weng},
  \bibinfo{person}{Dusit Niyato}, {and} \bibinfo{person}{Qiang Yang}.}
  \bibinfo{year}{2020}\natexlab{}.
\newblock \showarticletitle{A sustainable incentive scheme for federated
  learning}.
\newblock \bibinfo{journal}{\emph{IEEE Intelligent Systems}}
  \bibinfo{volume}{35}, \bibinfo{number}{4} (\bibinfo{year}{2020}),
  \bibinfo{pages}{58--69}.
\newblock


\bibitem[Yu et~al\mbox{.}(2006)]%
        {yu2006privacy}
\bibfield{author}{\bibinfo{person}{Hwanjo Yu}, \bibinfo{person}{Jaideep
  Vaidya}, {and} \bibinfo{person}{Xiaoqian Jiang}.}
  \bibinfo{year}{2006}\natexlab{}.
\newblock \showarticletitle{Privacy-preserving svm classification on vertically
  partitioned data}. In \bibinfo{booktitle}{\emph{Pacific-asia conference on
  knowledge discovery and data mining}}. Springer, \bibinfo{pages}{647--656}.
\newblock


\bibitem[Yu et~al\mbox{.}(2022)]%
        {yu2022privacy}
\bibfield{author}{\bibinfo{person}{Xiaopeng Yu}, \bibinfo{person}{Wei Zhao},
  \bibinfo{person}{Dianhua Tang}, \bibinfo{person}{Kai Liang}, {et~al\mbox{.}}}
  \bibinfo{year}{2022}\natexlab{}.
\newblock \showarticletitle{Privacy-Preserving Vertical Collaborative Logistic
  Regression without Trusted Third-Party Coordinator}.
\newblock \bibinfo{journal}{\emph{Security and Communication Networks}}
  \bibinfo{volume}{2022} (\bibinfo{year}{2022}).
\newblock


\bibitem[Yuan et~al\mbox{.}(2022)]%
        {yuan2022byzantine}
\bibfield{author}{\bibinfo{person}{Kun Yuan}, \bibinfo{person}{Zhaoxian Wu},
  {and} \bibinfo{person}{Qing Ling}.} \bibinfo{year}{2022}\natexlab{}.
\newblock \showarticletitle{A Byzantine-Resilient Dual Subgradient Method for
  Vertical Federated Learning}. In \bibinfo{booktitle}{\emph{ICASSP 2022-2022
  IEEE International Conference on Acoustics, Speech and Signal Processing
  (ICASSP)}}. IEEE, \bibinfo{pages}{4273--4277}.
\newblock


\bibitem[Yunhong et~al\mbox{.}(2009)]%
        {yunhong2009privacy}
\bibfield{author}{\bibinfo{person}{Hu Yunhong}, \bibinfo{person}{Fang Liang},
  {and} \bibinfo{person}{He Guoping}.} \bibinfo{year}{2009}\natexlab{}.
\newblock \showarticletitle{Privacy-preserving SVM classification on vertically
  partitioned data without secure multi-party computation}. In
  \bibinfo{booktitle}{\emph{2009 fifth international conference on natural
  computation}}, Vol.~\bibinfo{volume}{1}. IEEE, \bibinfo{pages}{543--546}.
\newblock


\bibitem[Zeng et~al\mbox{.}(2022)]%
        {zeng2022over}
\bibfield{author}{\bibinfo{person}{Xiangyu Zeng}, \bibinfo{person}{Shuhao Xia},
  \bibinfo{person}{Kai Yang}, \bibinfo{person}{Youlong Wu}, {and}
  \bibinfo{person}{Yuanming Shi}.} \bibinfo{year}{2022}\natexlab{}.
\newblock \showarticletitle{Over-the-Air Computation for Vertical Federated
  Learning}. In \bibinfo{booktitle}{\emph{2022 IEEE International Conference on
  Communications Workshops (ICC Workshops)}}. IEEE, \bibinfo{pages}{788--793}.
\newblock


\bibitem[Zhang et~al\mbox{.}(2015)]%
        {zhang2015optimizing}
\bibfield{author}{\bibinfo{person}{Chen Zhang}, \bibinfo{person}{Peng Li},
  \bibinfo{person}{Guangyu Sun}, \bibinfo{person}{Yijin Guan},
  \bibinfo{person}{Bingjun Xiao}, {and} \bibinfo{person}{Jason Cong}.}
  \bibinfo{year}{2015}\natexlab{}.
\newblock \showarticletitle{Optimizing fpga-based accelerator design for deep
  convolutional neural networks}. In \bibinfo{booktitle}{\emph{Proceedings of
  the 2015 ACM/SIGDA international symposium on field-programmable gate
  arrays}}. \bibinfo{pages}{161--170}.
\newblock


\bibitem[Zhang et~al\mbox{.}(2021e)]%
        {zhang2021aegis}
\bibfield{author}{\bibinfo{person}{Cengguang Zhang}, \bibinfo{person}{Junxue
  Zhang}, \bibinfo{person}{Di Chai}, {and} \bibinfo{person}{Kai Chen}.}
  \bibinfo{year}{2021}\natexlab{e}.
\newblock \showarticletitle{Aegis: A Trusted, Automatic and Accurate
  Verification Framework for Vertical Federated Learning}. In
  \bibinfo{booktitle}{\emph{International Workshop on Federated Learning for
  User Privacy and Data Confidentiality in Conjunction with IJCAI 2021
  (FL-IJCAI'21)}}.
\newblock


\bibitem[Zhang et~al\mbox{.}(2018)]%
        {zhang2018feature}
\bibfield{author}{\bibinfo{person}{Gong-Duo Zhang}, \bibinfo{person}{Shen-Yi
  Zhao}, \bibinfo{person}{Hao Gao}, {and} \bibinfo{person}{Wu-Jun Li}.}
  \bibinfo{year}{2018}\natexlab{}.
\newblock \showarticletitle{Feature-distributed svrg for high-dimensional
  linear classification}.
\newblock \bibinfo{journal}{\emph{arXiv preprint arXiv:1802.03604}}
  (\bibinfo{year}{2018}).
\newblock


\bibitem[Zhang et~al\mbox{.}(2021d)]%
        {zhang2021federated}
\bibfield{author}{\bibinfo{person}{Huanding Zhang}, \bibinfo{person}{Tao Shen},
  \bibinfo{person}{Fei Wu}, \bibinfo{person}{Mingyang Yin},
  \bibinfo{person}{Hongxia Yang}, {and} \bibinfo{person}{Chao Wu}.}
  \bibinfo{year}{2021}\natexlab{d}.
\newblock \showarticletitle{Federated Graph Learning--A Position Paper}.
\newblock \bibinfo{journal}{\emph{arXiv preprint arXiv:2105.11099}}
  (\bibinfo{year}{2021}).
\newblock


\bibitem[Zhang et~al\mbox{.}({[n.\,d.]})]%
        {zhang2023flash}
\bibfield{author}{\bibinfo{person}{Junxue Zhang}, \bibinfo{person}{Xiaodian
  Cheng}, \bibinfo{person}{Wei Wang}, \bibinfo{person}{Liu Yang},
  \bibinfo{person}{Jinbin Hu}, {and} \bibinfo{person}{Kai Chen}.}
  \bibinfo{year}{[n.\,d.]}\natexlab{}.
\newblock \showarticletitle{{FLASH:} Towards a High-performance Hardware
  Acceleration Architecture for Cross-silo Federated Learning}. In
  \bibinfo{booktitle}{\emph{{NSDI} '23: 20th USENIX Symposium on Networked
  Systems Design and Implementation, BOSTON, MA, USA, April 17–19, 2023}}.
  \bibinfo{publisher}{{USENIX}}.
\newblock


\bibitem[Zhang et~al\mbox{.}(2022a)]%
        {zhang2022sok}
\bibfield{author}{\bibinfo{person}{Junxue Zhang}, \bibinfo{person}{Xiaodian
  Cheng}, \bibinfo{person}{Liu Yang}, \bibinfo{person}{Jinbin Hu},
  \bibinfo{person}{Ximeng Liu}, {and} \bibinfo{person}{Kai Chen}.}
  \bibinfo{year}{2022}\natexlab{a}.
\newblock \showarticletitle{SoK: Fully Homomorphic Encryption Accelerators}.
\newblock \bibinfo{journal}{\emph{arXiv preprint arXiv:2212.01713}}
  (\bibinfo{year}{2022}).
\newblock


\bibitem[Zhang et~al\mbox{.}(2022b)]%
        {zhang2022adaptive}
\bibfield{author}{\bibinfo{person}{Jie Zhang}, \bibinfo{person}{Song Guo},
  \bibinfo{person}{Zhihao Qu}, \bibinfo{person}{Deze Zeng},
  \bibinfo{person}{Haozhao Wang}, \bibinfo{person}{Qifeng Liu}, {and}
  \bibinfo{person}{Albert~Y Zomaya}.} \bibinfo{year}{2022}\natexlab{b}.
\newblock \showarticletitle{Adaptive vertical federated learning on unbalanced
  features}.
\newblock \bibinfo{journal}{\emph{IEEE Transactions on Parallel and Distributed
  Systems}} \bibinfo{volume}{33}, \bibinfo{number}{12} (\bibinfo{year}{2022}),
  \bibinfo{pages}{4006--4018}.
\newblock


\bibitem[Zhang et~al\mbox{.}(2021a)]%
        {zhang2021desirable}
\bibfield{author}{\bibinfo{person}{Qingsong Zhang}, \bibinfo{person}{Bin Gu},
  \bibinfo{person}{Zhiyuan Dang}, \bibinfo{person}{Cheng Deng}, {and}
  \bibinfo{person}{Heng Huang}.} \bibinfo{year}{2021}\natexlab{a}.
\newblock \showarticletitle{Desirable Companion for Vertical Federated
  Learning: New Zeroth-Order Gradient Based Algorithm}. In
  \bibinfo{booktitle}{\emph{Proceedings of the 30th ACM International
  Conference on Information \& Knowledge Management}}.
  \bibinfo{pages}{2598--2607}.
\newblock


\bibitem[Zhang et~al\mbox{.}(2021c)]%
        {zhang2021asysqn}
\bibfield{author}{\bibinfo{person}{Qingsong Zhang}, \bibinfo{person}{Bin Gu},
  \bibinfo{person}{Cheng Deng}, \bibinfo{person}{Songxiang Gu},
  \bibinfo{person}{Liefeng Bo}, \bibinfo{person}{Jian Pei}, {and}
  \bibinfo{person}{Heng Huang}.} \bibinfo{year}{2021}\natexlab{c}.
\newblock \showarticletitle{AsySQN: Faster Vertical Federated Learning
  Algorithms with Better Computation Resource Utilization}. In
  \bibinfo{booktitle}{\emph{Proceedings of the 27th ACM SIGKDD Conference on
  Knowledge Discovery \& Data Mining}}. \bibinfo{pages}{3917--3927}.
\newblock


\bibitem[Zhang et~al\mbox{.}(2021b)]%
        {zhang2021secure}
\bibfield{author}{\bibinfo{person}{Qingsong Zhang}, \bibinfo{person}{Bin Gu},
  \bibinfo{person}{Cheng Deng}, {and} \bibinfo{person}{Heng Huang}.}
  \bibinfo{year}{2021}\natexlab{b}.
\newblock \showarticletitle{Secure Bilevel Asynchronous Vertical Federated
  Learning with Backward Updating}. In \bibinfo{booktitle}{\emph{Proceedings of
  the AAAI Conference on Artificial Intelligence}}, Vol.~\bibinfo{volume}{35}.
  \bibinfo{pages}{10896--10904}.
\newblock


\bibitem[Zhang et~al\mbox{.}(2022e)]%
        {zhang2022secure}
\bibfield{author}{\bibinfo{person}{Rui Zhang}, \bibinfo{person}{Hongwei Li},
  \bibinfo{person}{Meng Hao}, \bibinfo{person}{Hanxiao Chen}, {and}
  \bibinfo{person}{Yuan Zhang}.} \bibinfo{year}{2022}\natexlab{e}.
\newblock \showarticletitle{Secure feature selection for vertical federated
  learning in ehealth systems}. In \bibinfo{booktitle}{\emph{ICC 2022-IEEE
  International Conference on Communications}}. IEEE,
  \bibinfo{pages}{1257--1262}.
\newblock


\bibitem[Zhang et~al\mbox{.}(2020a)]%
        {zhang2020enabling}
\bibfield{author}{\bibinfo{person}{Xiaoli Zhang}, \bibinfo{person}{Fengting
  Li}, \bibinfo{person}{Zeyu Zhang}, \bibinfo{person}{Qi Li},
  \bibinfo{person}{Cong Wang}, {and} \bibinfo{person}{Jianping Wu}.}
  \bibinfo{year}{2020}\natexlab{a}.
\newblock \showarticletitle{Enabling execution assurance of federated learning
  at untrusted participants}. In \bibinfo{booktitle}{\emph{IEEE INFOCOM
  2020-IEEE Conference on Computer Communications}}. IEEE,
  \bibinfo{pages}{1877--1886}.
\newblock


\bibitem[Zhang et~al\mbox{.}(2022c)]%
        {zhang2022embedded}
\bibfield{author}{\bibinfo{person}{Yong Zhang}, \bibinfo{person}{Ying Hu},
  \bibinfo{person}{Xiaozhi Gao}, \bibinfo{person}{Dunwei Gong},
  \bibinfo{person}{Yinan Guo}, \bibinfo{person}{Kaizhou Gao}, {and}
  \bibinfo{person}{Wanqiu Zhang}.} \bibinfo{year}{2022}\natexlab{c}.
\newblock \showarticletitle{An embedded vertical-federated feature selection
  algorithm based on particle swarm optimisation}.
\newblock \bibinfo{journal}{\emph{CAAI Transactions on Intelligence
  Technology}} (\bibinfo{year}{2022}).
\newblock


\bibitem[Zhang et~al\mbox{.}(2020b)]%
        {zhang2020vertical}
\bibfield{author}{\bibinfo{person}{Yirun Zhang}, \bibinfo{person}{Qirui Wu},
  {and} \bibinfo{person}{Mohammad Shikh-Bahaei}.}
  \bibinfo{year}{2020}\natexlab{b}.
\newblock \showarticletitle{Vertical Federated Learning Based
  Privacy-Preserving Cooperative Sensing in Cognitive Radio Networks}. In
  \bibinfo{booktitle}{\emph{2020 IEEE Globecom Workshops (GC Wkshps}}. IEEE,
  \bibinfo{pages}{1--6}.
\newblock


\bibitem[Zhang and Zhu(2020)]%
        {zhang2020additively}
\bibfield{author}{\bibinfo{person}{Yifei Zhang} {and} \bibinfo{person}{Hao
  Zhu}.} \bibinfo{year}{2020}\natexlab{}.
\newblock \showarticletitle{Additively homomorphical encryption based deep
  neural network for asymmetrically collaborative machine learning}.
\newblock \bibinfo{journal}{\emph{arXiv preprint arXiv:2007.06849}}
  (\bibinfo{year}{2020}).
\newblock


\bibitem[Zhang et~al\mbox{.}(2022d)]%
        {zhang2022data}
\bibfield{author}{\bibinfo{person}{Zhixian Zhang}, \bibinfo{person}{Xinchao
  Li}, {and} \bibinfo{person}{Shiyou Yang}.} \bibinfo{year}{2022}\natexlab{d}.
\newblock \showarticletitle{Data Pricing in Vertical Federated Learning}. In
  \bibinfo{booktitle}{\emph{2022 IEEE/CIC International Conference on
  Communications in China (ICCC)}}. IEEE, \bibinfo{pages}{932--937}.
\newblock


\bibitem[Zhang et~al\mbox{.}(2022f)]%
        {zhang2022low}
\bibfield{author}{\bibinfo{person}{Zezhong Zhang}, \bibinfo{person}{Guangxu
  Zhu}, {and} \bibinfo{person}{Shuguang Cui}.}
  \bibinfo{year}{2022}\natexlab{f}.
\newblock \showarticletitle{Low-Latency Cooperative Spectrum Sensing via
  Truncated Vertical Federated Learning}. In \bibinfo{booktitle}{\emph{2022
  IEEE Globecom Workshops (GC Wkshps)}}. IEEE, \bibinfo{pages}{1858--1863}.
\newblock


\bibitem[Zhao et~al\mbox{.}(2022a)]%
        {zhao2022accel}
\bibfield{author}{\bibinfo{person}{Jiaqi Zhao}, \bibinfo{person}{Hui Zhu},
  \bibinfo{person}{Fengwei Wang}, \bibinfo{person}{Rongxing Lu},
  \bibinfo{person}{Hui Li}, \bibinfo{person}{Zhongmin Zhou}, {and}
  \bibinfo{person}{Haitao Wan}.} \bibinfo{year}{2022}\natexlab{a}.
\newblock \showarticletitle{ACCEL: an efficient and privacy-preserving
  federated logistic regression scheme over vertically partitioned data}.
\newblock \bibinfo{journal}{\emph{Science China Information Sciences}}
  \bibinfo{volume}{65}, \bibinfo{number}{7} (\bibinfo{year}{2022}),
  \bibinfo{pages}{170307}.
\newblock


\bibitem[Zhao et~al\mbox{.}(2023)]%
        {zhao2023vflr}
\bibfield{author}{\bibinfo{person}{Jiaqi Zhao}, \bibinfo{person}{Hui Zhu},
  \bibinfo{person}{Fengwei Wang}, \bibinfo{person}{Rongxing Lu},
  \bibinfo{person}{Ermei Wang}, \bibinfo{person}{Linfeng Li}, {and}
  \bibinfo{person}{Hui Li}.} \bibinfo{year}{2023}\natexlab{}.
\newblock \showarticletitle{VFLR: An Efficient and Privacy-Preserving Vertical
  Federated Framework for Logistic Regression}.
\newblock \bibinfo{journal}{\emph{IEEE Transactions on Cloud Computing}}
  (\bibinfo{year}{2023}).
\newblock


\bibitem[Zhao et~al\mbox{.}(2022b)]%
        {zhao2022sgboost}
\bibfield{author}{\bibinfo{person}{Jiaqi Zhao}, \bibinfo{person}{Hui Zhu},
  \bibinfo{person}{Wei Xu}, \bibinfo{person}{Fengwei Wang},
  \bibinfo{person}{Rongxing Lu}, {and} \bibinfo{person}{Hui Li}.}
  \bibinfo{year}{2022}\natexlab{b}.
\newblock \showarticletitle{SGBoost: An Efficient and Privacy-Preserving
  Vertical Federated Tree Boosting Framework}.
\newblock \bibinfo{journal}{\emph{IEEE Transactions on Information Forensics
  and Security}}  \bibinfo{volume}{18} (\bibinfo{year}{2022}),
  \bibinfo{pages}{1022--1036}.
\newblock


\bibitem[Zheng et~al\mbox{.}(2018)]%
        {zheng2018blockchain}
\bibfield{author}{\bibinfo{person}{Zibin Zheng}, \bibinfo{person}{Shaoan Xie},
  \bibinfo{person}{Hong-Ning Dai}, \bibinfo{person}{Xiangping Chen}, {and}
  \bibinfo{person}{Huaimin Wang}.} \bibinfo{year}{2018}\natexlab{}.
\newblock \showarticletitle{Blockchain challenges and opportunities: A survey}.
\newblock \bibinfo{journal}{\emph{International journal of web and grid
  services}} \bibinfo{volume}{14}, \bibinfo{number}{4} (\bibinfo{year}{2018}),
  \bibinfo{pages}{352--375}.
\newblock


\bibitem[Zhou et~al\mbox{.}(2020)]%
        {zhou2020vertically}
\bibfield{author}{\bibinfo{person}{Jun Zhou}, \bibinfo{person}{Chaochao Chen},
  \bibinfo{person}{Longfei Zheng}, \bibinfo{person}{Huiwen Wu},
  \bibinfo{person}{Jia Wu}, \bibinfo{person}{Xiaolin Zheng},
  \bibinfo{person}{Bingzhe Wu}, \bibinfo{person}{Ziqi Liu}, {and}
  \bibinfo{person}{Li Wang}.} \bibinfo{year}{2020}\natexlab{}.
\newblock \showarticletitle{Vertically federated graph neural network for
  privacy-preserving node classification}.
\newblock \bibinfo{journal}{\emph{arXiv preprint arXiv:2005.11903}}
  (\bibinfo{year}{2020}).
\newblock


\bibitem[Zhou et~al\mbox{.}(2021)]%
        {zhou2021privacy}
\bibfield{author}{\bibinfo{person}{Zhou Zhou}, \bibinfo{person}{Youliang Tian},
  {and} \bibinfo{person}{Changgen Peng}.} \bibinfo{year}{2021}\natexlab{}.
\newblock \showarticletitle{Privacy-Preserving Federated Learning Framework
  with General Aggregation and Multiparty Entity Matching}.
\newblock \bibinfo{journal}{\emph{Wireless Communications and Mobile
  Computing}}  \bibinfo{volume}{2021} (\bibinfo{year}{2021}).
\newblock


\bibitem[Zhu et~al\mbox{.}(2021a)]%
        {zhu2021pivodl}
\bibfield{author}{\bibinfo{person}{Hangyu Zhu}, \bibinfo{person}{Rui Wang},
  \bibinfo{person}{Yaochu Jin}, {and} \bibinfo{person}{Kaitai Liang}.}
  \bibinfo{year}{2021}\natexlab{a}.
\newblock \showarticletitle{PIVODL: Privacy-preserving vertical federated
  learning over distributed labels}.
\newblock \bibinfo{journal}{\emph{IEEE Transactions on Artificial
  Intelligence}} (\bibinfo{year}{2021}).
\newblock


\bibitem[Zhu et~al\mbox{.}(2021b)]%
        {zhu2021from}
\bibfield{author}{\bibinfo{person}{Hangyu Zhu}, \bibinfo{person}{Haoyu Zhang},
  {and} \bibinfo{person}{Yaochu Jin}.} \bibinfo{year}{2021}\natexlab{b}.
\newblock \showarticletitle{From federated learning to federated neural
  architecture search: a survey}.
\newblock \bibinfo{journal}{\emph{Complex \& Intelligent Systems}}
  \bibinfo{volume}{7}, \bibinfo{number}{2} (\bibinfo{year}{2021}),
  \bibinfo{pages}{639--657}.
\newblock


\bibitem[Zou et~al\mbox{.}(2022)]%
        {zou2022defending}
\bibfield{author}{\bibinfo{person}{Tianyuan Zou}, \bibinfo{person}{Yang Liu},
  \bibinfo{person}{Yan Kang}, \bibinfo{person}{Wenhan Liu},
  \bibinfo{person}{Yuanqin He}, \bibinfo{person}{Zhihao Yi},
  \bibinfo{person}{Qiang Yang}, {and} \bibinfo{person}{Ya-Qin Zhang}.}
  \bibinfo{year}{2022}\natexlab{}.
\newblock \showarticletitle{Defending batch-level label inference and
  replacement attacks in vertical federated learning}.
\newblock \bibinfo{journal}{\emph{IEEE Transactions on Big Data}}
  (\bibinfo{year}{2022}).
\newblock


\end{thebibliography}

\end{document}